\documentclass[10pt,twocolumn,letterpaper]{article}

\usepackage[pagenumbers]{arxiv}

\usepackage[dvipsnames]{xcolor}
\usepackage{makecell}
\usepackage{tabularx}

\definecolor{arxivblue}{rgb}{0.21,0.49,0.74}
\usepackage[pagebackref,breaklinks,colorlinks,citecolor=arxivblue]{hyperref}

\title{MExECON: Multi-view Extended Explicit Clothed humans \\ Optimized via Normal integration}

\author{Fulden Ece U\u{g}ur  \hspace{0.6cm} Rafael Redondo \hspace{0.6cm} Albert Barreiro \hspace{0.6cm} Stefan Hristov \hspace{0.6cm} Roger Marí\\
Eurecat, Centre Tecnològic de Catalunya, Barcelona, Spain\\
{\tt\small fuldenece.ugur@eurecat.org}
}

\begin{document}
\maketitle
\begin{abstract}
This work presents MExECON, a novel pipeline for 3D reconstruction of clothed human avatars from sparse multi-view RGB images. Building on the single-view method ECON, MExECON extends its capabilities to leverage multiple viewpoints, improving geometry and body pose estimation. At the core of the pipeline is the proposed Joint Multi-view Body Optimization (JMBO) algorithm, which fits a single SMPL-X body model jointly across all input views, enforcing multi-view consistency. The optimized body model serves as a low-frequency prior that guides the subsequent surface reconstruction, where geometric details are added via normal map integration. MExECON integrates normal maps from both front and back views to accurately capture fine-grained surface details such as clothing folds and hairstyles. All multi-view gains are achieved without requiring any network re-training. Experimental results show that MExECON consistently improves fidelity over the single-view baseline and achieves competitive performance compared to modern few-shot 3D reconstruction methods.
\end{abstract}    
\section{Introduction}
\label{sec:intro}

Reconstructing high-fidelity 3D human avatars from images is a central challenge in computer vision, with broad applications in virtual reality, gaming, and telepresence. In this domain, single-view reconstruction methods are appealing due to their simple and minimal input requirements. However, they face fundamental limitations from self-occlusions and depth ambiguity, often leading to incomplete, inaccurate or non-detailed outputs. On the other hand, multi-view reconstruction methods can recover 360-degree geometry more accurately, at the expense of higher computational cost~\cite{correia20233d}. The enhanced detail provided by multi-view methods can be particularly valuable for applications involving avatars and personalized experiences in digital platforms such as the metaverse, where high-fidelity representations are essential for natural and realistic human interaction~\cite{radiah2023influence}.

\begin{figure}[t]
    \centering
    \begin{tabular}{c}
    {\small{ECON}~\cite{xiu2023econ}}   \hspace{0.48cm} {\small{MExECON} (Ours)}   \hspace{0.5cm}  {\small{Ground truth}} \\
    \hspace{-0.3cm} \includegraphics[width=1.0\linewidth]{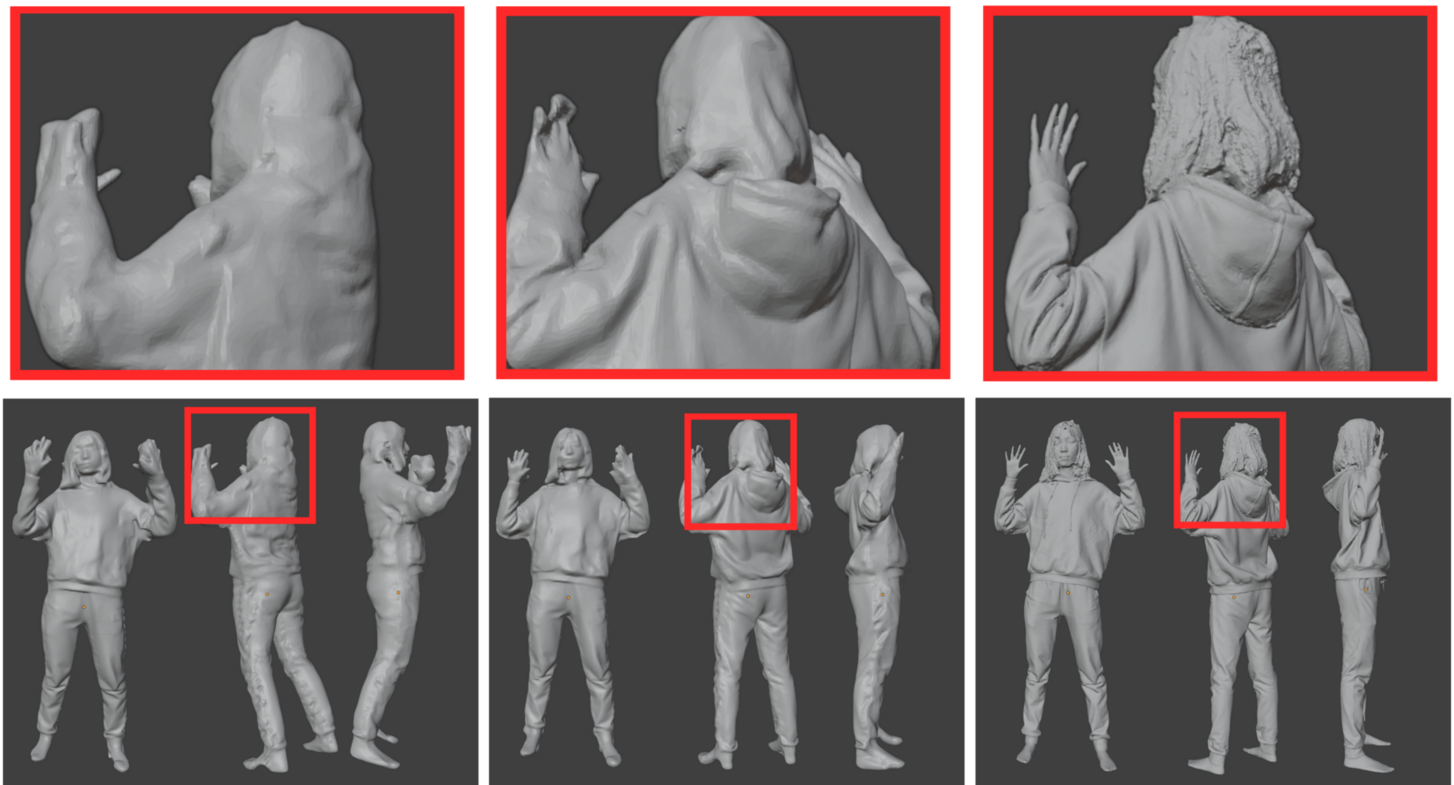}
    \end{tabular}
    \caption{Single-view avatar reconstruction methods like ECON are prone to error in unseen areas; MExECON extends ECON to multiple views, yielding more accurate pose and body geometry.}
    \label{fig:qualitative_comparison_2}
\end{figure}

In an era where image capture is increasingly affordable and efficient, it is practical and desirable for state-of-the-art methods to support both single-image and multi-view inputs within a unified framework. Avoiding separate formulations enhances usability across a wider range of applications.

Motivated by this principle, we build upon ECON~\cite{xiu2023econ}, a single-view method that combines parametric body model fitting with normal map integration to generate detailed 3D human avatars. We propose MExECON ({\bf{M}}ulti-view {\bf{Ex}}tended {\bf{ECON}}), an extended pipeline that supports arbitrary multi-view inputs while maintaining compatibility with the single-image setup. Our key contributions are:
\begin{itemize}
     \item[-] A Joint Multi-view Body Optimization (JMBO) algorithm that estimates a single, consistent body model from all available views. JMBO aligns the body pose and shape with image features while regularizing sensitive areas.
     \item[-] A surface reconstruction framework that integrates normal maps from both front and back views—unlike~\cite{xiu2023econ}, which uses only the front—without requiring distinct normal estimators for each viewpoint.
\end{itemize}
MExECON does not require any re-training of the original network components and generates more realistic avatars compared to ECON, while achieving competitive performance against state-of-the-art few-shot 3D reconstruction methods, such as VGGT~\cite{wang2025vggt}.

\section{Related work}
\label{sec:related_work}

Image-based human reconstruction methods can be broadly categorized as explicit or implicit, depending on the type of geometric representation they employ.

{\bf{Explicit or parametric methods}} represent the human body geometry using interpretable parameters related to physical attributes. Their discrete and controllable nature makes them suitable for animation tasks,  but also limits expressiveness and level of detail. The Skinned Multi-Person Linear (SMPL) model \cite{loper2015smpl} introduced a vertex-based representation controlled by shape and pose parameters, establishing an efficient, animatable body standard widely adopted.
SMPL-X \cite{pavlakos2019expressive} incorporates additional parameters for facial expressiveness and fully articulated hands. STAR \cite{osman2020star} simplified SMPL to ensure local encoding of pose-deformation. Other explicit approaches incorporate hair and clothing via 3D offsets added to the skinned body mesh~\cite{alldieck2019learning, alldieck2018video, xiang2020monoclothcap}, enabling controllable and custom, detailed avatars. Recent methods such as ECON~\cite{xiu2023econ} and CanonicalFusion~\cite{shin2024canonicalfusion} directly predict clothed meshes conditioned to an underlying SMPL-X body and aggregate surface details from image observations.

{\bf{Implicit methods}} represent the body geometry using continuous functions, such as occupancy fields or signed distance functions (SDF), rather than discrete parameterizations. This flexibility supports complex shapes and details but hinders animation and editing. PIFu~\cite{saito2019pifu} uses a network to encode clothed humans as occupancy fields by aligning per-pixel features with 3D points. PIFuHD~\cite{saito2020pifuhd} improved resolution using a corse-to-fine architecture conditioned on image-predicted normal maps. ICON~\cite{xiu2022icon} proposed a local-feature based SDF encoder guided with normal maps predicted from images and SMPL estimates. PaMIR~\cite{zheng2021pamir} conditioned the prediction of an implicit field on a voxelized SMPL mesh.
Despite these advances, implicit methods often lead to non-realistic human shapes, especially for complex or uncommon poses. This issue is often mitigated by incorporating depth or geometry priors~\cite{he2020geo, dong2022pina}.

The latest advances in image-based 3D modeling, including neural rendering and diffusion, are also being rapidly extended to human avatars.
Neural rendering methods learn geometry through differentiable volumetric rendering: NeRFs~\cite{peng2021animatable, su2021nerf, weng2022humannerf} model the scene geometry as a continuous density field, while Gaussian Splatting variants~\cite{li2024animatable, qian20243dgsavatar, shao2024splattingavatar} use an explicit set of Gaussian kernels. These methods are particularly notable for their novel view synthesis capabilities and their unsupervised nature.
Within diffusion models, text-to-3D pipelines leverage score distillation~\cite{poole2022dreamfusion} to align diffusion-generated multi-view 2D images or normal maps with underlying 3D representations (deformable parametric models~\cite{liao2024tada}, Gaussian kernels~\cite{li2025simavatar} or NeRFs~\cite{wang2023rodin, cao2024dreamavatar}). Parallel to diffusion models, large-scale transformer architectures like VGGT~\cite{wang2025vggt} offer a faster alternative, enabling dense 3D reconstruction from arbitrary multi-view collections, including up to hundreds of images.

\subsection{ECON in a nutshell}
\label{sec:econ_nutshell}

\setlength{\belowdisplayskip}{0.2cm} \setlength{\belowdisplayshortskip}{0.2cm}
\setlength{\abovedisplayskip}{0.2cm} \setlength{\abovedisplayshortskip}{0.2cm}

Our work builds on the single-view ECON~\cite{xiu2023econ} pipeline for 3D human reconstruction. ECON defines global shape and pose based on a \mbox{SMPL-X} body model that guides the reconstruction process. Free-form details such as clothes and hair are added through a normal map integration process. The main steps of the method, in sequential order, are:

\noindent{\textbf{SMPL-X initialization.}}
An SMPL-X model $\mathcal{M}^{\text{b}}$ is predicted by the pre-trained network PIXIE~\cite{feng2021collaborative} based on the RGB input front image $\mathcal{I}$.

\noindent{\textbf{Normal map prediction.}}
Clothed front $\widehat{\mathcal{N}}_{\text{F}}^{\text{c}}$ and back $\widehat{\mathcal{N}}_{\text{B}}^{\text{c}}$ normal maps are predicted by two pre-trained generator networks $\mathcal{G}$ based on
$\mathcal{I}$ and front $\mathcal{N}_{\text{F}}^{\text{b}}$ and back $\mathcal{N}_{\text{B}}^{\text{b}}$ skinned body normals calculated from $\mathcal{M}^{\text{b}}$.
\begin{equation}
    \widehat{\mathcal{N}}_{\text{F}}^{\text{c}} = \mathcal{G}_{\text{F}}^{\text{N}}(\mathcal{I}, \mathcal{N}_{\text{F}}^{\text{b}}) \quad \text{and} \quad
    \widehat{\mathcal{N}}_{\text{B}}^{\text{c}} = \mathcal{G}_{\text{B}}^{\text{N}}(\mathcal{I}, \mathcal{N}_{\text{B}}^{\text{b}}).
    \label{eq:econ_clothed_normals}
\end{equation}

\noindent{\textbf{SMPL-X optimization.}}
The SMPL-X body $\mathcal{M}^{\text{b}}$ is aligned with the clothed normal maps by minimizing the loss
\begin{equation}
    \mathcal{L}_{\text{SMPL-X}} = \mathcal{L}_{\text{Silhouette}} + \mathcal{L}_{\text{Normals}}
    + \mathcal{L}_{\text{Landmarks}},
    \label{eq:econ_smplx_opt}
\end{equation}
where $\mathcal{L}_{\text{Silhouette}}$ aligns the body with the clothed silhouette in $\mathcal{I}$, $\mathcal{L}_{\text{Normals}}$ penalizes differences between clothed normals and skinned body normals, and $\mathcal{L}_{\text{Landmarks}}$ penalizes differences between 2D joint landmarks in $\mathcal{I}$ and the reprojection of 3D joints from the SMPL-X body $\mathcal{M}^{\text{b}}$.

\noindent{\textbf{Normal integration.}}
The depth-aware bilateral normal integration d-BiNI lifts clothed normal maps $\{\widehat{\mathcal{N}}_{\text{F}}^{\text{c}}, \widehat{\mathcal{N}}_{\text{B}}^{\text{c}}\}$ to clothed depth maps $\{\widehat{\mathcal{Z}}_{\text{F}}^{\text{c}}, \widehat{\mathcal{Z}}_{\text{B}}^{\text{c}}\}$, additionally constraining the junction of both surfaces using skinned body depth maps $\{  \mathcal{Z}_{\text{F}}^{\text{b}}, \mathcal{Z}_{\text{B}}^{\text{b}} \}$ calculated from $\mathcal{M}^{\text{b}}$:
\begin{equation}
\text{d-BiNI}\left(
\widehat{\mathcal{N}}_{\text{F}}^{\text{c}}, \widehat{\mathcal{N}}_{\text{B}}^{\text{c}}, \mathcal{Z}_{\text{F}}^{\text{b}}, \mathcal{Z}_{\text{B}}^{\text{b}}
\right) \rightarrow \widehat{\mathcal{Z}}_{\text{F}}^{\text{c}}, \widehat{\mathcal{Z}}_{\text{B}}^{\text{c}}.
\label{eq:econ_dBiNI}
\end{equation}
The clothed depth maps $\{\widehat{\mathcal{Z}}_{\text{F}}^{\text{c}}$,  $\widehat{\mathcal{Z}}_{\text{B}}^{\text{c}}\}$ are jointly optimized and define the two corresponding partial surfaces $\{\mathcal{M}_{\text{F}}$,  $\mathcal{M}_{\text{B}}\}$.

\noindent{\textbf{Shape completion.}} The pre-trained IF-Nets+ network is used to fill the missing geometry in the partial surfaces $\{ \mathcal{M}_{\text{F}}$, $\mathcal{M}_{\text{B}} \}$ guided by the body model $\mathcal{M}^{\text{b}}$  to encourage shape consistency in the predicted completed
mesh $\mathcal{R}_{\text{IF}}$. Poisson Surface Reconstruction (PSR)~\cite{kazhdan2006poisson} is finally used to stitch $\mathcal{M}_{\text{F}}$ and $\mathcal{M}_{\text{B}}$ with the parts of $\mathcal{R}_{\text{IF}}$ originally missing:
\begin{equation}
    \text{PSR}\left( \, \mathcal{R}_{\text{IF}}( \widehat{\mathcal{Z}}_{\text{F}}^{\text{c}}, \widehat{\mathcal{Z}}_{\text{B}}^{\text{c}}, \mathcal{M}^{\text{b}}), \, \mathcal{M}_{\text{F}}, \, \mathcal{M}_{\text{B}}  \right) \rightarrow \mathcal{R},
    \label{eq:econ_final_mesh}
\end{equation}
where $\mathcal{R}$ is the final clothed watertight avatar mesh.
\begin{figure*}[t]
    \centering
    \includegraphics[width=1.0\linewidth]{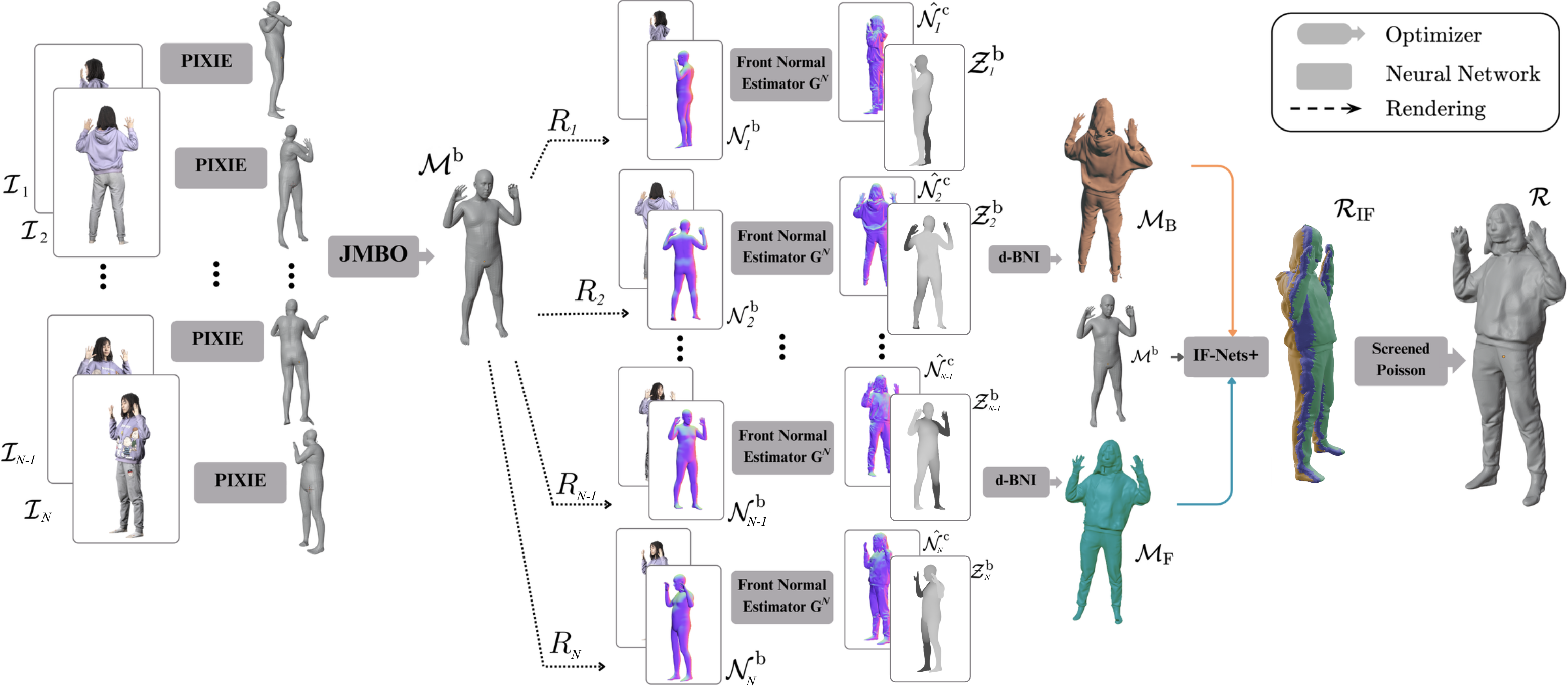}
    \caption{MExECON 3D avatar reconstruction pipeline. First, the JMBO algorithm estimates a single, optimized SMPL-X body model $\mathcal{M}^{\text{b}}$ from $N$ input RGB images.
    Second, clothed normal maps $\widehat{\mathcal{N}}_{i}^{\text{c}}$ are predicted for each view $\mathcal{I}_i$, conditioned on body normals $\mathcal{N}_{i}^{\text{b}}$ from $\mathcal{M}^{\text{b}}$. The front and back clothed normal maps, together with body depth maps $\mathcal{Z}_{i}^{\text{b}}$ from $\mathcal{M}^{\text{b}}$, feed the d-BiNI algorithm to generate two partial 3D surfaces $\{\mathcal{M}_{\text{F}}$, $\mathcal{M}_{\text{B}}\}$. These are merged by IF-Nets+ into a unified mesh $\mathcal{R}_{\text{IF}}$ to finally produce a complete watertight mesh $\mathcal{R}$.} 
    \label{fig:pipeline_overview}
\end{figure*}

\section{Methodology}
\label{sec:method}

Given a set of $N$ RGB images captured around a static subject, the proposed MExECON pipeline reconstructs a high-quality 3D mesh that captures the subject's pose and geometric details,
as illustrated in Fig.~\ref{fig:pipeline_overview}. 
Our method is structured in two key parts. The first part (Sec.~\ref{sec:jmbo}) estimates a prior multi-view consistent body model, while the second part (Sec.~\ref{sec:front_back_rec}) incorporates fine-grained details via surface reconstruction with front and back normal map integration.

\subsection{Joint Multi-view Body Optimization (JMBO)}
\label{sec:jmbo}

JMBO is the initial optimization process that estimates a single SMPL-X body~\cite{pavlakos2019expressive} ($\mathcal{M}^{\text{b}}$ in Fig.~\ref{fig:pipeline_overview}) from $N$ available views and their camera parameters. This step ensures that the body model is not affected by unseen or self-occluded body areas, as is often the case with single-view methods.

\noindent
{\bf{SMPL-X initialization.}} Each image in the input set $\{\mathcal{I}_i\}_{i=1}^N$ is used to predict a separate SMPL-X estimate $\widehat{\mathcal{M}}_i^{\text{b}}$ using PIXIE~\cite{feng2021collaborative}. The shape and pose parameters of all per-view predictions $\{\widehat{\mathcal{M}}_i^{\text{b}}\}_{i=1}^N$ are averaged to initialize $\mathcal{M}^{\text{b}}$.

\setlength{\belowdisplayskip}{0.25cm} \setlength{\belowdisplayshortskip}{0.25cm}
\setlength{\abovedisplayskip}{0.25cm} \setlength{\abovedisplayshortskip}{0.25cm}

\noindent
{\bf{SMPL-X optimization.}} The final $\mathcal{M}^{\text{b}}$ is optimized across all views by minimizing the JMBO cost function,
\begin{equation}
\mathcal{L}_{\text{SMPL-X}}=\mathcal{L}_{\text {Silhouette}}+\lambda_{n} \mathcal{L}_{\text {Normal }}+\lambda_{l} \mathcal{L}_{\text {Landmark }}+\lambda_{h} \mathcal{L}_{\text {Head}},
\label{eq:jmbo}
\end{equation}
where $\lambda_n = 0.2$ and $\lambda_l = \lambda_h = 0.1$
are empirically adjusted hyperparameters to balance the contribution of each term.
Note that Eq.~\eqref{eq:jmbo} is the multi-view extension of \eqref{eq:econ_smplx_opt}.

$\mathcal{L}_{\text{Silhouette}}$ aligns the body contour $\mathcal{S}_{i}^{\text{b}}$, projected from $\mathcal{M}^{\text{b}}$, with the clothed silhouette $\mathcal{S}_{i}^{\text{c}}$ in each image $\mathcal{I}_i$:
\begin{equation}
\mathcal{L}_{\text {Silhouette}}= \sum\nolimits_{i}^N||\mathcal{S}_{i}^{\text{b}}-\mathcal{S}_{i}^{\text{c}}||_1.
\label{eq:silhouette_loss}
\end{equation}

$\mathcal{L}_{\text{Normals}}$ aligns the body normals $\mathcal{N}_{i}^{\text{b}}$, rendered from $\mathcal{M}^{\text{b}}$, with the clothed normal maps $\widehat{\mathcal{N}}_{i}^{\text{c}}$:
\begin{equation}
\mathcal{L}_{\text {Normals}}= \sum\nolimits_{i}^N||\mathcal{N}_i^{\text{b}} - \widehat{\mathcal{N}}_{i}^{\text{c}}||_1.
\label{eq:normals_loss}
\end{equation}
Here, all $\widehat{\mathcal{N}}_{i}^{\text{c}}$ are predicted using the ECON front normal estimator $\mathcal{G}_{\text{F}}^{\text{N}}$ followed by a rotation $R_{\text{F} \to i}$ to map the normal coordinates from frontal view $\mathcal{I}_{\text{F}}$ to the corresponding view $\mathcal{I}_i$ according to the known camera parameters.

$\mathcal{L}_{\text{Landmarks}}$ aligns the body joints $\mathcal{J}_{i}^{\text{b}}$ projected from $\mathcal{M}^{\text{b}}$ with 2D image landmarks $\mathcal{J}_{i}^{\text{c}}$ detected via MediaPipe~\cite{lugaresi2019mediapipe}.
\begin{equation}
\mathcal{L}_{\text {Landmarks}}=\sum\nolimits_{i}^N \frac{\sum_{k} c_{i, k} \cdot||\mathcal{J}_{i, k}^{\text{b}}-\mathcal{J}_{i, k}^{\text{c}}||_{2}}{\sum_{k} c_{i, k}},
\label{eq:landmarks_loss}
\end{equation}
where $c_{i,k}$ is a confidence score for the $k$-th landmark detection in image $\mathcal{I}_i$~\cite{lugaresi2019mediapipe}.

\begin{figure}
\centering
    \begin{tabular}{@{\hskip -0cm}c@{\hskip 0.1cm}c@{\hskip 0.1cm}c}
        {\footnotesize JMBO without $\mathcal{L}_{\text{Head}}$} & {\footnotesize JMBO with $\mathcal{L}_{\text{Head}}$} & {\footnotesize Ground-truth avatar } \\
    \includegraphics[width=0.32\linewidth]{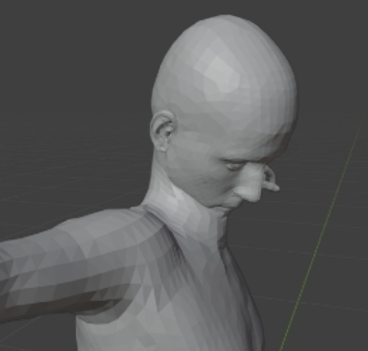} &
    \includegraphics[width=0.32\linewidth]{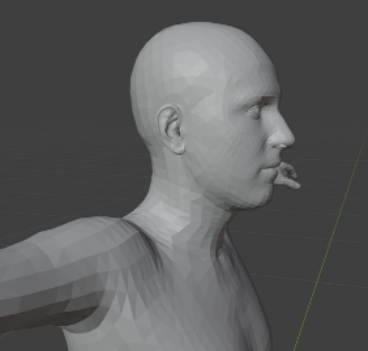} &
    \includegraphics[width=0.32\linewidth]{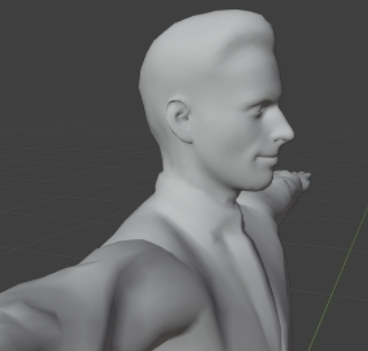}
    \end{tabular}
    \caption{JMBO introduces a regularization term $\mathcal{L}_{\text{Head}}$ to prevent unrealistic head positions in the multi-view consistent body model.}
    \label{fig:head_tilt}
\end{figure}

$\mathcal{L}_{\text{Head}}$ is a novel regularization term that encourages the body model $\mathcal{M}^{\text{b}}$ to maintain a forward-facing, upright head orientation during the multi-view optimization process:
\begin{equation}
\mathcal{L}_{\text{Head}} = 1 - \frac{\mathbf{v}_h \cdot \mathbf{v}_y}{\|\mathbf{v}_h\| \|\mathbf{v}_y\|},
\label{eq:head_pitch}
\end{equation}
where $\mathbf{v}_h$ is the head pitch vector and $\mathbf{v}_y=\left[0,1,0\right]$ is the canonical forward-facing direction in 3D world coordinates. As shown in Fig.~\ref{fig:head_tilt}, the inclusion of $\mathcal{L}_{\text{Head}}$ was motivated by the empirical observation that head position is highly sensitive to the multi-view landmark fitting. Note that head and hands concentrate the majority of landmarks used in~\eqref{eq:landmarks_loss}.

As shown in Fig.~\ref{fig:pipeline_overview}, the optimized SMPL-X body $\mathcal{M}^{\text{b}}$ estimated by JMBO is a skinned model and lacks accessory details such as hair or clothing. The main function
of this multi-view consistent body model is to act as a global prior for the subsequent front-back detailed reconstruction.

\subsection{Front-back normal-guided 3D reconstruction}
\label{sec:front_back_rec}

{\bf{Front-back reconstruction.}} Similar to ECON, the proposed MExECON reconstructs and merges two distinct front and back partial surfaces $\{\mathcal{M}_{\text{F}}, \mathcal{M}_{\text{B}}\}$, as illustrated in Fig.~\ref{fig:pipeline_overview}. Fine-grained details are incorporated into the partial surfaces by integrating front and back clothed normal maps $\{\widehat{\mathcal{N}}_{\text{F}}^{\text{c}},\widehat{\mathcal{N}}_{\text{B}}^{\text{c}}\}$ through \mbox{d-BiNI} optimization, as in \eqref{eq:econ_dBiNI}. However, here the partial surfaces are aligned with the multi-view SMPL-X estimated with JMBO.

\begin{figure}[t]
\centering
   \begin{tabular}{@{\hskip -0.0cm}c@{\hskip 0.01cm}@{\hskip 0.1cm}c@{\hskip 0.08cm}|@{\hskip 0.08cm}c@{\hskip 0.08cm}|@{\hskip 0.05cm}c@{\hskip -0.18cm}c@{\hskip 0cm}}
    & {\footnotesize Input} & {\footnotesize Front normals $\widehat{\mathcal{N}}_{\text{F}}^{\text{c}}$} & \multicolumn{2}{c}{{\footnotesize Back normals $\widehat{\mathcal{N}}_{\text{B}}^{\text{c}}$}} \\ \hline
    & & & & \\[-0.35cm]
    \rotatebox[origin=lb]{90}{\footnotesize \hspace{0.7cm} ECON~\cite{xiu2023econ}} &
    \includegraphics[width=0.18\linewidth]{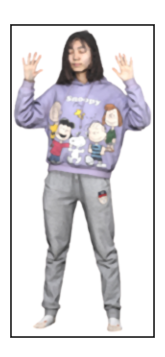} &
    \includegraphics[width=0.18\linewidth]{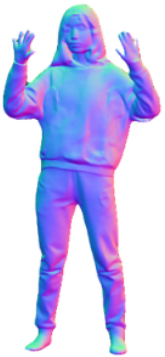} &
    \multicolumn{2}{c}{\includegraphics[width=0.18\linewidth]{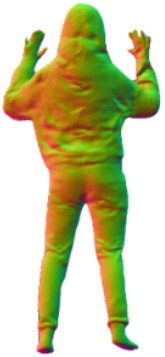}} \\
    [-0.1cm]
    & {\footnotesize $\mathcal{I}_{\text{F}}$} &
    {\footnotesize $\mathcal{G}_{\text{F}}^{\text{N}}(\mathcal{I}_{\text{F}}, \mathcal{N}_{\text{F}}^{\text{b}})$ } &
    \multicolumn{2}{c}{{\footnotesize $\mathcal{G}_{\text{B}}^{\text{N}}(\mathcal{I}_{\text{F}}, \mathcal{N}_{\text{B}}^{\text{b}})$ }} 
    \\[0.1cm]  \hline
    & & & &  \\[-0.35cm]
    \rotatebox[origin=lb]{90}{\footnotesize \hspace{0.3cm} MExECON (Ours)} &
    \includegraphics[width=0.18\linewidth]{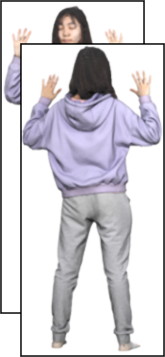} &
    \includegraphics[width=0.18\linewidth]{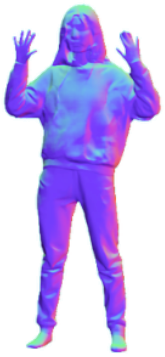} &
    \includegraphics[width=0.18\linewidth]{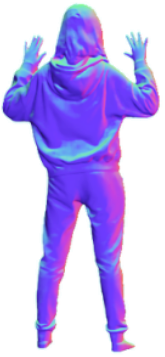} &
    \includegraphics[width=0.18\linewidth]{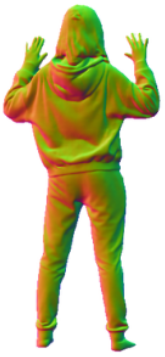} \\[-0.1cm]
    & {\footnotesize $\mathcal{I}_{\text{F}}$ and $\mathcal{I}_{\text{B}}$} &
    {\footnotesize $\mathcal{G}_{\text{F}}^{\text{N}}(\mathcal{I}_{\text{F}}, \mathcal{N}_{\text{F}}^{\text{b}})$ } &
    {\footnotesize $\mathcal{G}_{\text{F}}^{\text{N}}(\mathcal{I}_{\text{B}}, \mathcal{N}_{\text{B}}^{\text{b}})$} &
    {\hspace{0.2cm} \footnotesize $R_{\text{F} \to  \text{B}} \!\!\left( \mathcal{G}_{\text{F}}^{\text{N}}(\mathcal{I}_{\text{B}}, \mathcal{N}_{\text{B}}^{\text{b}}) \right)$} 
\end{tabular}
    \caption{ECON vs MExECON prediction of front and back clothed normal maps, $\widehat{\mathcal{N}}_{\text{F}}^{\text{c}}$ and $\widehat{\mathcal{N}}_{\text{B}}^{\text{c}}$. ECON relies on two different front and back normal estimators, $\mathcal{G}_{\text{F}}^{\text{N}}$ and $\mathcal{G}_{\text{B}}^{\text{N}}$.
    MExECON relies solely on $\mathcal{G}_{\text{F}}^{\text{N}}$
    and improves 
    fine-grained details in $\widehat{\mathcal{N}}_{\text{B}}^{\text{c}}$
    by using the now-available back view $\mathcal{I}_{\text{B}}$ followed by a known camera rotation $R_{\text{F} \to \text{B}}$. In addition, the clothed normal map prediction takes advantage of the body normal maps $\{\mathcal{N}_{\text{F}}^{\text{b}},  \mathcal{N}_{\text{B}}^{\text{b}}\}$ calculated from the SMPL-X body model optimized with JMBO. }
    \label{fig:improved_back_normals}
\end{figure}

\setlength{\belowdisplayskip}{0.25cm} \setlength{\belowdisplayshortskip}{0.25cm}
\setlength{\abovedisplayskip}{0.25cm} \setlength{\abovedisplayshortskip}{0.25cm}

MExECON also introduces a key modification in the computation of the back clothed normal map $\widehat{\mathcal{N}}_{\text{B}}^{\text{c}}$ integrated into the back partial surface $\mathcal{M}_{\text{B}}$. Instead of inferring this map from the front view $\mathcal{I}_{\text{F}}$, as in ECON, MExECON leverages the now-available back view $\mathcal{I}_{\text{B}}$ to predict $\widehat{\mathcal{N}}_{\text{B}}^{\text{c}}$. More specifically, $\widehat{\mathcal{N}}_{\text{B}}^{\text{c}}$ is predicted by feeding the ECON front normal estimator $\mathcal{G}_{\text{F}}^{\text{N}}$ with $\mathcal{I}_{\text{B}}$ and the skinned body back normals $\mathcal{N}_{\text{B}}^{\text{b}}$ from the SMPL-X prior, then transforming the output with the known relative rotation between the front and back cameras $R_{\text{F} \to \text{B}}$, expressed as:
\begin{equation}
  \widehat{\mathcal{N}}_{\text{B}}^{\text{c}} = R_{\text{F} \to \text{B}} \!\left( \mathcal{G}_{\text{F}}^{\text{N}}(\mathcal{I}_{\text{B}},  \mathcal{N}_{\text{B}}^{\text{b}}) \right).
  \label{eq:mexecon_back_normals}
\end{equation}

As shown in Fig.~\ref{fig:improved_back_normals}, this modification significantly improves the prediction of the back clothed normal map, reconstructing fine-grained details such as accessory objects (hoods, belts, or backpacks) or complex hairstyles (long hair, buns, or ponytails) observed in the back view. Such details are lost or smoothed out in the equivalent 
map predicted by ECON.
In addition, Eq.~\eqref{eq:mexecon_back_normals} also circumvents the need for distinct normal estimator networks for the front and back, and can be extended to any other viewpoint.

\vspace{0.2cm}

\begin{figure}
\centering
    \begin{tabular}{@{\hskip -0cm}c@{\hskip 0.1cm}c@{\hskip 0.1cm}c}
    \includegraphics[width=0.32\linewidth]{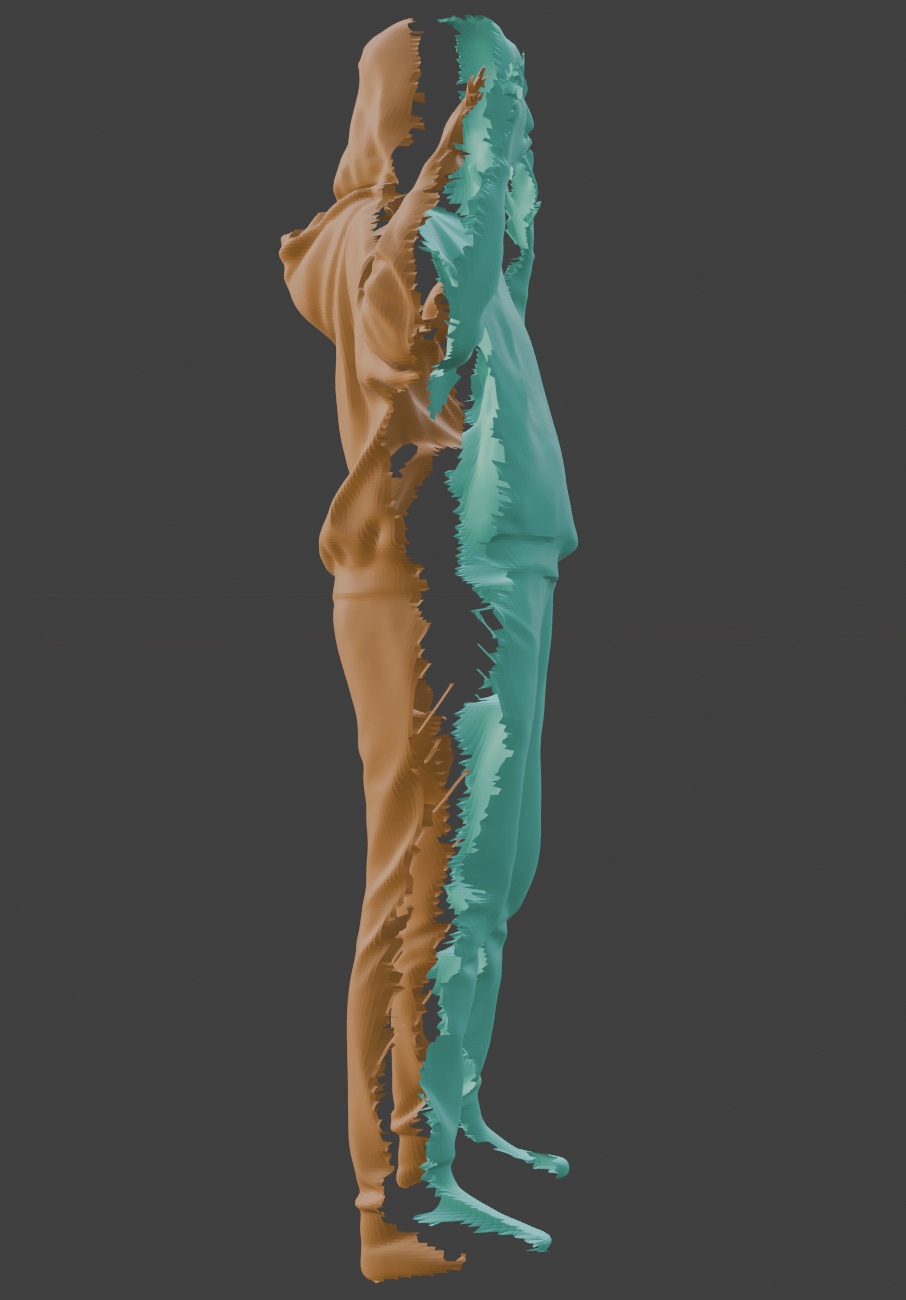} &
    \includegraphics[width=0.32\linewidth]{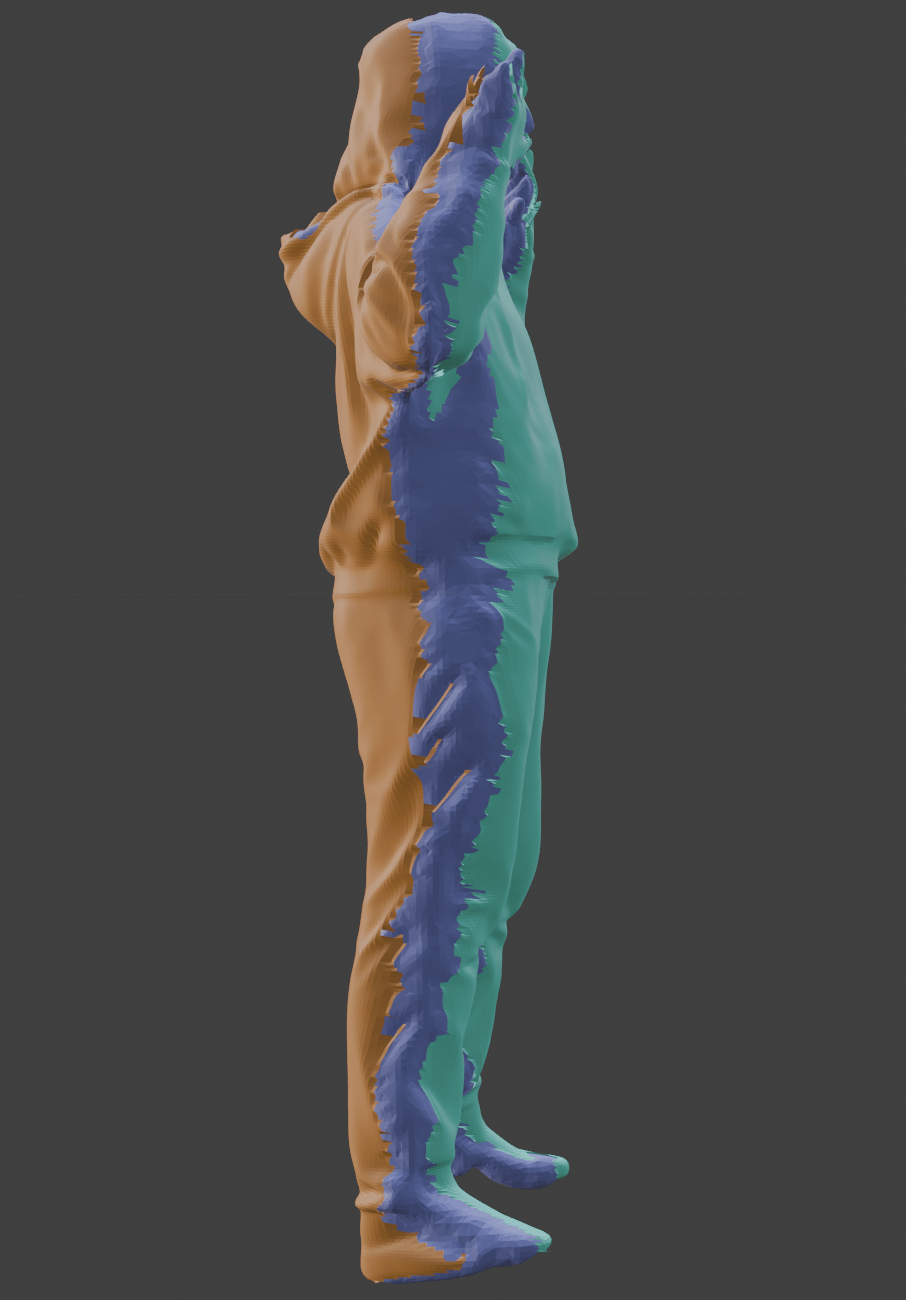} &
    \includegraphics[width=0.32\linewidth]{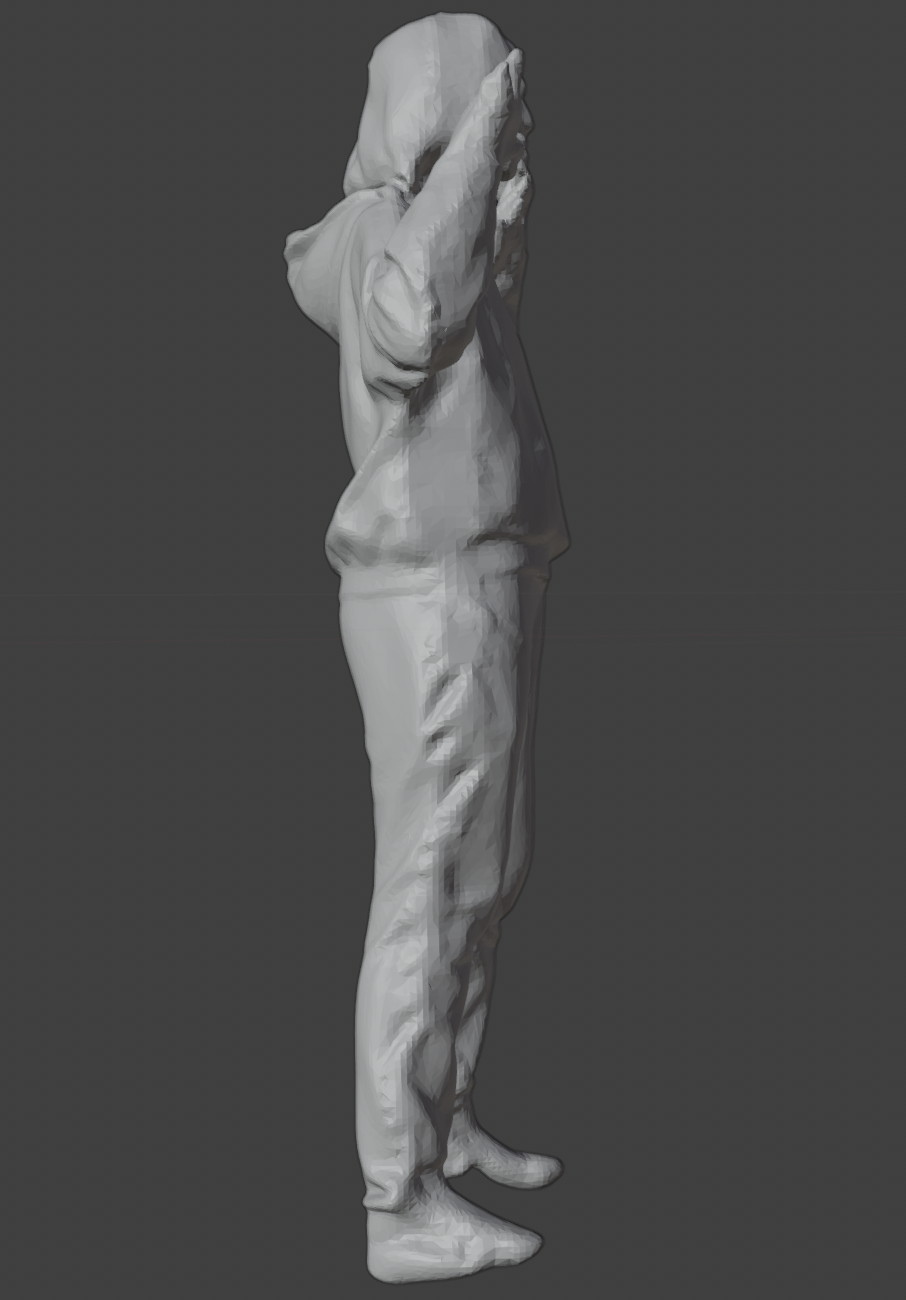}
    \end{tabular}
    \caption{Surface completion with IF-Nets+. Left to right: Input partial surfaces $\mathcal{M}_{\text{F}}$ (green) and $\mathcal{M}_{\text{B}}$ (orange), partial surfaces and additional area $\mathcal{R}_{\text{IF}}$ filled by If-Nets+ (purple), final watertight mesh $\mathcal{R}$ extracted with Poisson Surface Reconstruction~\cite{kazhdan2006poisson}.}
    \label{fig:surface_completion}
\end{figure}

\noindent
{\bf{Shape completion.}} 
This step aims to fill the gap between the partial surfaces $\{\mathcal{M}_{\text{F}}, \mathcal{M}_{\text{B}}\}$. 
MExECON retains compatibility with the original formulation~\eqref{eq:econ_final_mesh} behind the extraction of the final mesh $\mathcal{R}$. As illustrated in Fig.~\ref{fig:surface_completion}, the IF-Nets+ network is used to predict a continuous occupancy field of the entire body based on the SMPL-X model $\mathcal{M}^{\text{b}}$ and the depth information of the clothed front-back surfaces. A smooth, complete mesh $\mathcal{R}_{\text{IF}}$ is extracted from the IF-Nets+ occupancy field using Marching cubes~\cite{lorensen1998marching}. The final watertight mesh $\mathcal{R}$ is obtained by using PSR~\cite{kazhdan2006poisson} to stitch $\mathcal{M}_{\text{F}}$ and $\mathcal{M}_{\text{B}}$ with the new filled regions.

\section{Experiments}
\label{sec:experiments}

\subsection{Data}

A test set of 20 clothed human avatars was curated from the THuman 2.1 dataset~\cite{tao2021function4d}, which provides high-quality ground-truth 3D scans in the form of textured meshes. The test set spans a diverse range of clothing styles, poses and body shapes. The selected THuman avatar identifiers are: 0004, 0006, 0007, 0019, 0021, 0023, 0053, 0054, 0055, 0057, 0070, 0071, 0072, 0074, 0081, 0083, 0120, 0170, 0181 and 0445.

\begin{figure}
\centering
    \includegraphics[width=0.99\linewidth]{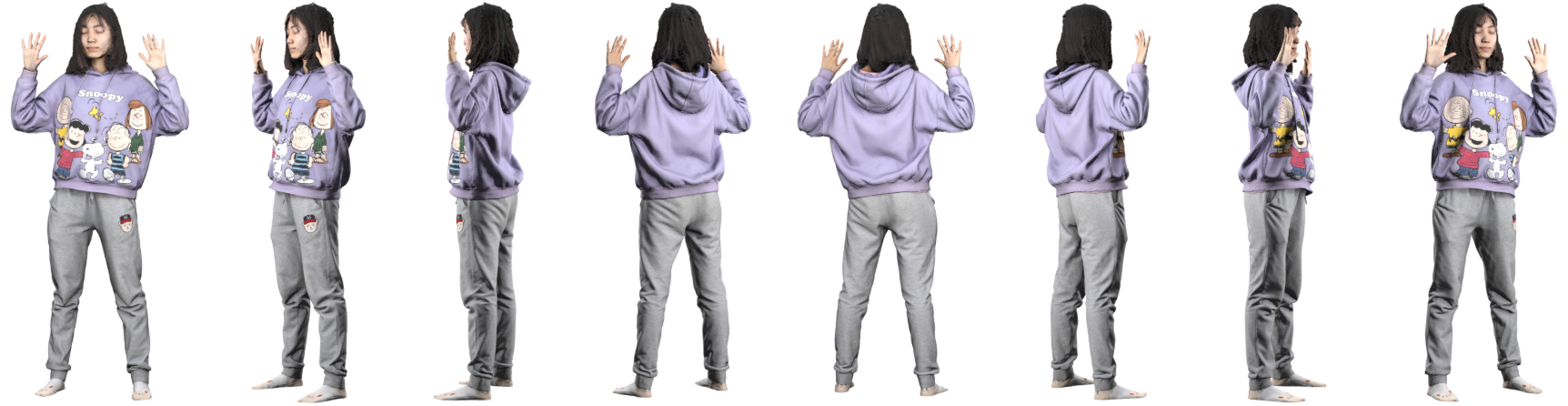}
    \caption{Example 8-view RGB image set rendered with virtual cameras positioned around subject 0021 from THuman~2.1~\cite{tao2021function4d}.}
    \label{fig:input_rgb}
\end{figure}

For each avatar, a multi-view image set was rendered using Blender~\cite{blender}, employing eight virtual cameras uniformly distributed over 360 degrees around the subject, starting from a frontal view. Image size was set to $1080 \times 1920$ pixels and camera elevation was fixed at eye level across all viewpoints. The resulting synthetic RGB images, along with the corresponding intrinsic and extrinsic camera parameters, were used as input to our MExECON pipeline. Fig.~\ref{fig:input_rgb} shows an example of a complete 8-view set.

\subsection{Metrics}
\label{sec:metrics}

The quality of the reconstructed meshes is assessed against the ground-truth meshes from the THuman 2.1 dataset~\cite{tao2021function4d} using four standard metrics: Chamfer Distance, Point-to-Surface (P2S) Accuracy, P2S Completeness, and Normal Consistency. All metrics are reported in millimeters (mm) or degrees (deg.) and averaged over the 20-avatar test set. In all metrics, lower values indicate higher geometric fidelity.

\noindent
\textbf{Chamfer Distance.} Bi-directional measure of geometric accuracy, i.e., average distance from each surface point to the closest points on the ground-truth surface and vice versa.

\noindent
\textbf{P2S Accuracy.} One-directional measure of geometric accuracy, i.e., average distance from each surface point to the closest points on the ground-truth surface. 

\noindent
\textbf{P2S Completeness.} One-directional measure of how completely the reconstruction covers the reference geometry, i.e., average distance from each ground-truth surface point to the closest points on the reconstructed mesh.

\noindent
\textbf{Normal Consistency.} Angular difference between the reconstructed surface normals and the ground-truth normals. It captures the preservation of fine details, such as clothing folds. For each surface point, we compute the normal vector error in degrees using point‑to‑surface matching as in \cite{Knapitsch2017}. Unlike the L2 difference between normal maps from fixed viewpoints proposed in ECON, this metric is view independent and can be applied to any pair of aligned meshes. 

\subsection{Evaluation}
The conducted experiments show that MExECON substantially improves upon the single-view ECON baseline by effectively leveraging multi-view information. Separate evaluations of the JMBO body prior and the final clothed mesh geometry are discussed to emphasize the role of each core contribution of MExECON.

\noindent
\textbf{SMPL-X body model accuracy.} The estimation of an accurate SMPL-X body prior is critical to achieve a high-fidelity reconstruction. We compared the proposed JMBO methodology (Sec.~\ref{sec:jmbo}) for estimating the SMPL-X parameters against the original ECON single front-view prediction and a straightforward multi-view average from independent per-view predictions. As shown in Table~\ref{tab:jmbo_metrics}, JMBO significantly improves the output body shape and pose, as well as the alignment with the available 2D images. The total loss is reduced by over 74\% (from 0.229 to 0.059) and corresponds to the weighted sum of the silhouette, landmark, and normal losses defined in Eq.~\eqref{eq:jmbo}.

\begin{figure*}
\centering
    \begin{tabular}{@{\hskip -0cm}c@{\hskip 0.1cm}c@{\hskip 0.1cm}c}
    {\small Front-view SMPL-X (ECON)} & {\small 8-view JMBO SMPL-X (Ours)} & {\small Ground-truth avatar} \\
    \includegraphics[width=0.32\linewidth]{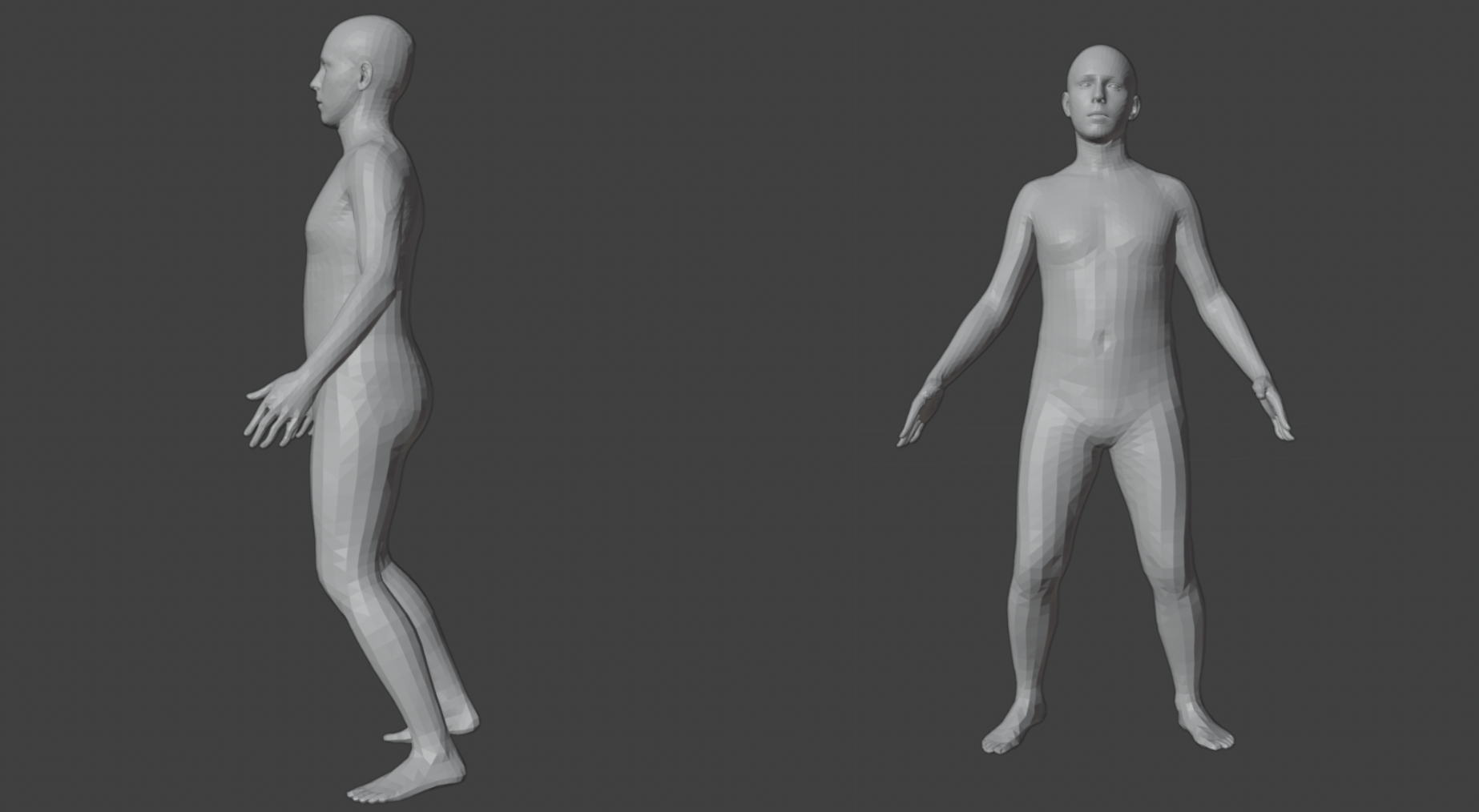} &
    \includegraphics[width=0.32\linewidth]{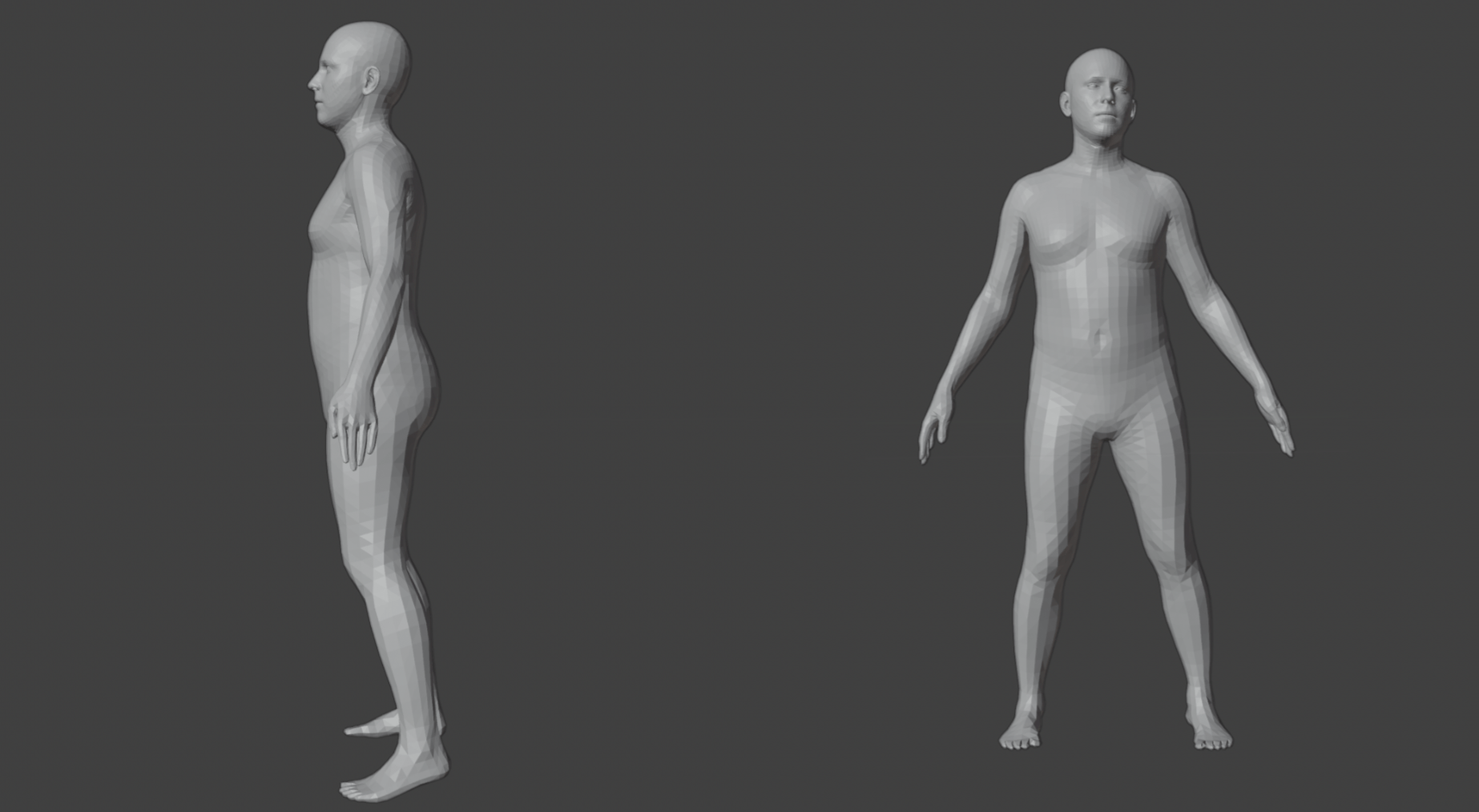} &
    \includegraphics[width=0.32\linewidth]{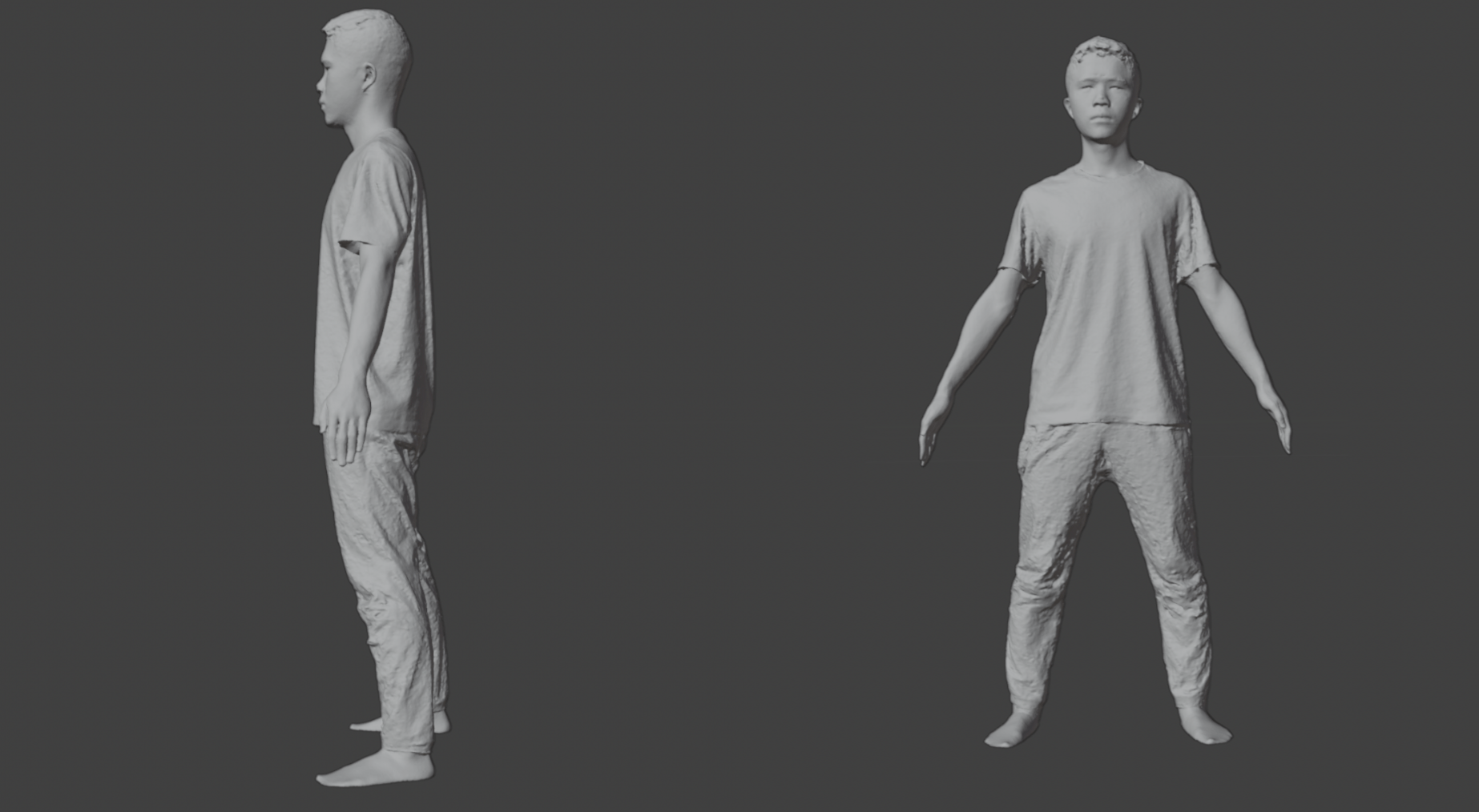} \\
    \includegraphics[width=0.32\linewidth]{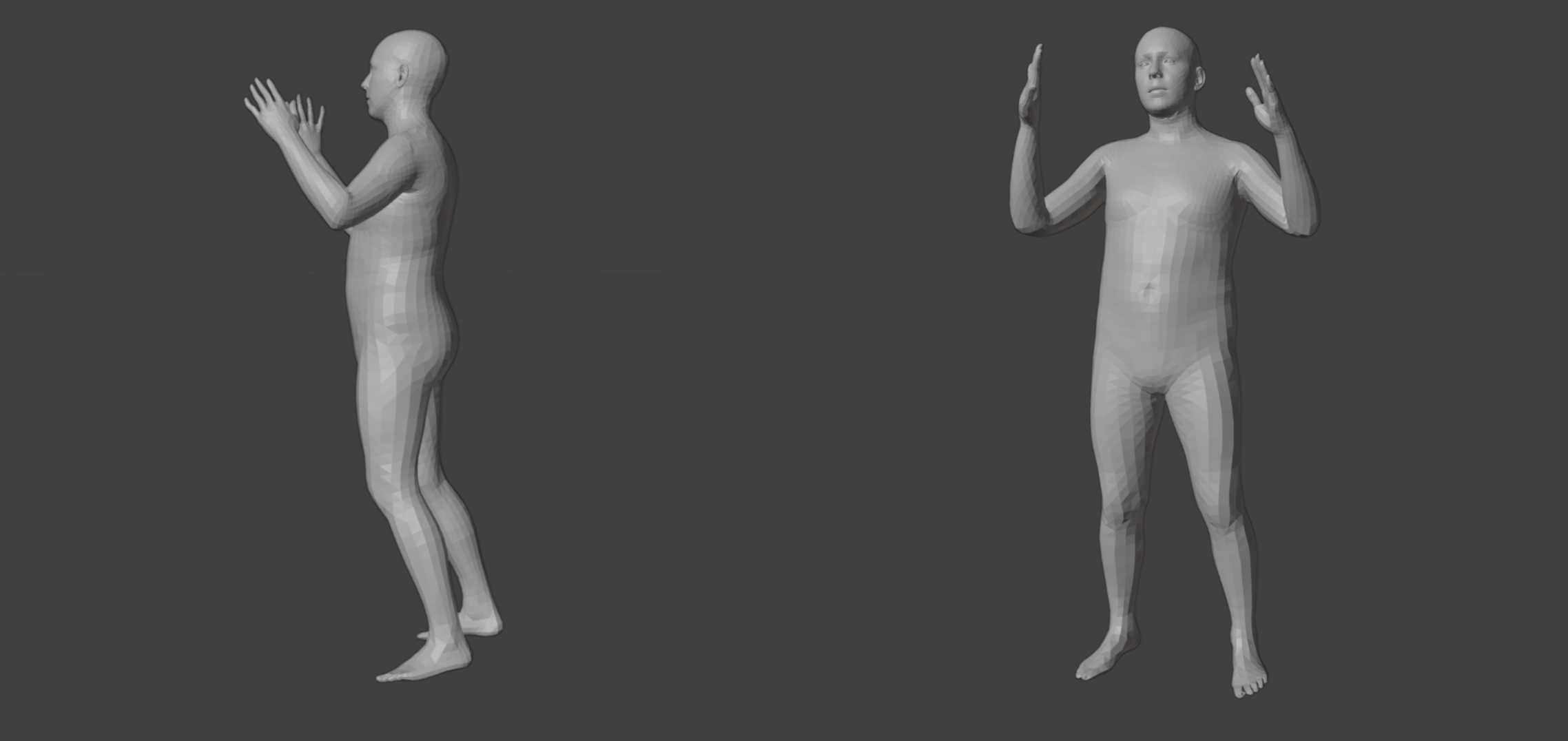} &
    \includegraphics[width=0.32\linewidth]{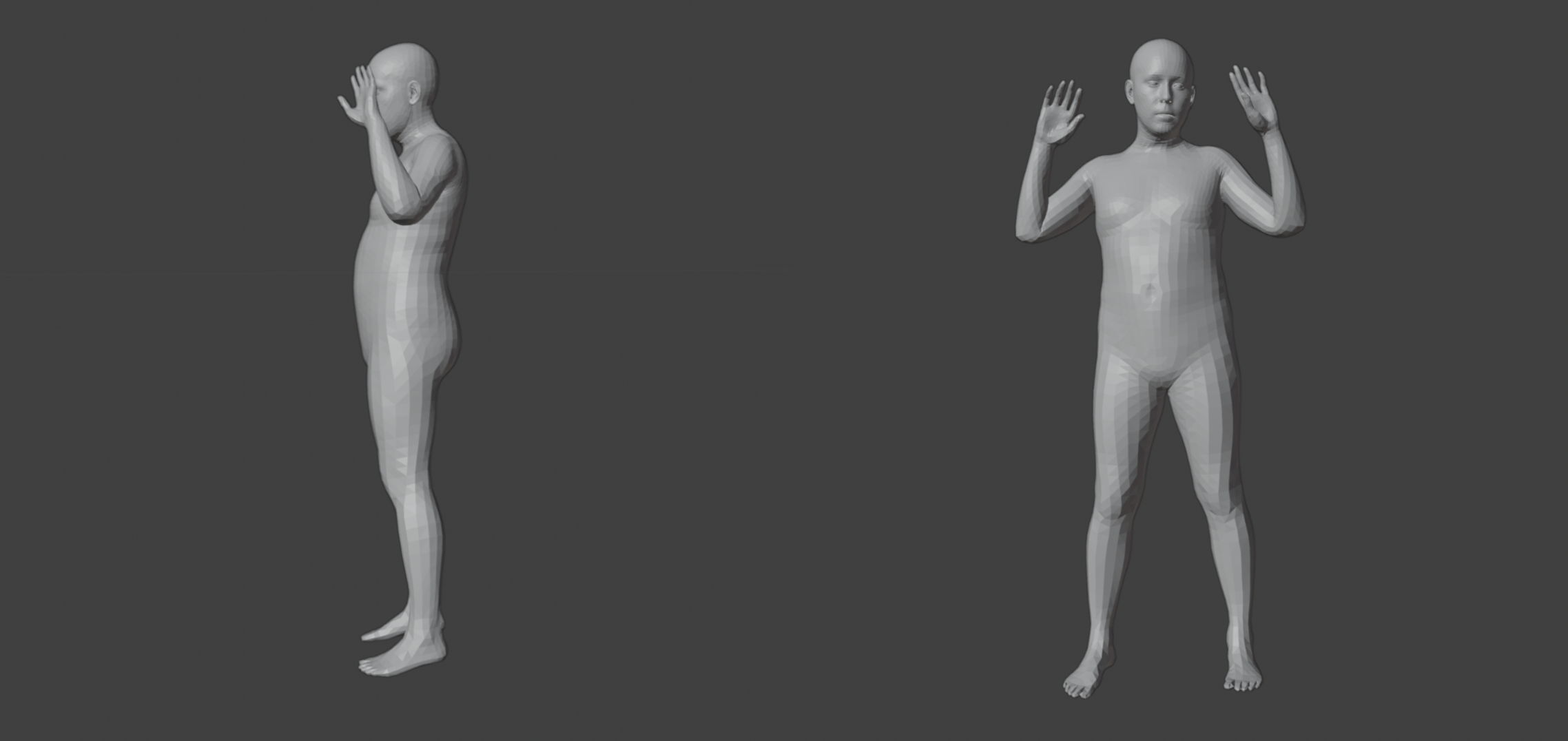} &
    \includegraphics[width=0.32\linewidth]{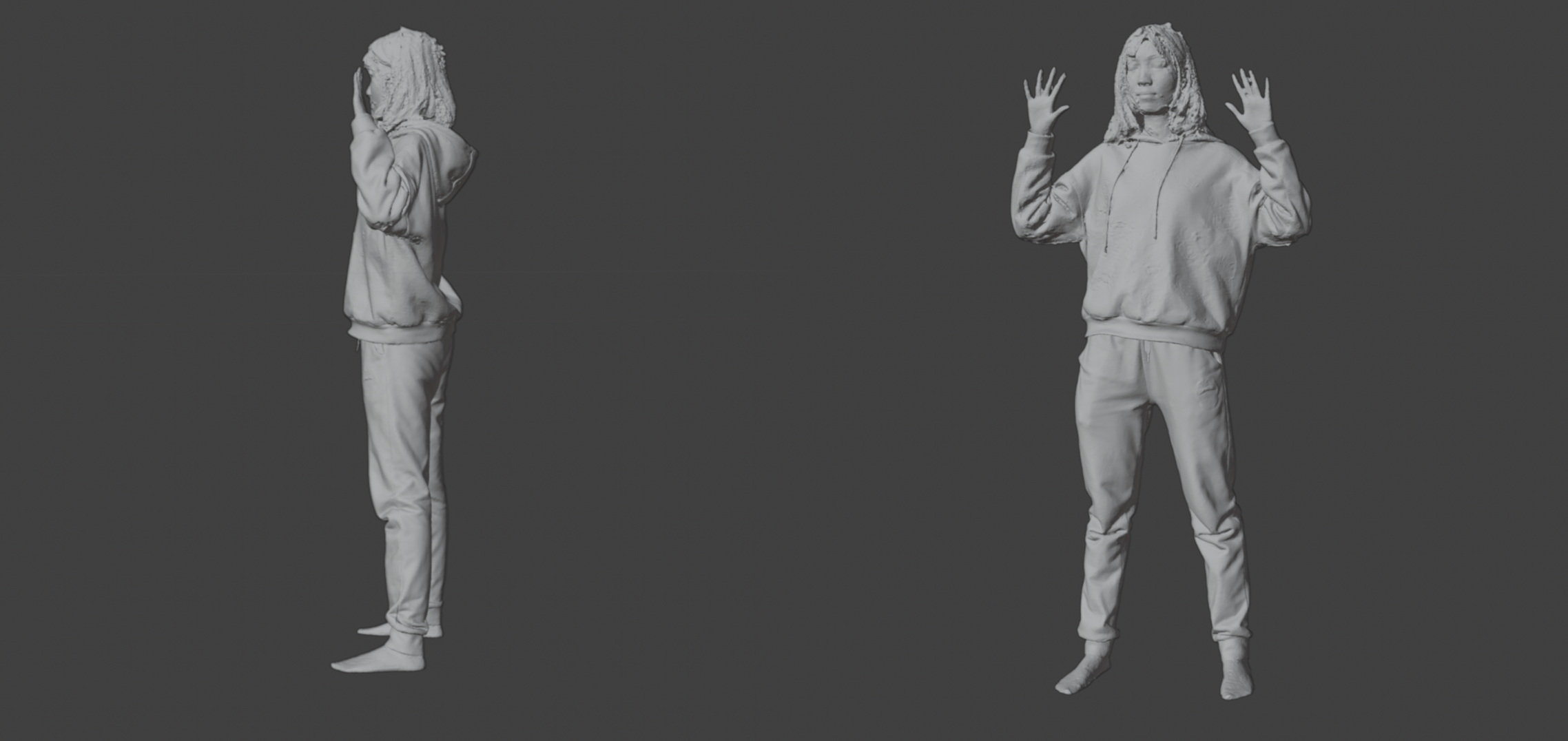}
    \end{tabular}
    \caption{JMBO refines the body shape and pose of the SMPL-X prior that guides the clothed avatar reconstruction. In contrast, single-view SMPL-X estimates, as used in ECON, suffer from reduced accuracy due to depth ambiguities and self-occlusions in unseen body regions.
    }
    \label{fig:jmbo_qualitative}
\end{figure*}

The qualitative results in Fig.~\ref{fig:jmbo_qualitative} visually corroborate the numerical improvements. Single-view SMPL-X predictions exhibit pose inaccuracies due to depth ambiguities (e.g., bent legs, overextended arms) and exaggerated body shapes (e.g., oversized torso caused by loose clothing in the bottom example). In contrast, the JMBO prediction exhibits a stable, upright posture with body proportions closely aligned to the ground-truth avatar.

\begin{table}[t]
  \centering
  \label{tab:final_geometry_metrics}
  { \small
  \begin{tabular}{@{\hskip 0.02cm}l@{\hskip 0.01cm}c@{\hskip 0.1cm}c@{\hskip 0.1cm}c@{\hskip 0.1cm}c@{\hskip 0.1cm}c}
    \cmidrule(lr){2-5}
     & \makecell{Silhouette\\Loss $\downarrow$} &
           \makecell{Landmark\\loss $\downarrow$} &
           \makecell{Normal\\loss $\downarrow$} &
           \makecell{Total\\loss $\downarrow$} \\
    \midrule
    {\small Front-view SMPL-X} & 0.103 & 1.259 & 0.013 & 0.229 \\
    {\small Multi-view avg. SMPL-X} & 0.092 & 0.169 & 0.012 & 0.113 \\
    {\small JMBO SMPL-X} & \textbf{0.041} & \textbf{0.153} & \textbf{0.008} & \textbf{0.059} \\  
    \bottomrule
  \end{tabular}
  }
  \caption{JMBO ablation study. The proposed JMBO algorithm improves the alignment between the output SMPL-X body model and the image silhouettes, and yields lower values across the loss terms in~\eqref{eq:jmbo}. Metrics are computed per image across the 8-view set of each avatar and averaged over the 20-avatar test set.}
  \label{tab:jmbo_metrics}
\end{table}

\begin{table}[t]
  \centering
  { \small
  \begin{tabular}{@{\hskip 0.02cm}l@{\hskip 0.2cm}c@{\hskip 0.2cm}c@{\hskip 0.2cm}c@{\hskip 0.2cm}c@{\hskip 0.2cm}c}
    \cmidrule(lr){2-5}
     & \makecell{Chamfer\\(mm) $\downarrow$} &
           \makecell{P2S Acc.\\\ (mm) $\downarrow$} &
           \makecell{P2S Comp.\\\ (mm) $\downarrow$} &
           \makecell{Normal\\(deg.) $\downarrow$} \\
    \midrule
    ECON \cite{xiu2023econ} & 40.56 & 39.52 & 37.21 & 45.79 \\
    2-view MExECON & 32.76 & 29.72 & 31.18 & 44.03 \\
    8-view MExECON & \textbf{23.12} & \textbf{21.26} & 22.84 & \textbf{39.54} \\ 
    8-view VGGT \cite{wang2025vggt} & 23.87 & 22.21 & \textbf{19.84} & 41.38 \\
    \bottomrule
  \end{tabular}
  }
  \caption{Mesh geometry comparison of ECON, MExECON for 2 and 8 input images, and VGGT. Results are averaged over the 20-avatar test set.
  }
  \label{tab:geometry_metrics}
\end{table}

\noindent
\textbf{Final clothed mesh accuracy.} The benefits of the multi-view JMBO SMPL-X prior propagate to the final clothed avatar, which also incorporates fine-grained details from the front and back normal maps. We compared the original ECON pipeline with the proposed 8-view MExECON, a 2-view variant of our approach using only front and back views, and the state-of-the-art multi-view 3D reconstruction method VGGT~\cite{wang2025vggt}.
VGGT is a large-scale transformer architecture that predicts both the camera parameters of the input views and a dense point cloud of the scene geometry, from which the avatar mesh is extracted using PSR~\cite{kazhdan2006poisson}.

As shown in Table~\ref{tab:geometry_metrics}, MExECON significantly outperforms ECON across all metrics. In particular, the \mbox{8-view} MExECON reduces the Chamfer Distance by 42\% (from 40.56 to 23.12 mm), indicating improved overall geometric alignment. P2S Accuracy and Completeness are reduced by 46\% and 38\%, respectively, demonstrating enhanced precision and surface coverage. This is particularly evident in the reconstruction of originally occluded regions like the back.
Finally, Normal Consistency improves by 13\%, indicating an improvement of fine-grained surface details.

The 2-view MExECON variant achieves intermediate performance between the single-view ECON baseline and the 8-view counterpart. This highlights the importance of incorporating at least one additional viewpoint to resolve ambiguities. Increasing the input viewpoints to 8 further boosts performance, demonstrating the ability of our method to scale effectively with the number of viewpoints.

\begin{figure}
\centering
    \begin{tabular}{@{\hskip -0cm}c@{\hskip 0.1cm}c@{\hskip 0.1cm}c}
        {\footnotesize 8-view VGGT~\cite{wang2025vggt}} & {\footnotesize 8-view MExECON (Ours)} & {\footnotesize Ground truth} \\
    \includegraphics[width=0.32\linewidth]{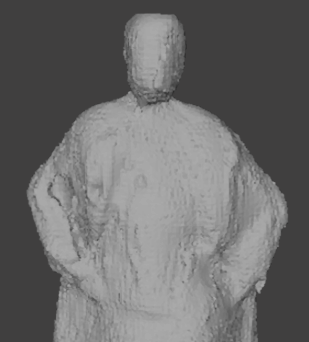} &
    \includegraphics[width=0.32\linewidth]{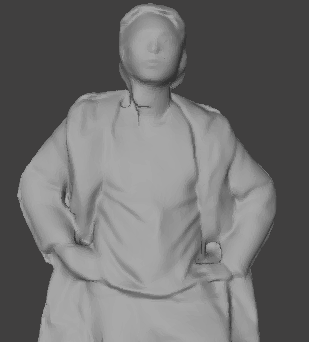} &
    \includegraphics[width=0.32\linewidth]{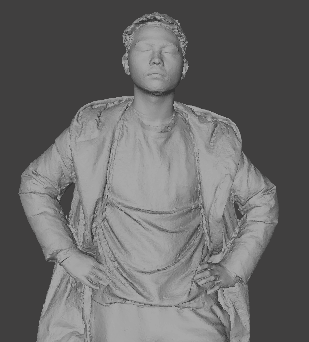}
    \end{tabular}
    \caption{VGGT~\cite{wang2025vggt} recovers accurate overall shape and surface coverage but smooths out fine details. Our 8-view MExECON results preserve high-frequency details while maintaining strong geometric alignment with the ground truth.}
    \label{fig:vggt_comparison}
\end{figure}

VGGT provides robust surface coverage, achieving the lowest P2S Completeness among all methods. In contrast, the surface completion step in ECON and MExECON can lead to incomplete limbs in complex poses, as discussed in Sec.~\ref{sec:limitations}. While VGGT avatars achieve accurate overall shape, they lack fine-grained details, resulting in higher Chamfer Distance and Normal Consistency errors. Fig.~\ref{fig:vggt_comparison} highlights the qualitative difference in terms of geometric details. Note that VGGT is trained on a large collection of real-world and synthetic datasets, and is not specifically tailored for human avatar reconstruction.

\begin{figure*}
\centering
    \begin{tabular}{@{\hskip -0cm}c@{\hskip 0.1cm}c@{\hskip 0.1cm}c@{\hskip 0.1cm}c@{\hskip 0.1cm}c@{\hskip 0.1cm}c}
     & RGB images & ECON & 2-view MExECON & 8-view MExECON & Ground truth \\
    \rotatebox[origin=lb]{90}{\footnotesize \hspace{1.5cm} Avatar ID 0004} &
\includegraphics[width=0.19\linewidth]{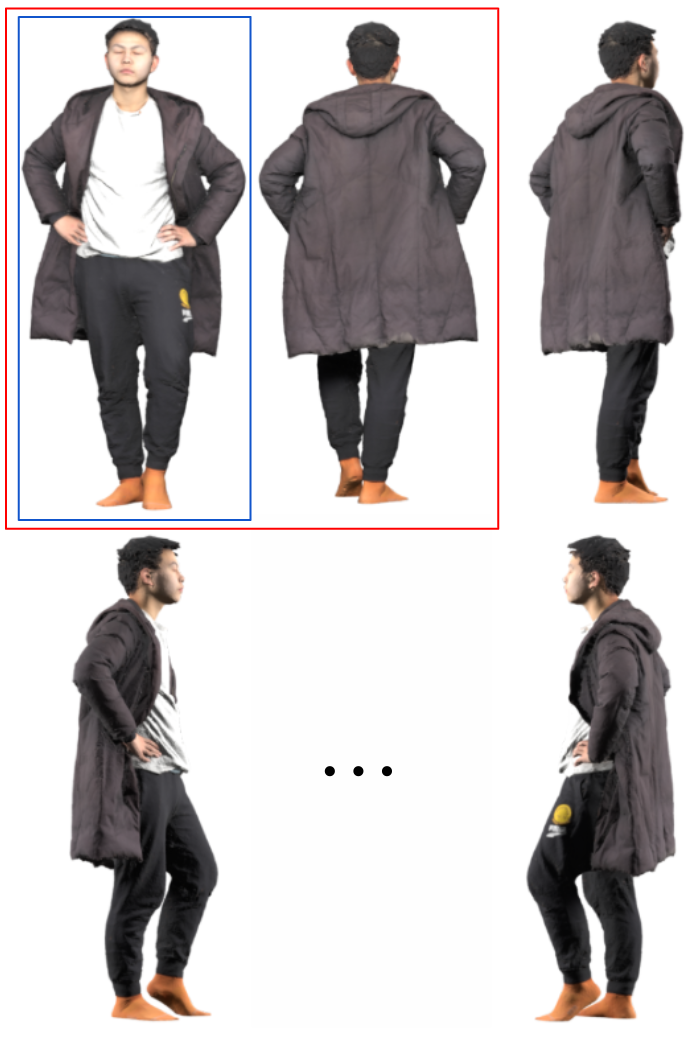} &
    \includegraphics[width=0.19\linewidth]{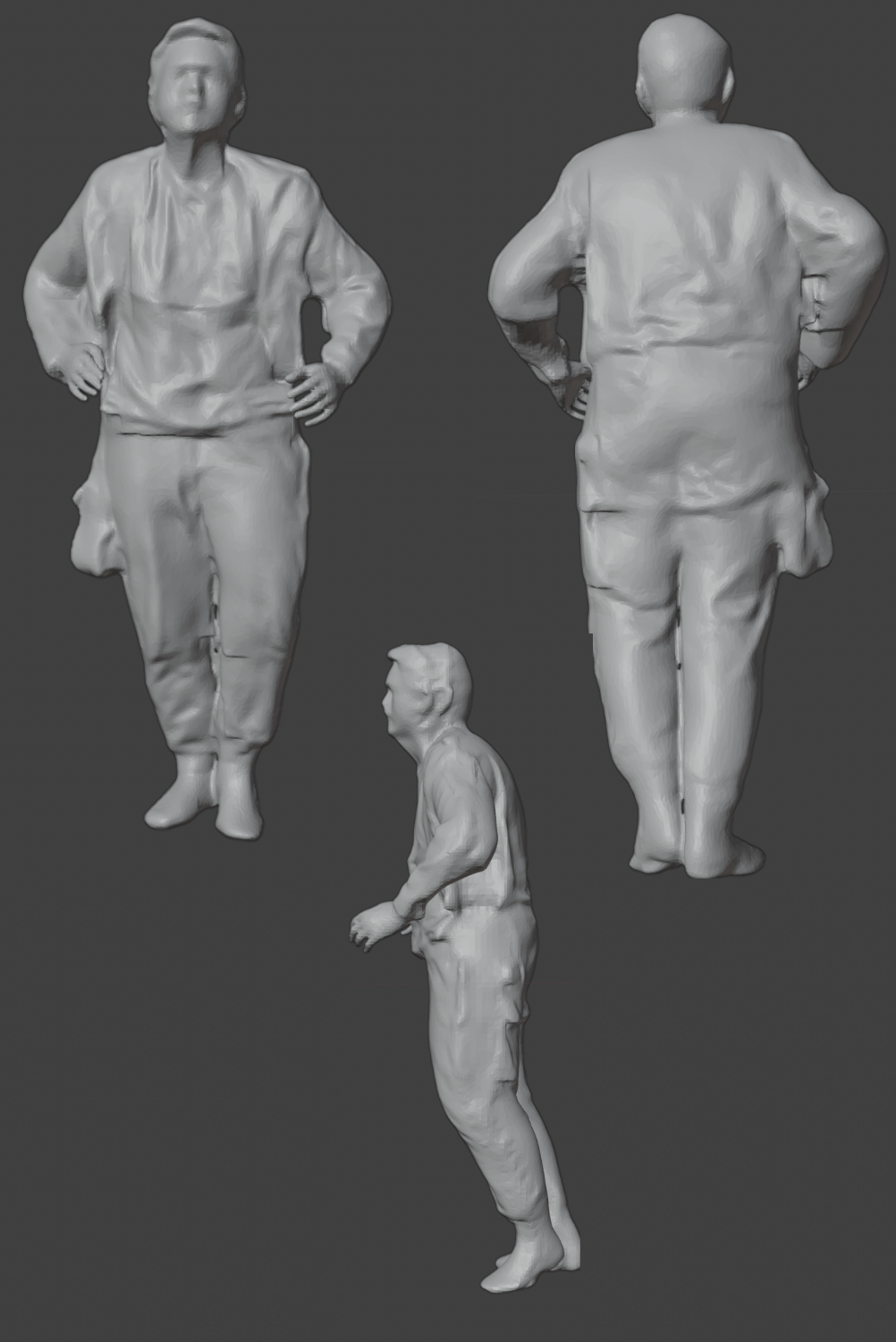} &
    \includegraphics[width=0.19\linewidth]{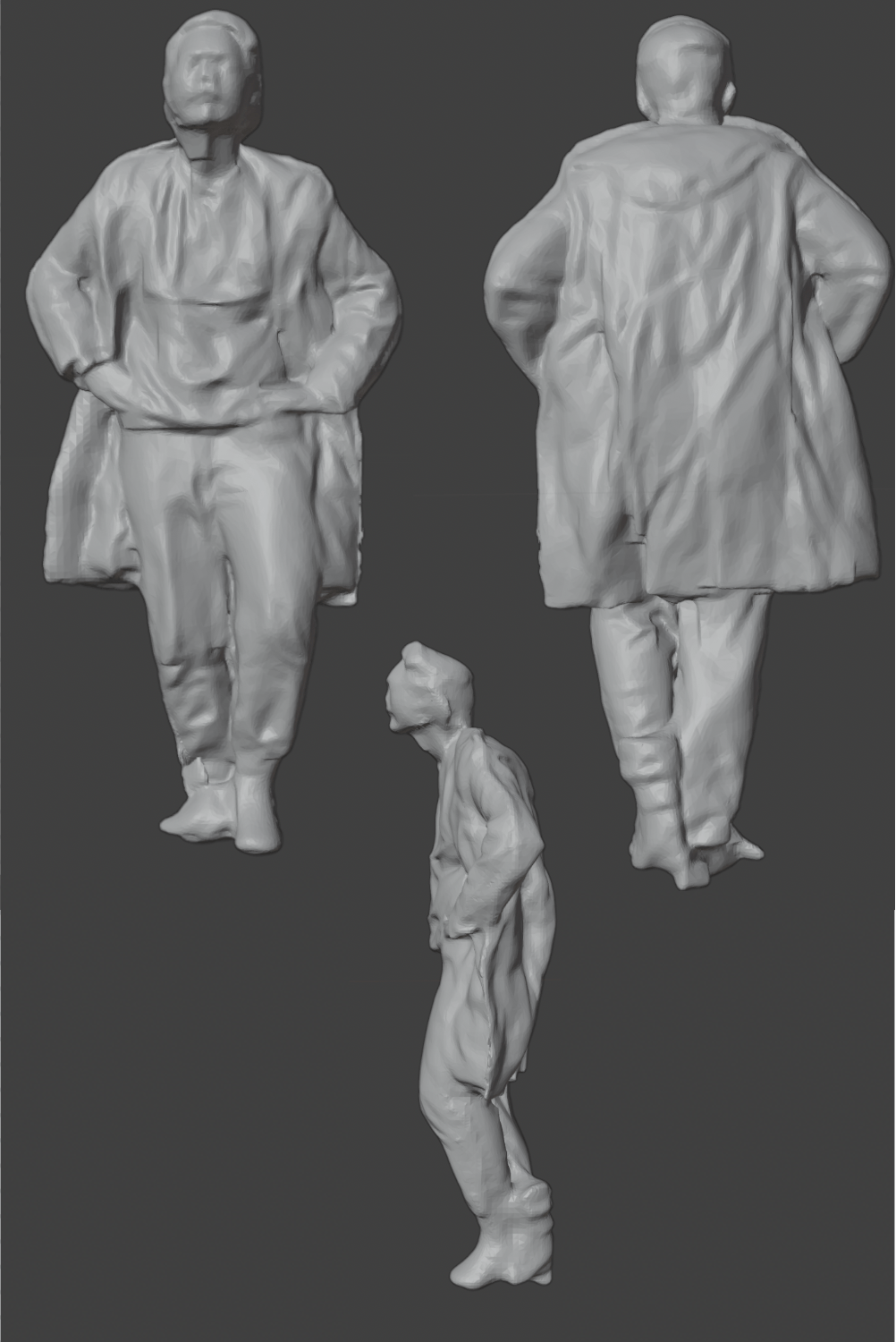} &
    \includegraphics[width=0.19\linewidth]{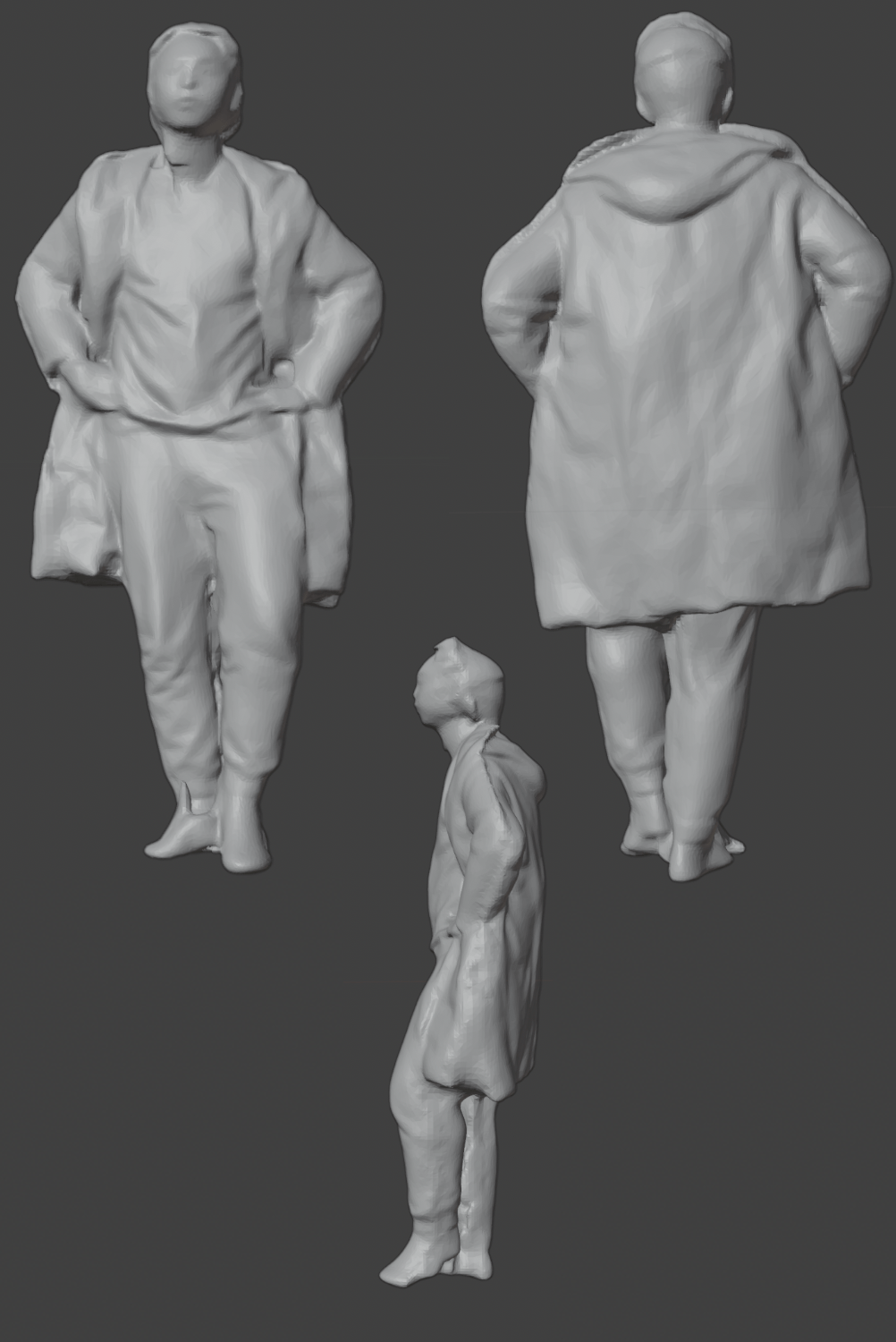} &
    \includegraphics[width=0.19\linewidth]{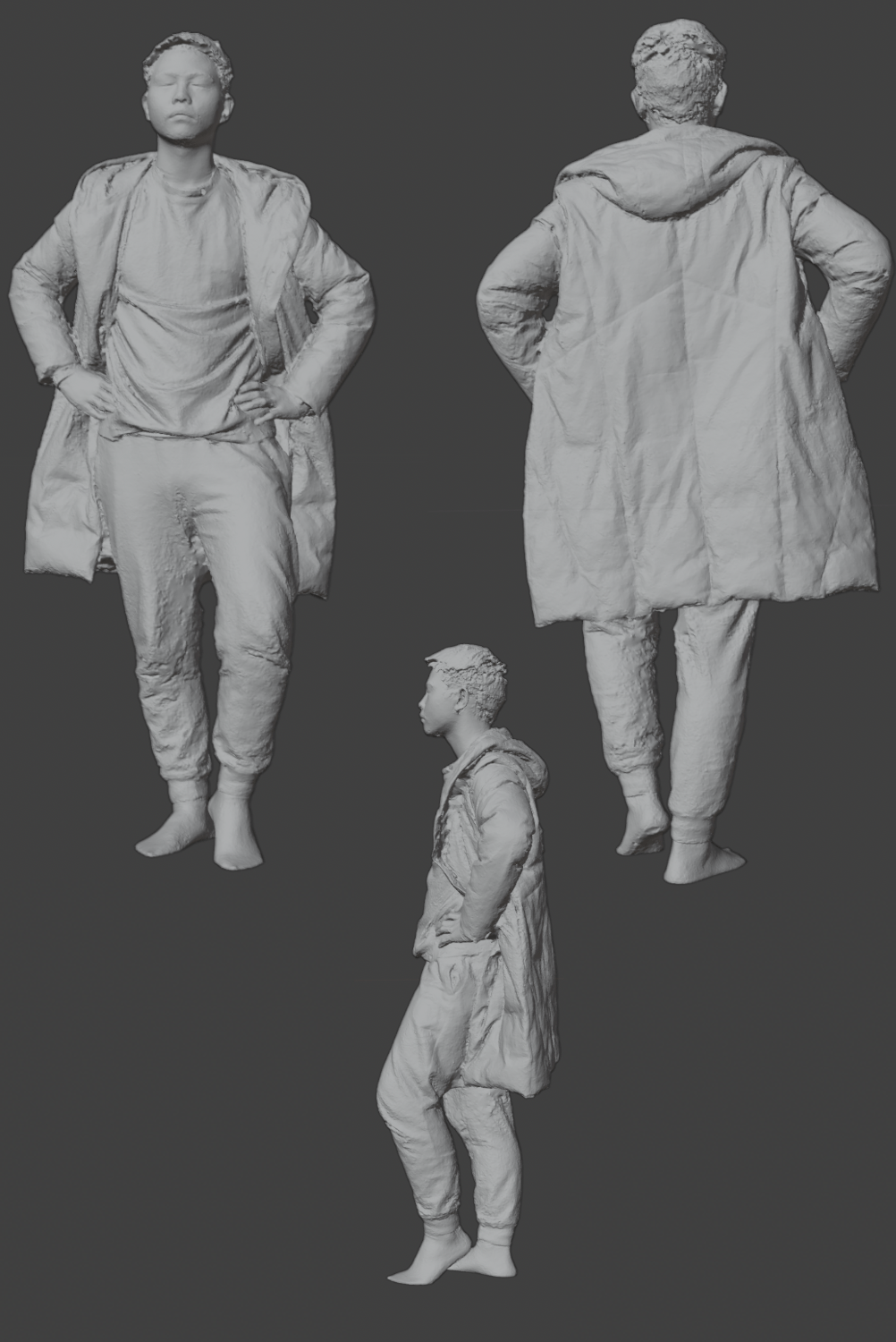} \\
\rotatebox[origin=lb]{90}{\footnotesize \hspace{1.5cm} Avatar ID 0021} &
\includegraphics[width=0.19\linewidth]{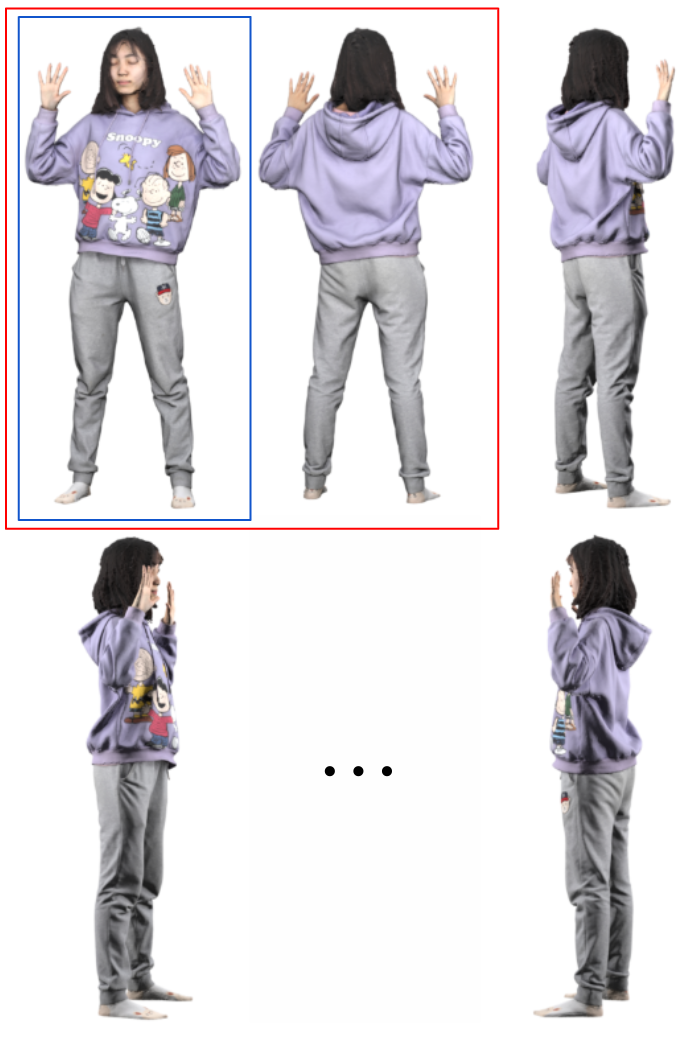} &
    \includegraphics[width=0.19\linewidth]{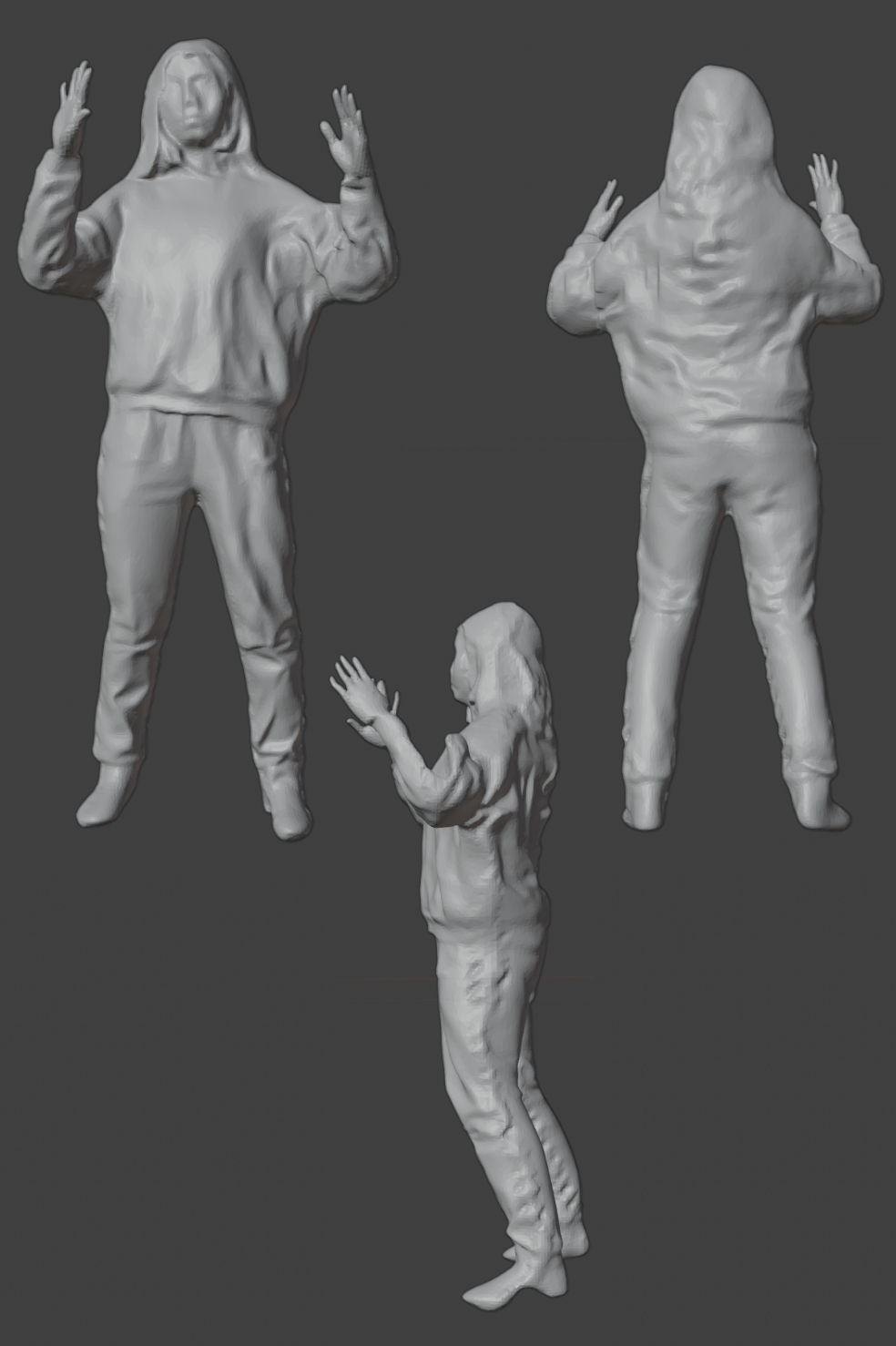} &
    \includegraphics[width=0.19\linewidth]{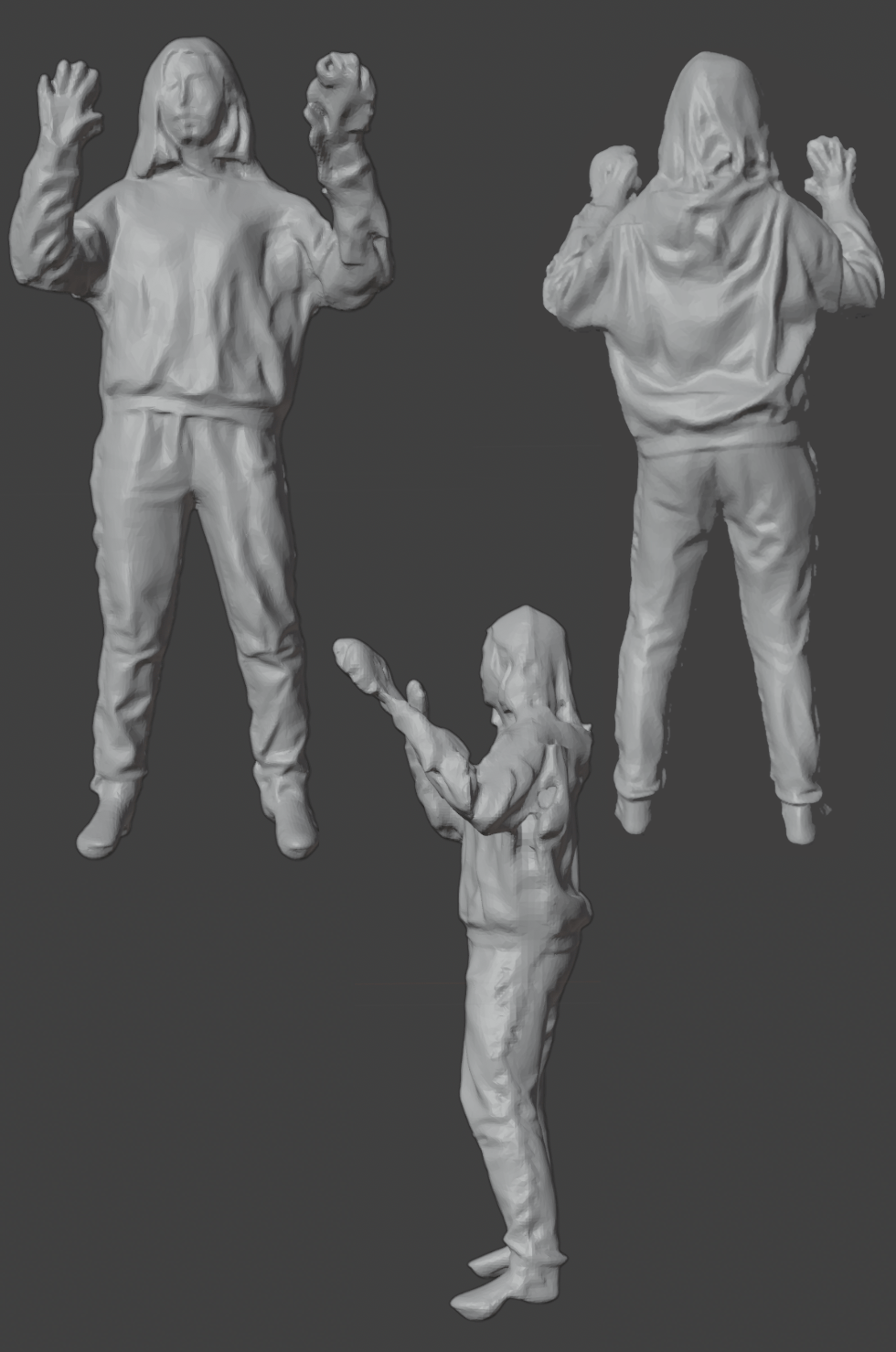} &
    \includegraphics[width=0.19\linewidth]{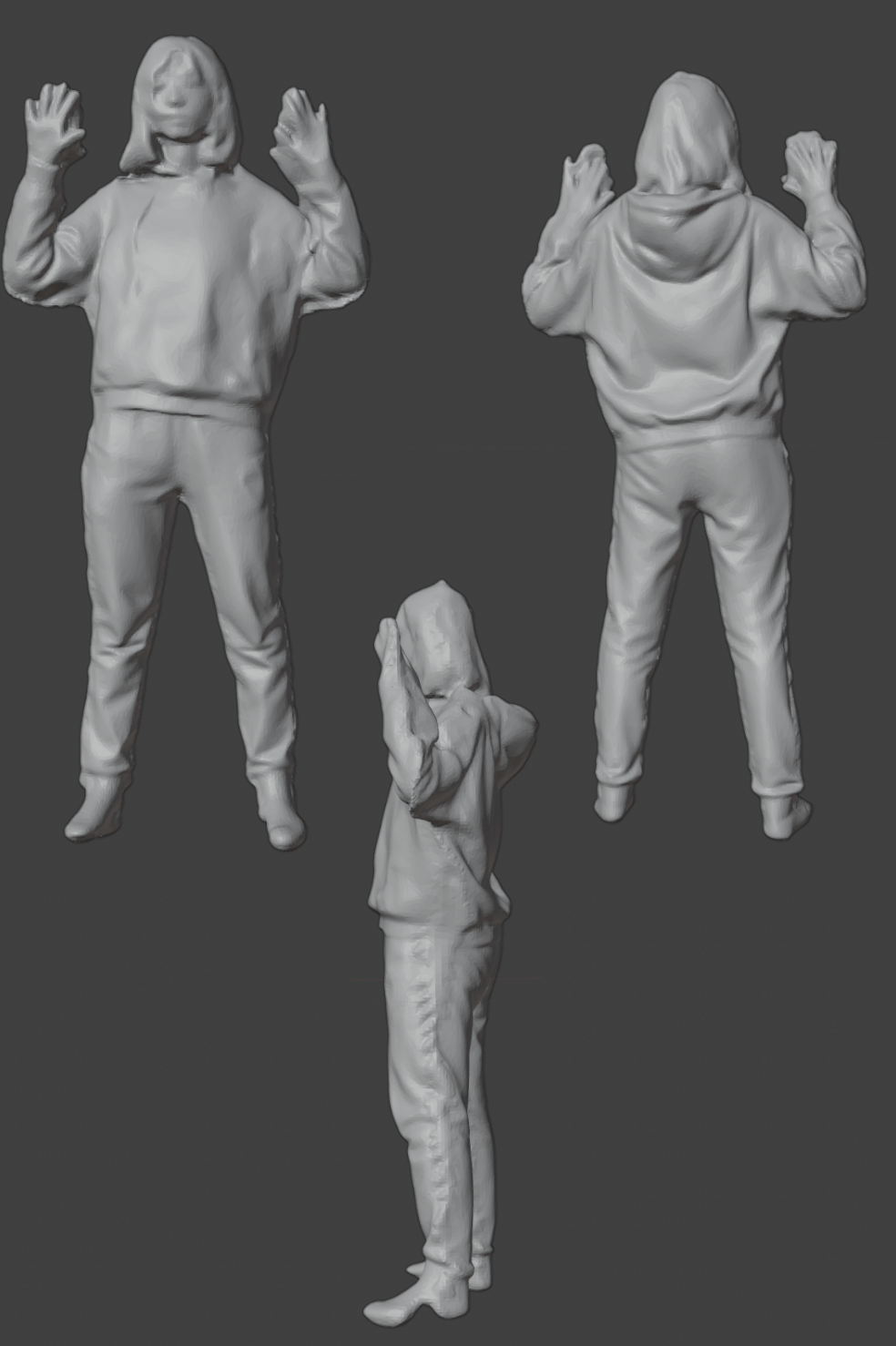} &
    \includegraphics[width=0.19\linewidth]{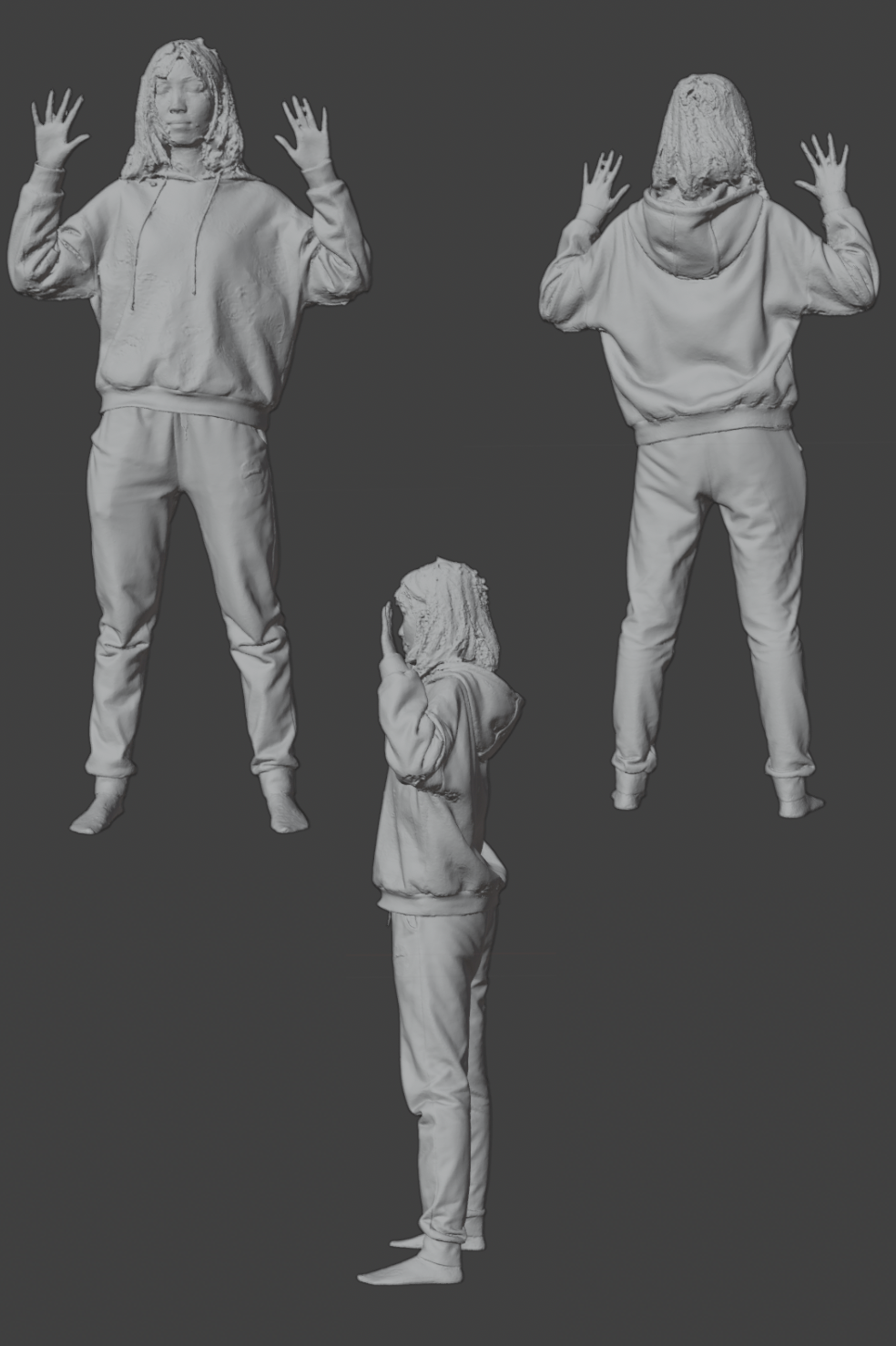} \\
\rotatebox[origin=lb]{90}{\footnotesize \hspace{1.5cm} Avatar ID 0070} &
\includegraphics[width=0.19\linewidth]{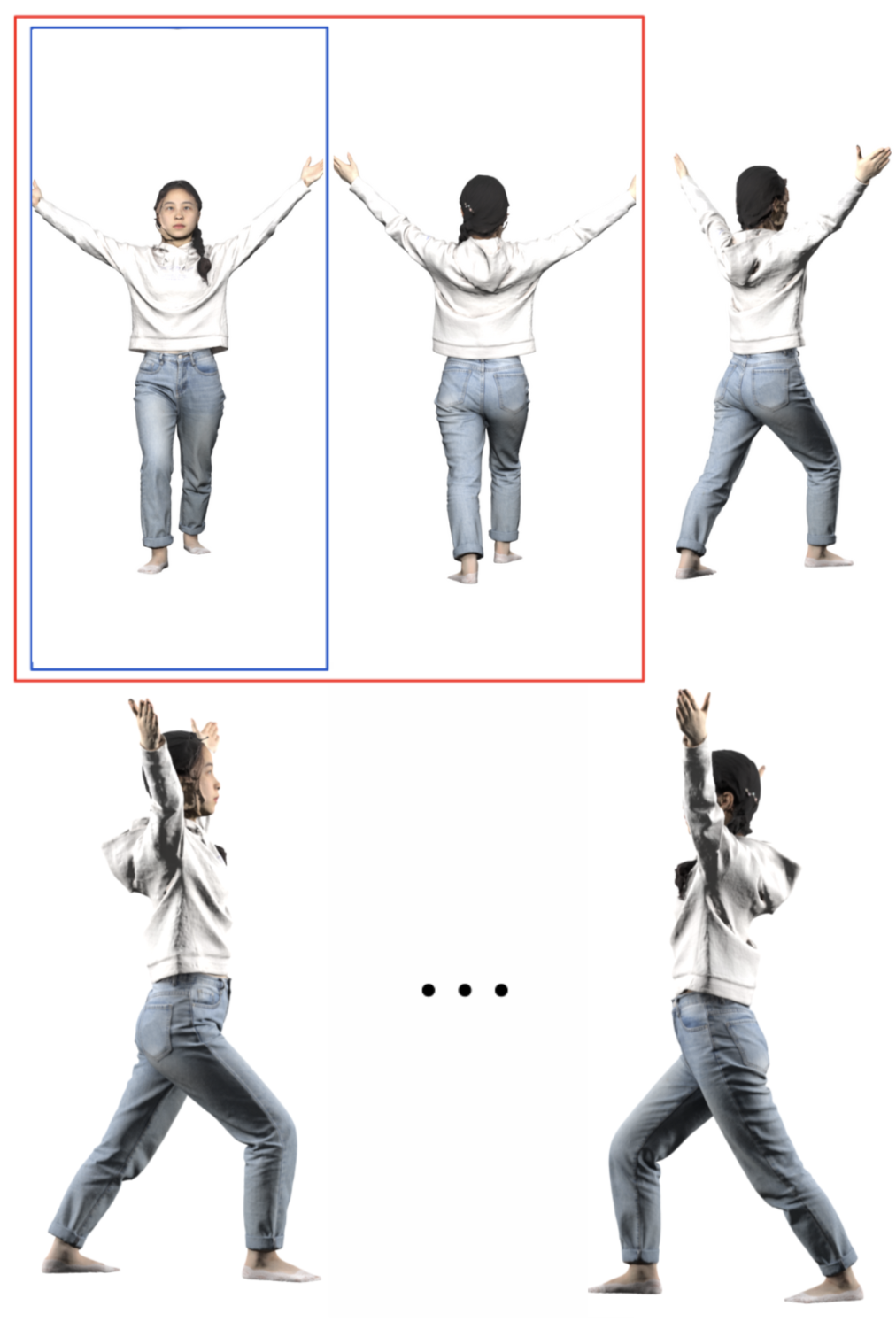} &
    \includegraphics[width=0.19\linewidth]{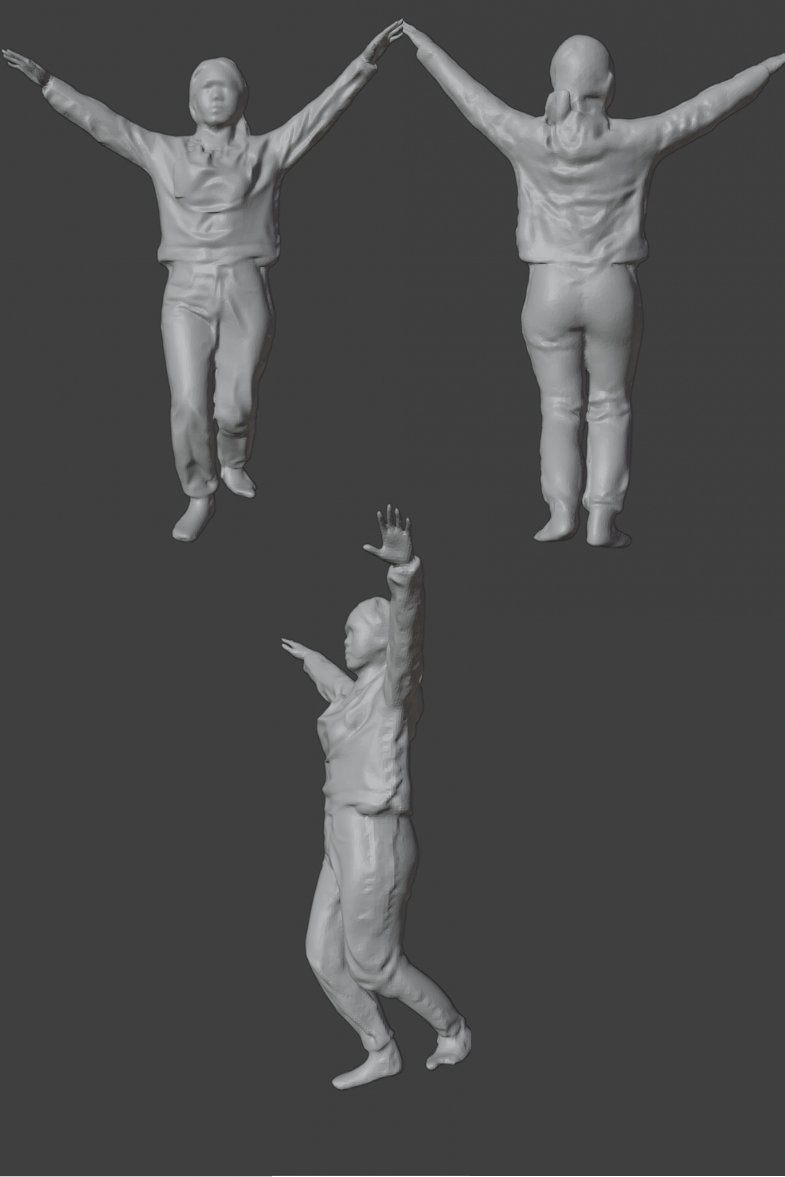} &
    \includegraphics[width=0.19\linewidth]{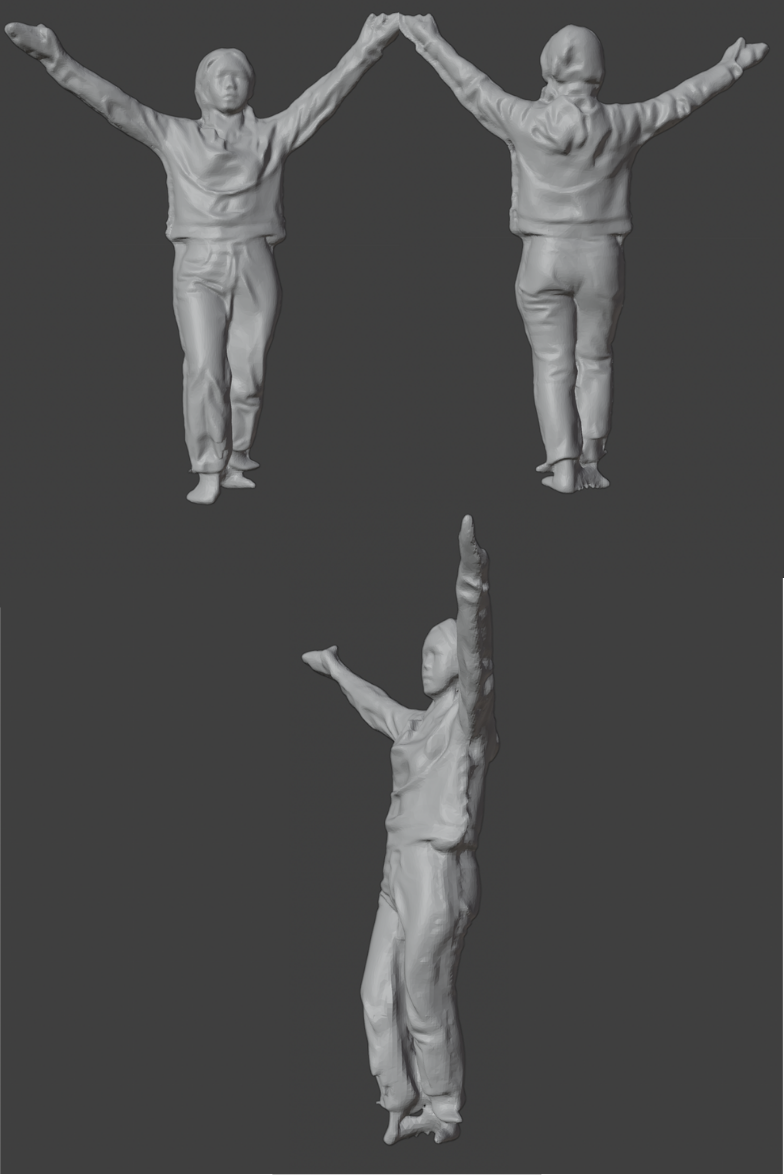} &
    \includegraphics[width=0.19\linewidth]{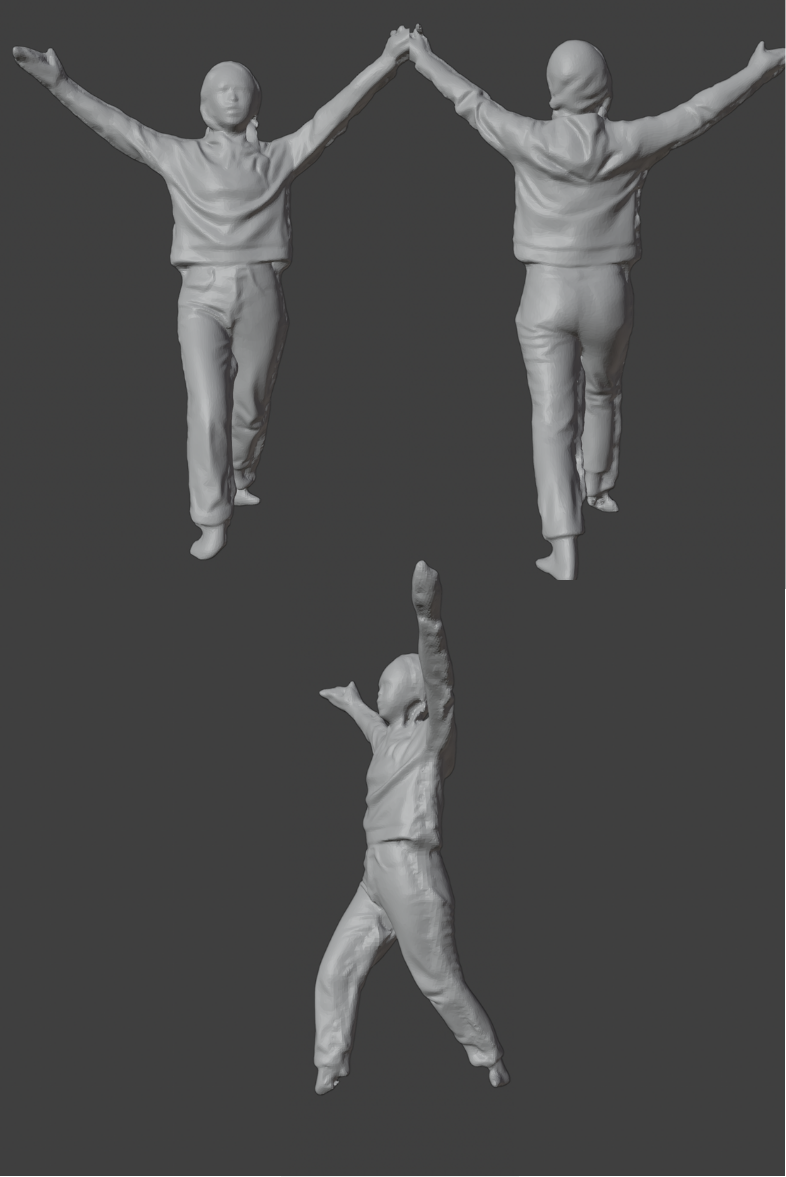} &
    \includegraphics[width=0.19\linewidth]{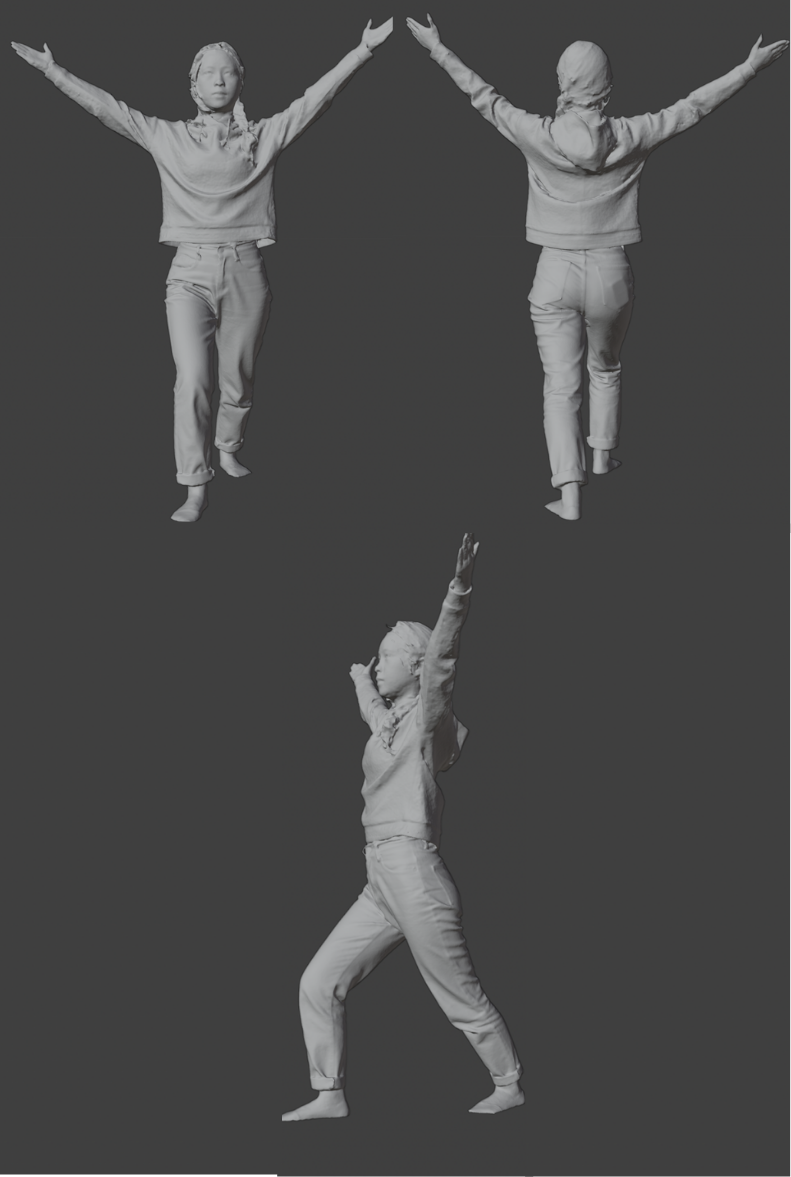} \\
\rotatebox[origin=lb]{90}{\footnotesize \hspace{1.5cm} Avatar ID 0445} &
\includegraphics[width=0.19\linewidth]{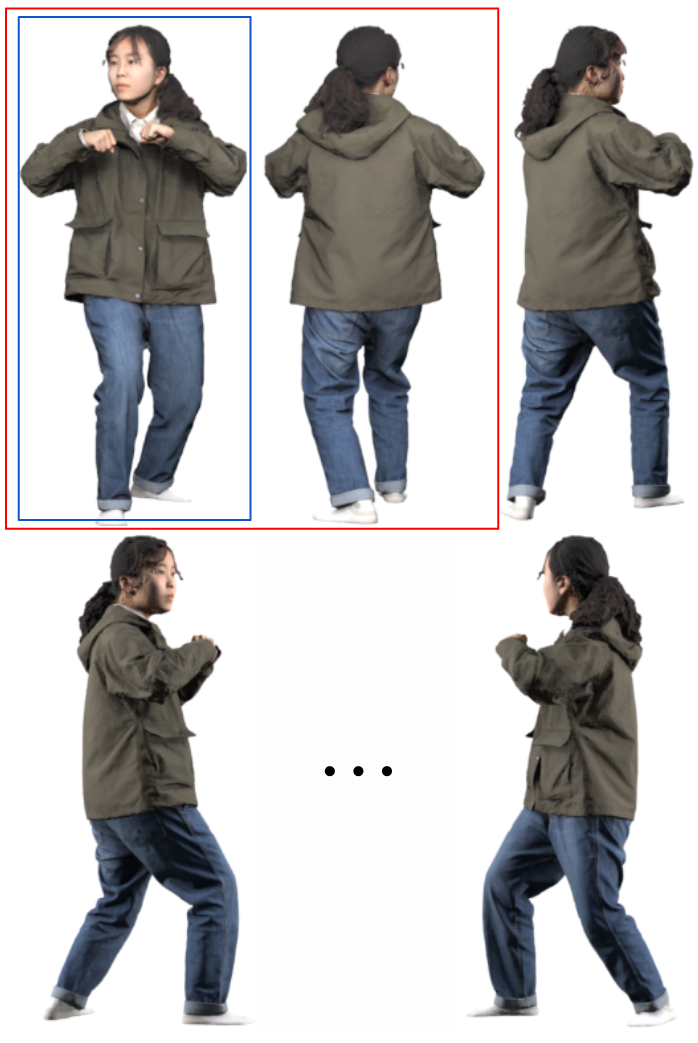} &
    \includegraphics[width=0.19\linewidth]{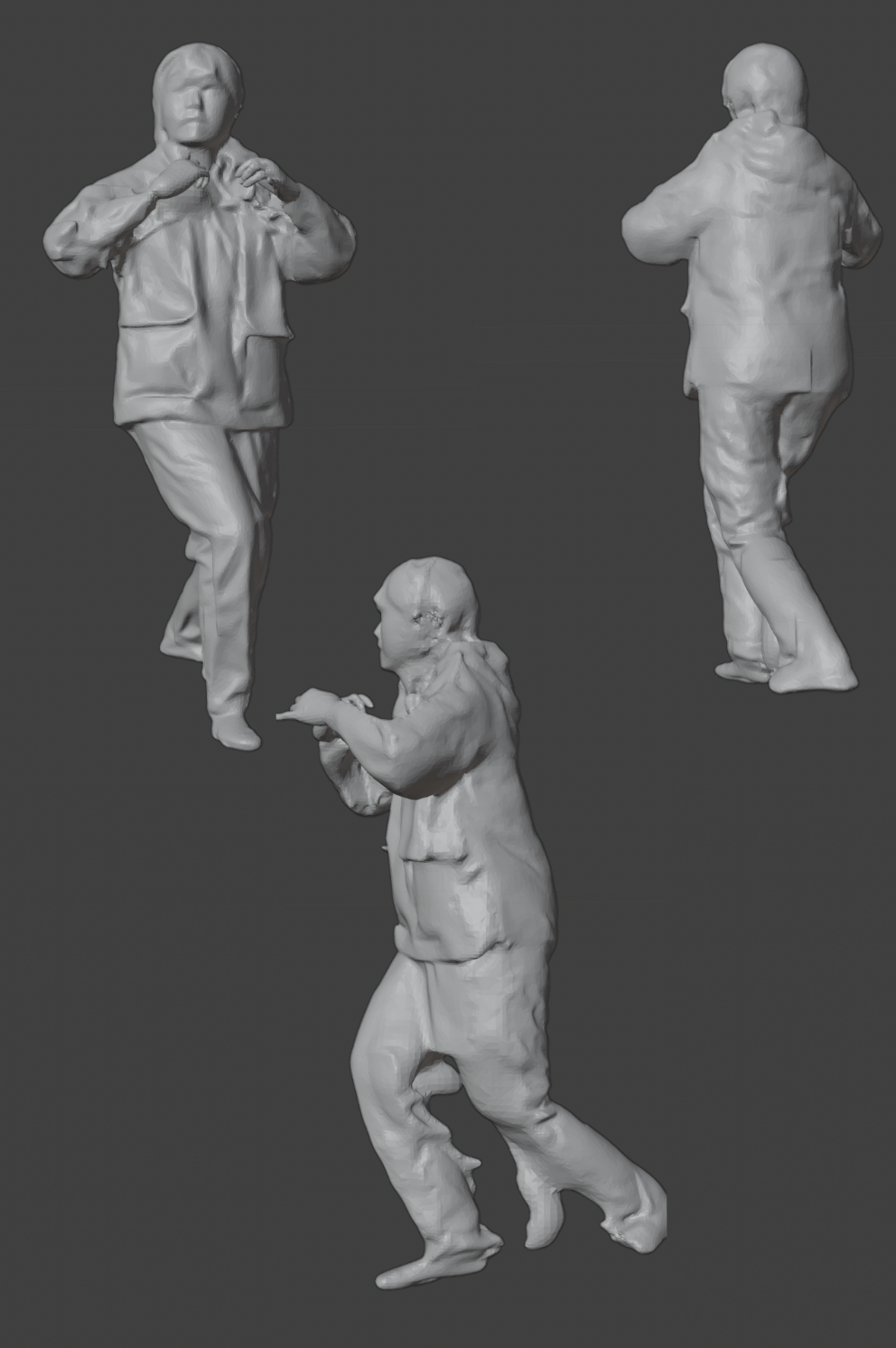} &
    \includegraphics[width=0.19\linewidth]{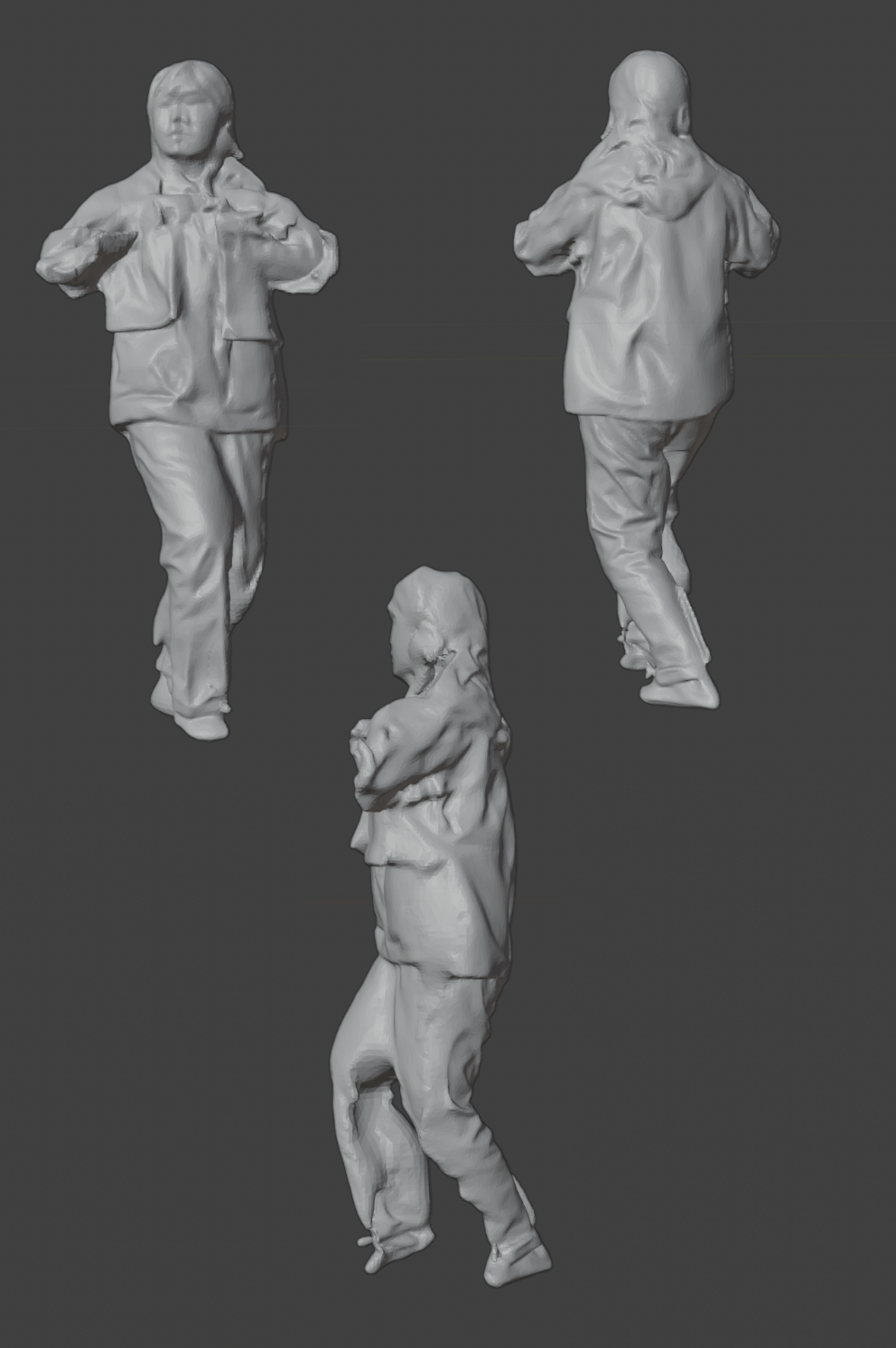} &
    \includegraphics[width=0.19\linewidth]{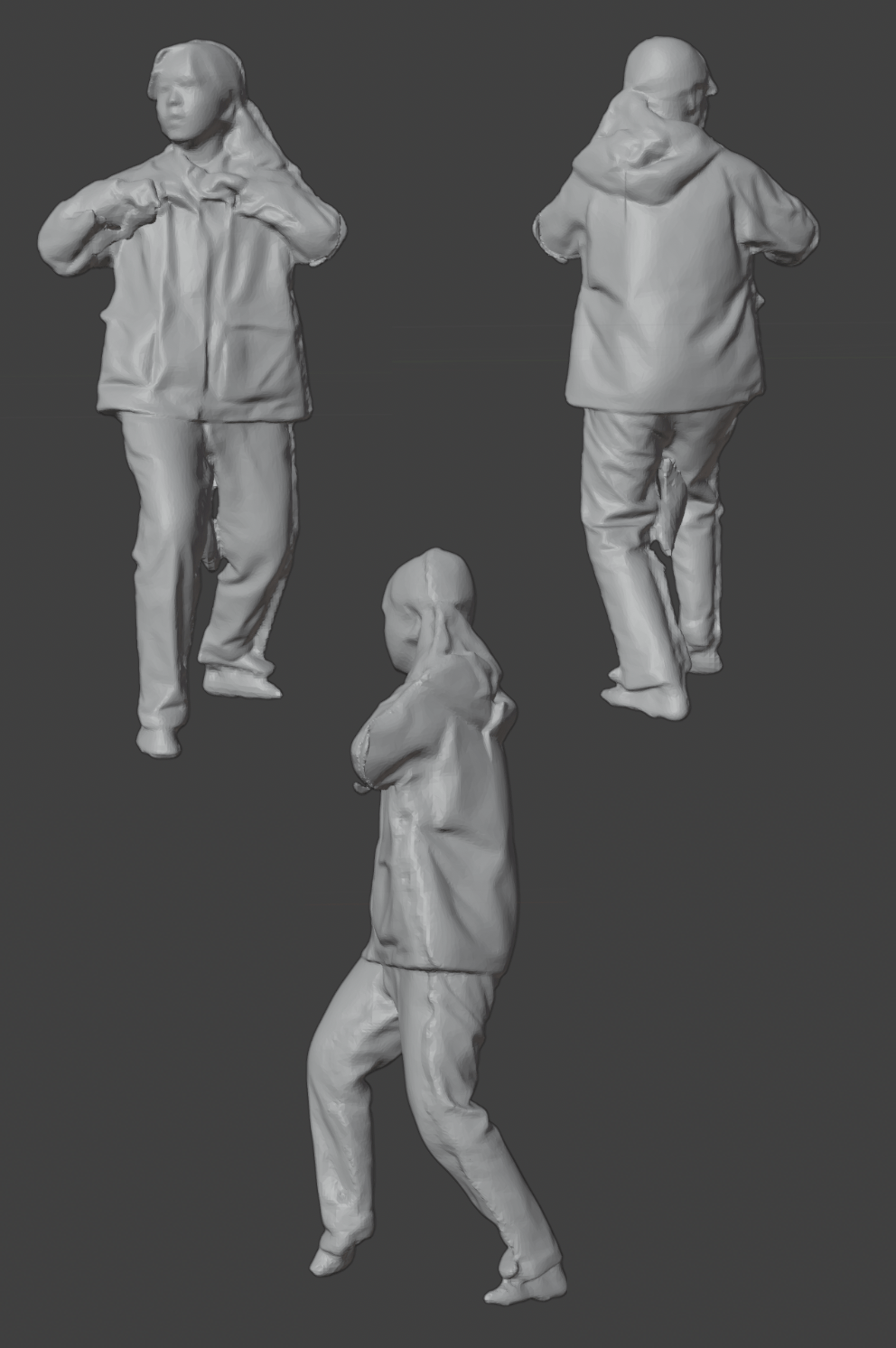} &
    \includegraphics[width=0.19\linewidth]{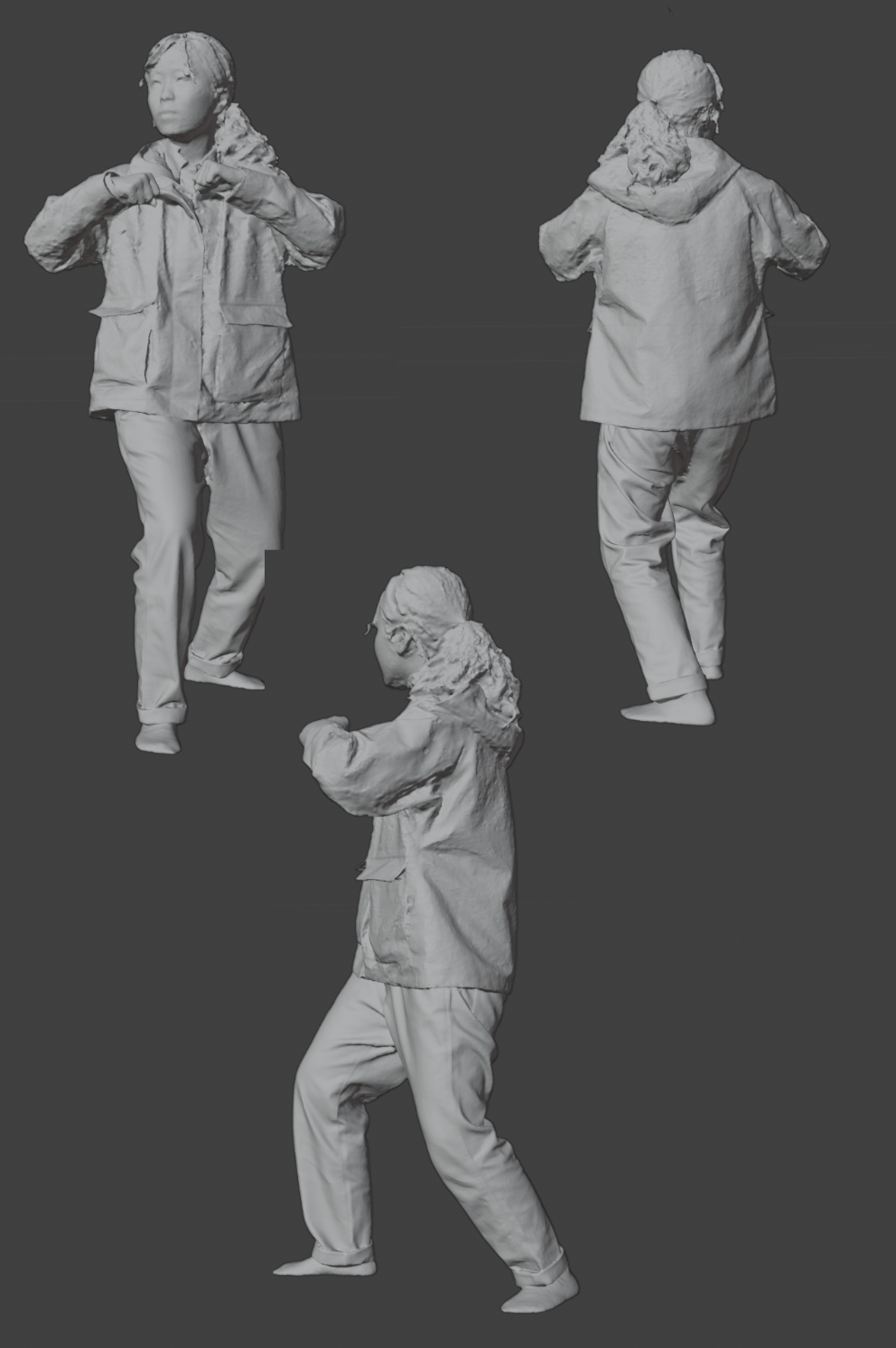} \\
    \end{tabular}
    \caption{Qualitative comparison. ECON takes a single front-view input (blue bounding boxes), 2-view MExECON uses front and back views (red bounding boxes), and 8-view MExECON uses all available views. MExECON consistently yields more accurate body shape and pose estimation, along with enhanced geometric detail (particularly in the back). All meshes are shown from 3 viewpoints for clarity. 
    }
    \label{fig:qualitative_results}
\end{figure*}

\begin{figure*}[t]
\centering
    \begin{tabular}{@{\hskip -0cm}c@{\hskip 0.1cm}c@{\hskip 0.1cm}c}
    {\small Avatar ID 0083} & {\small Avatar ID 0023} & {\small Avatar ID 0071} \\
    \includegraphics[width=0.32\linewidth]{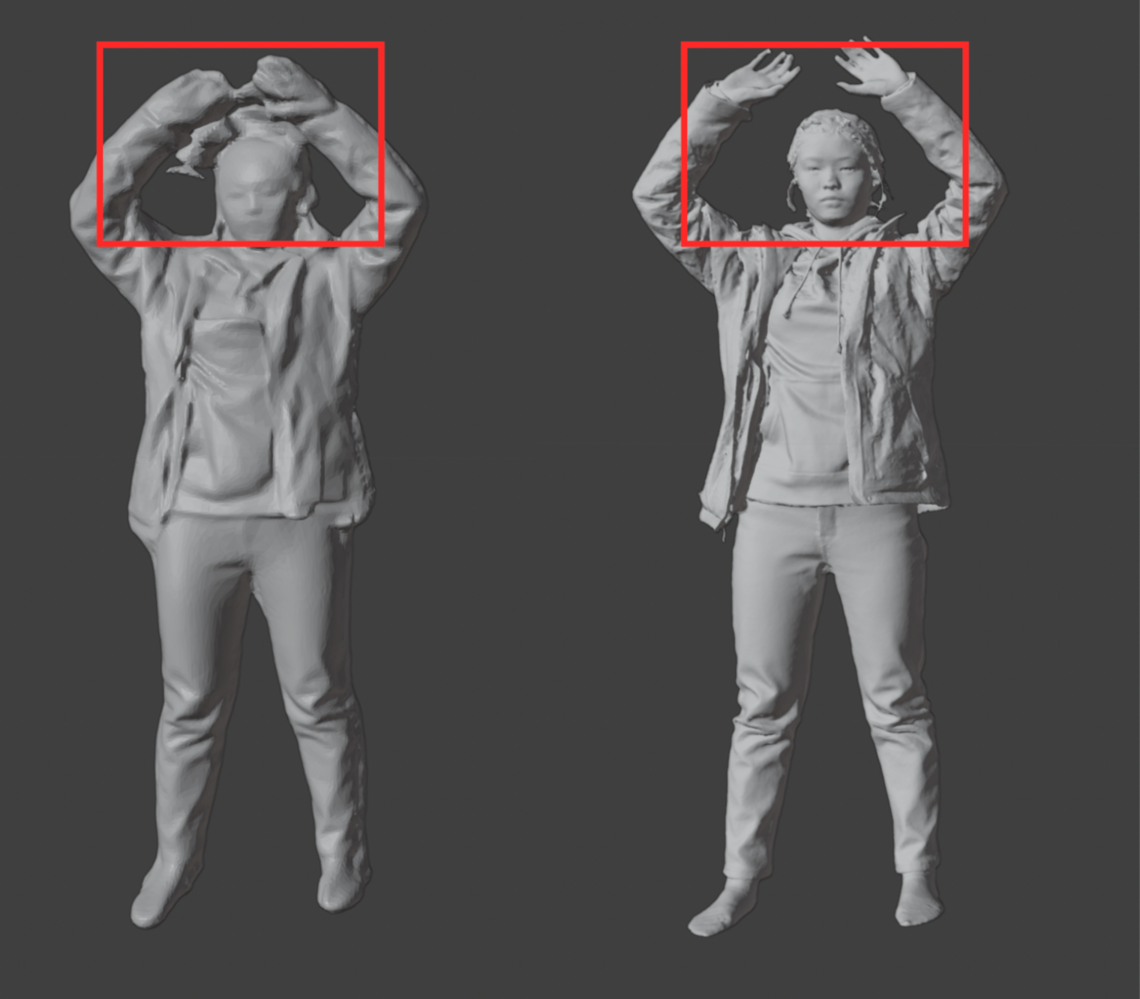} &
    \includegraphics[width=0.32\linewidth]{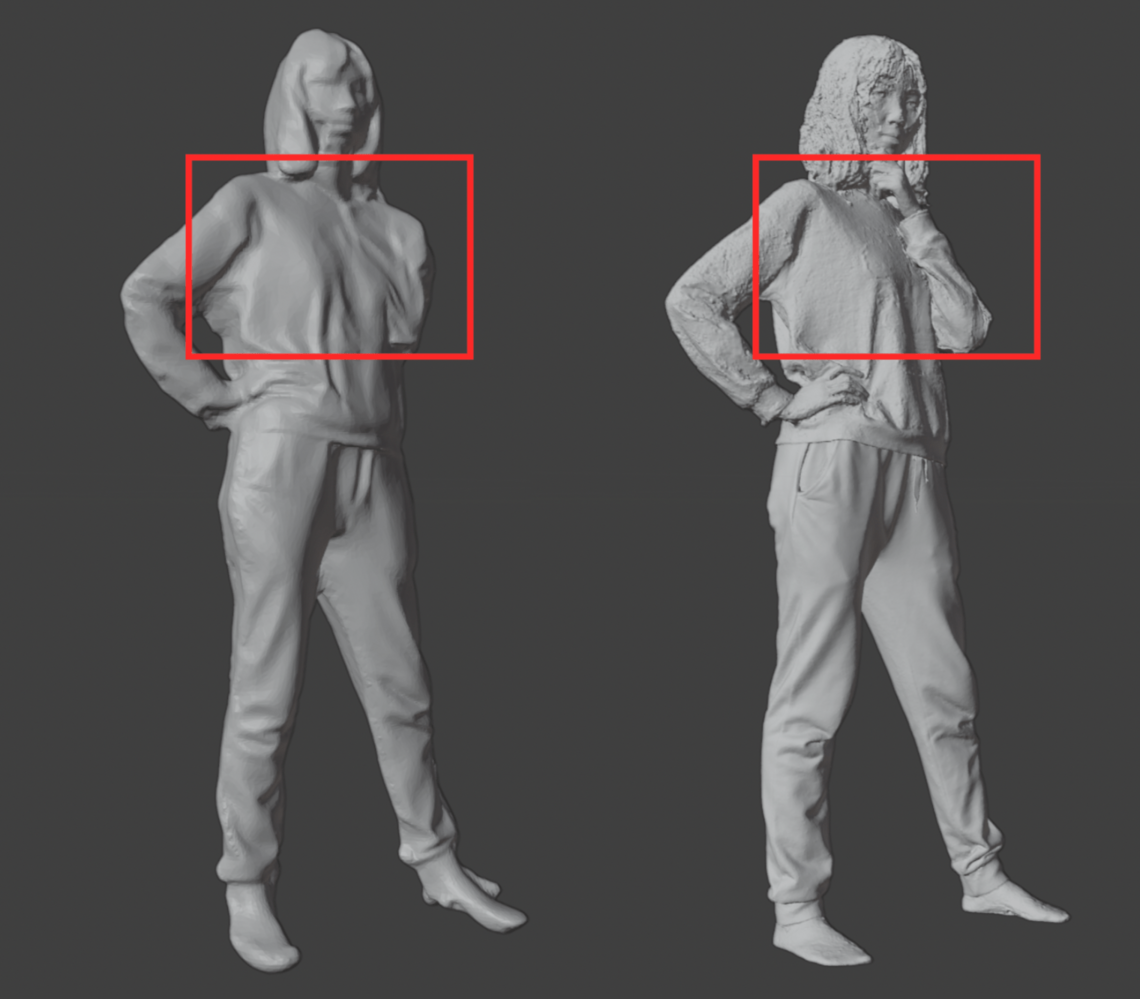} &
    \includegraphics[width=0.32\linewidth]{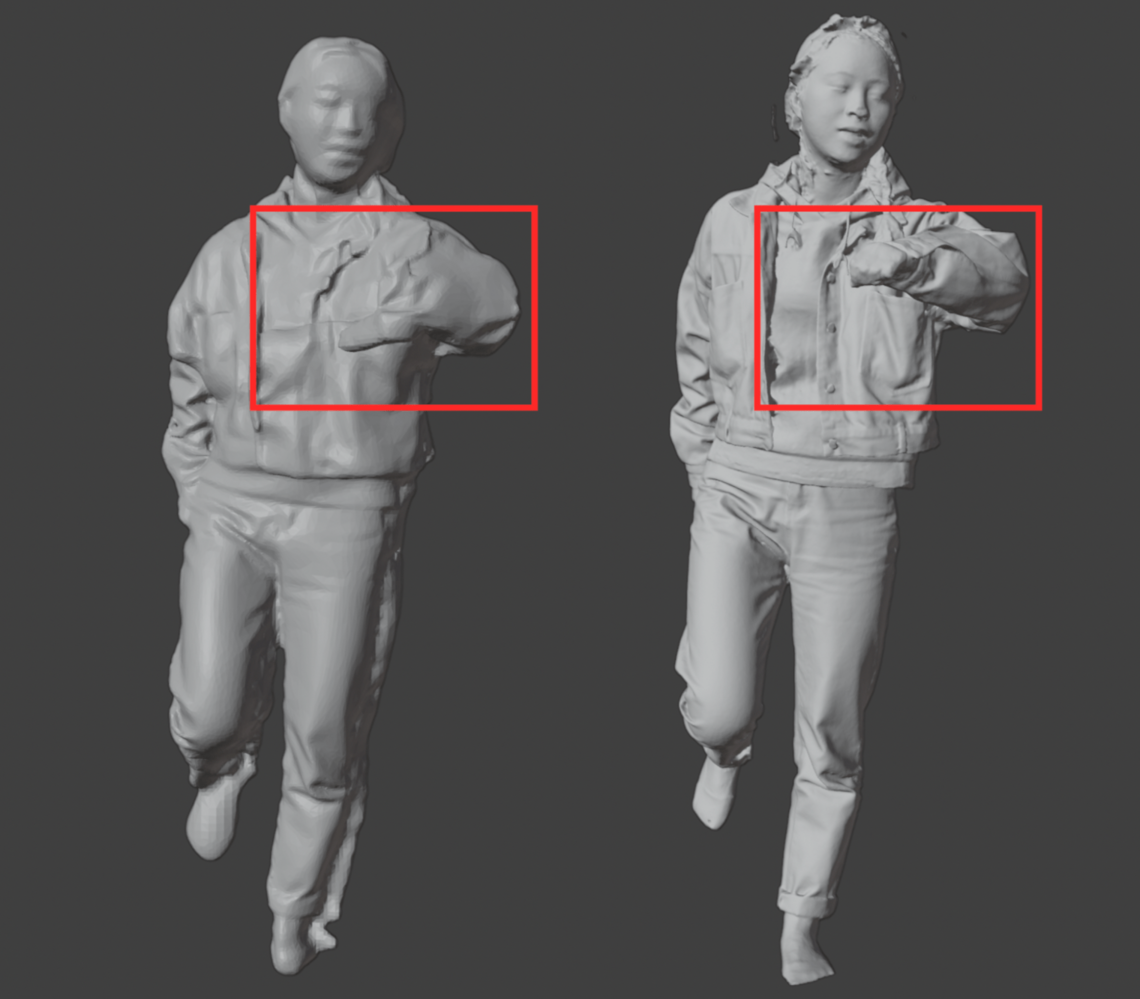} \\
    \end{tabular}
    \caption{Examples of reconstruction artifacts observed in MExECON outputs (left in each subfigure) compared to the corresponding ground-truth meshes (right). Red bounding boxes highlight common artifacts such as abnormal bulges or incomplete limb reconstruction.}
    \label{fig:artifact_examples}
\end{figure*}

Fig.~\ref{fig:qualitative_results} shows qualitative results for a subset of the test subjects. In the case of avatar 0021, the single-view ECON reconstruction fails to capture the geometry of the hooded garment on the back, producing a flattened and indistinct surface. In contrast, MExECON successfully reconstructs the corresponding geometry by incorporating information from the back-view image. Similarly, for avatar 0004, ECON exhibits an unnatural lean, while MExECON produces a standing posture as in the ground-truth mesh. Fig.~\ref{fig:qualitative_results} also shows a qualitative hierarchy: 8-view MExECON outperforms the 2-view variant, which in turn surpasses the single-view ECON baseline. VGGT results are not shown due to the significantly lower level of detail (as shown in Fig.~\ref{fig:vggt_comparison}), but are included in the appendix.

Note that Fig.~\ref{fig:qualitative_results} reveals more realistic hands in ECON avatars, achieved by replacing reconstructed hands with cropped SMPL-X hands. This step was omitted in MExECON but could be added without compatibility issues.

\subsection{Limitations and failure cases}
\label{sec:limitations}

{\bf{Calibrated cameras.}} MExECON assumes known, error-free camera parameters and the availability of front and back views, which are conditions restricted to synthetic data or controlled laboratory setups. Integrating automatic camera calibration would be essential for real scenarios.

\noindent 
{\bf{Static scene.}} The subject is assumed to remain static across the input images. Dynamic scenes would require estimating separate per-view SMPL-X pose parameters, potentially introducing inconsistencies in the JMBO algorithm (Sec.~\ref{sec:jmbo}).

\noindent
{\bf{Reconstruction artifacts.}} While MExECON produces high-quality geometry, it is not immune to artifacts. Visual inspection reveals that artifacts often appear when arms or legs are positioned close to the body, causing local surface ambiguities in the partial surface reconstructions. Such artifacts appear as unrealistic bulges in the transition area (subject 0083) or incomplete limb reconstructions (subject 0023 and 0071), as shown in Fig.~\ref{fig:artifact_examples}. We attribute these artifacts to the d-BiNI partial surface reconstruction and the IF-Nets+ surface completion stages, which struggle to estimate accurate depth information in these challenging areas.

\noindent
{\bf{Partial normal integration.}} MExECON retains ECON's front-back normal integration paradigm to avoid network re-training (Sec.~\ref{sec:front_back_rec}). While this is efficient, it means that geometric details from side views are not directly integrated into the final clothed surface. Their influence is limited to constraining the low-frequency JMBO prior. A full \mbox{$N$-view} normal integration could potentially capture finer details. However, this would require extending d-BiNI and re-training IF-Nets+, which are restricted to front and back inputs~\cite{xiu2023econ}.

\noindent
{\bf{Untextured output.}} Both ECON and MExECON output untextured meshes. Appearance realism could be improved by aggregating color from the input images via multi-view texturing post-processing. However, this aggregation presents challenges in shared regions that are visible in more than one view.
\section{Conclusion}
\label{sec:conclusion}

This work introduced MExECON, a multi-view pipeline for clothed 3D human avatar reconstruction. MExECON extends ECON~\cite{xiu2023econ} to support multi-view inputs while maintaining compatibility with the original single front-view framework. The body pose and shape of the output meshes are improved via the proposed JMBO algorithm, an optimization process that estimates a prior SMPL-X body model consistent across all views. The geometry of the output mesh is enhanced by normal map integration that captures details (clothing, hair) from front and back views, rather than relying solely on a front view. MExECON achieves multi-view gains with minimal overhead and no network re-training, while delivering performance competitive with modern few-shot multi-view 3D reconstruction approaches~\cite{wang2025vggt}. The main limitations are the need for a static scene and calibrated cameras, as well as the potential generation of artifacts after surface completion of areas occluded or unseen in the front and back views.
\clearpage

\appendix
\noindent
{\Large \bf{Appendix}}

\paragraph{Computation time and cost.} Table~\ref{tab:computational_cost} shows the computation time and GPU cost required by ECON~\cite{xiu2023econ}, MExECON and VGGT~\cite{wang2025vggt} to perform one avatar reconstruction. VGGT inference is the fastest, as it evaluates the entire transformer network at once in a single forward pass and does not involve iterative procedures like the SMPL-X optimization algorithms used in ECON and MExECON.

The difference in computation time between MExECON and ECON is mainly due to the SMPL-X body model initialization. MExECON sequentially computes separate per-view SMPL-X body models, which are averaged to initialize the proposed JMBO algorithm. While the initialization is more demanding, JMBO generally completes SMPL-X optimization in 35 iterations, compared to the 50 iterations employed in ECON. GPU memory usage in MExECON also increases with the number of views, as each JMBO iteration computes the gradients of the SMPL-X parameters with respect to all input views. However, this is an implementation choice rather than a methodological constraint. Minimizing the JMBO loss over smaller batches of different views across additional iterations could potentially reduce GPU memory demands to match those of baseline ECON.

\begin{table}[h]
  \centering
  { \small
  \begin{tabular}{@{\hskip 0.02cm}l@{\hskip 0.2cm}c@{\hskip 0.2cm}c}
    \cmidrule(lr){2-3}
     & Peak GPU memory & Computation time \\
    \midrule
    ECON \cite{xiu2023econ} & 8~GB & 45~sec. \\
    2-view MExECON & 12~GB & 1~min. \\
    8-view MExECON & 24~GB  & 2.5~min. \\ 
    8-view VGGT \cite{wang2025vggt} & 14~GB  & 5~sec. \\
    \bottomrule
  \end{tabular}
  }
  \caption{Approximate computation time and cost required by each method for one avatar reconstruction. All experiments were run on a single NVIDIA A100 40 GB GPU.}
  \label{tab:computational_cost}
\end{table}

\vspace{-0.2cm}

\begin{figure}[t]
\centering
    \begin{tabular}{@{\hskip -0cm}c@{\hskip 0.1cm}c}
    \rotatebox[origin=lb]{90}{\footnotesize \hspace{1.5cm} Avatar ID 0004} &
    \includegraphics[width=0.9\linewidth]{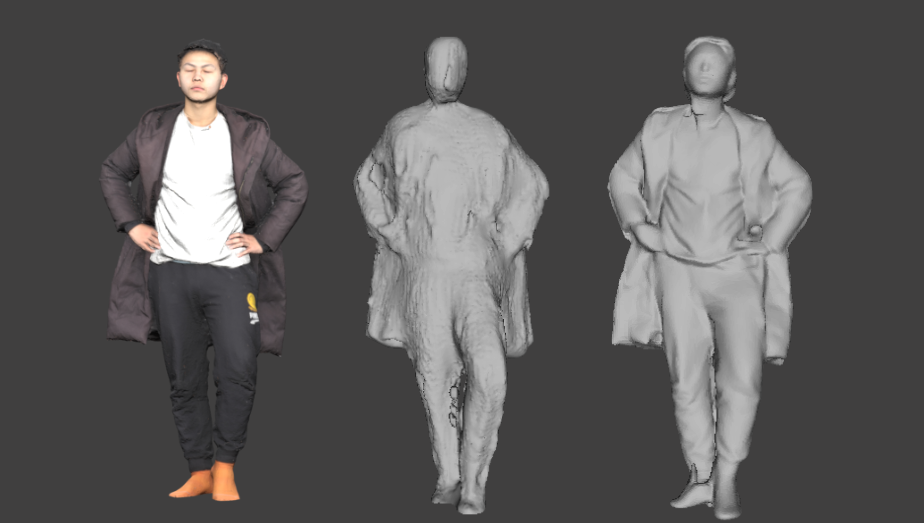} \\
    \rotatebox[origin=lb]{90}{\footnotesize \hspace{1.5cm} Avatar ID 0021} &
    \includegraphics[width=0.9\linewidth]{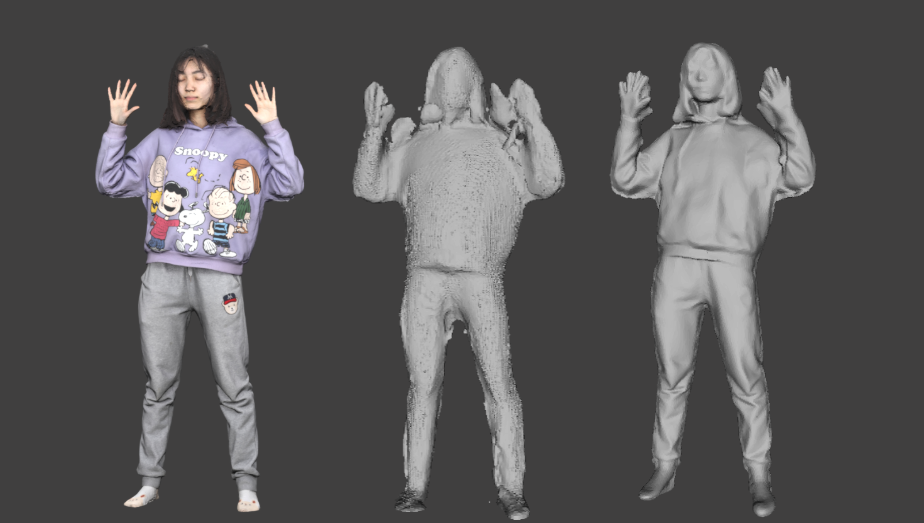} \\
    \rotatebox[origin=lb]{90}{\footnotesize \hspace{1.5cm} Avatar ID 0070} &
    \includegraphics[width=0.9\linewidth]{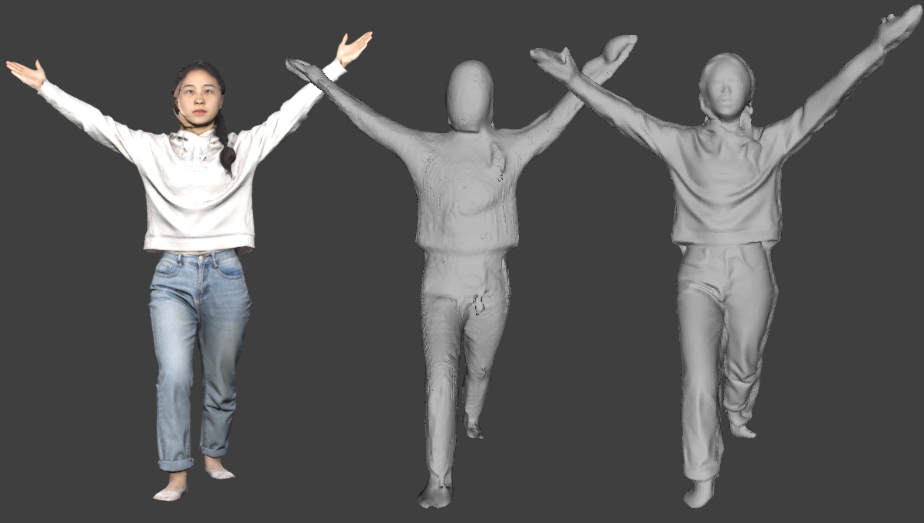} \\
    \rotatebox[origin=lb]{90}{\footnotesize \hspace{1.5cm} Avatar ID 0445} &
    \includegraphics[width=0.9\linewidth]{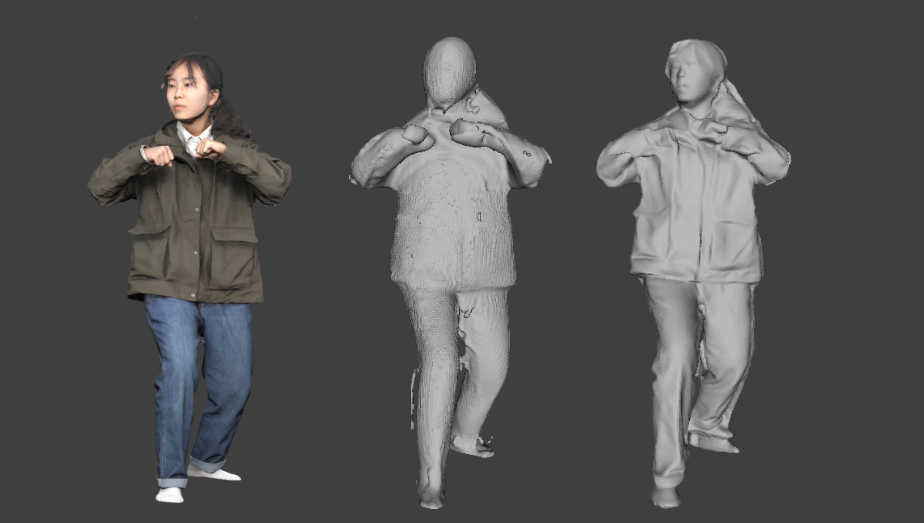} \\
    \end{tabular}
    \caption{VGGT vs MExECON qualitative results (8 input views). Left to right: front RGB view of the ground-turth avatar, VGGT reconstruction, MExECON reconstruction.}
    \label{fig:vggt_supp}
\end{figure}

\paragraph{VGGT comparison.} Fig.~\ref{fig:vggt_supp} shows full-body clothed avatar reconstructions by MExECON and VGGT~\cite{wang2025vggt}, using the same set of 8 input RGB images. The MExECON pipeline relies on the known camera parameters of the input views, whereas VGGT jointly estimates camera poses alongside the geometric reconstruction.\footnote{All VGGT results use the official implementation default parameters proposed in \textcolor{blue}{\href{https://github.com/facebookresearch/vggt}{https://github.com/facebookresearch/vggt}}.} Visual comparison shows that VGGT delivers accurate global shape and complete surface coverage but falls short in capturing the fine-grained geometric details that are well preserved in the MExECON results.

\setkeys{Gin}{draft=false}
\setkeys{Gin}{draft}

\setkeys{Gin}{draft=false}
\begin{figure*}[]
\centering
    \begin{tabular}{@{\hskip -0cm}c@{\hskip 0.1cm}c@{\hskip 0.1cm}c@{\hskip 0.1cm}c@{\hskip 0.1cm}c@{\hskip 0.1cm}c}
     & RGB images & ECON & 2-view MExECON & 8-view MExECON & Ground truth \\
     \rotatebox[origin=lb]{90}{\footnotesize \hspace{1.5cm} Avatar ID 0006} &
\includegraphics[width=0.19\linewidth]{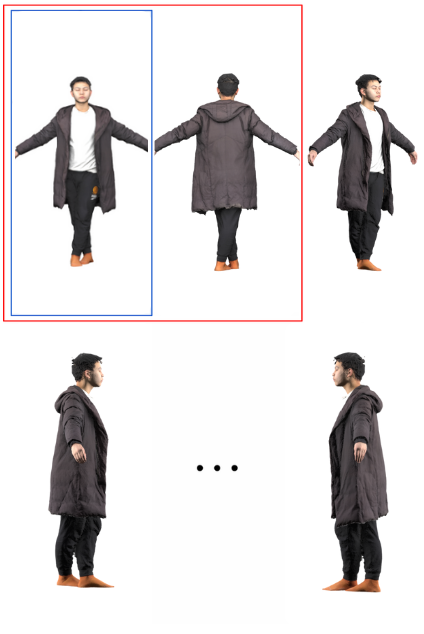} &
    \includegraphics[width=0.19\linewidth]{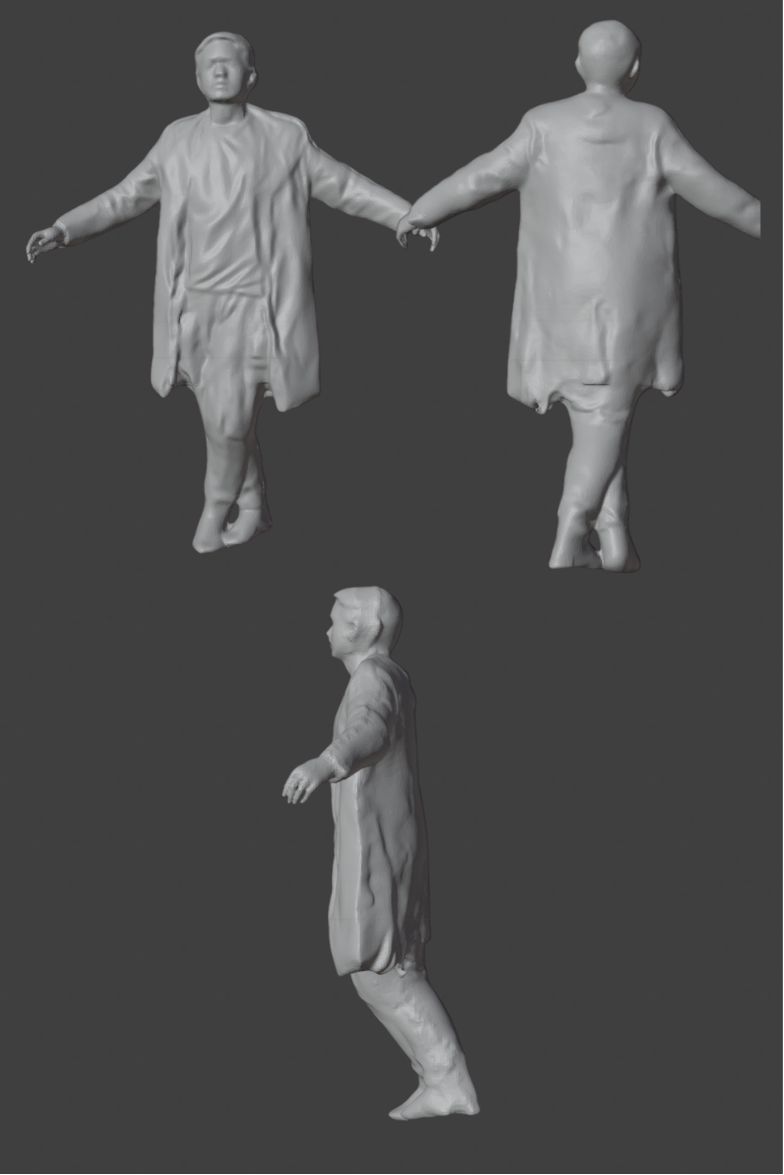} &
    \includegraphics[width=0.19\linewidth]{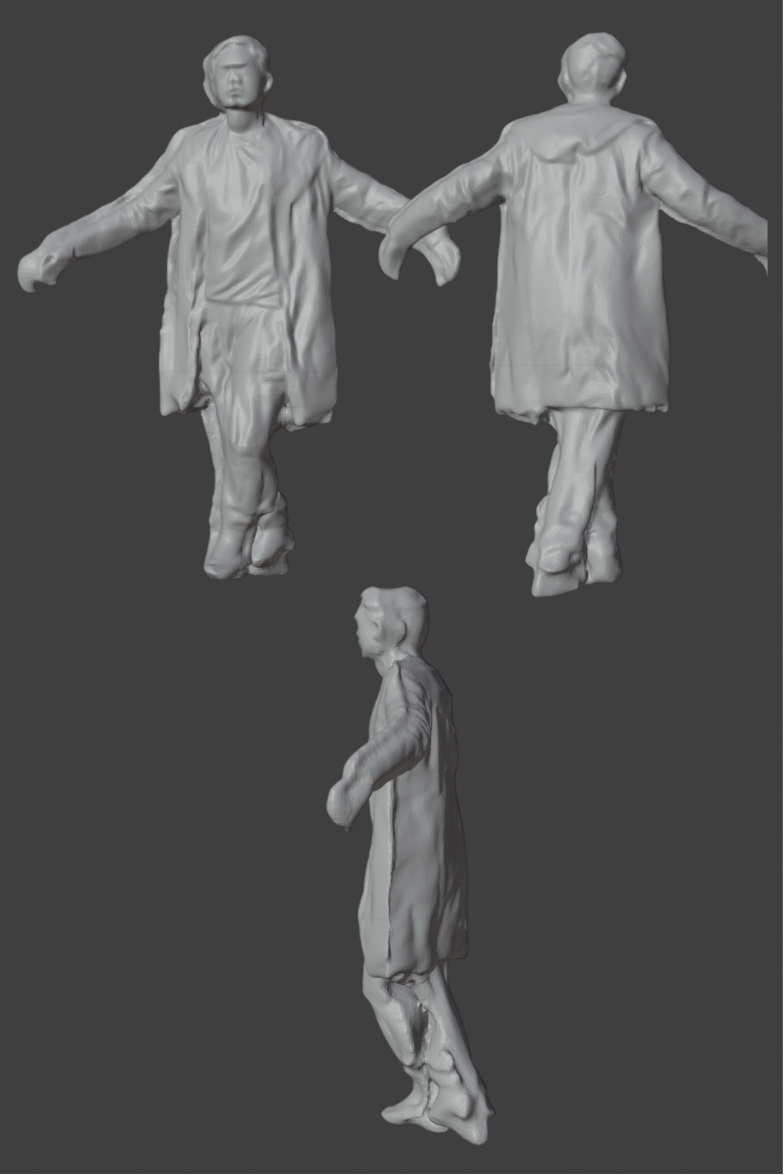} &
    \includegraphics[width=0.19\linewidth]{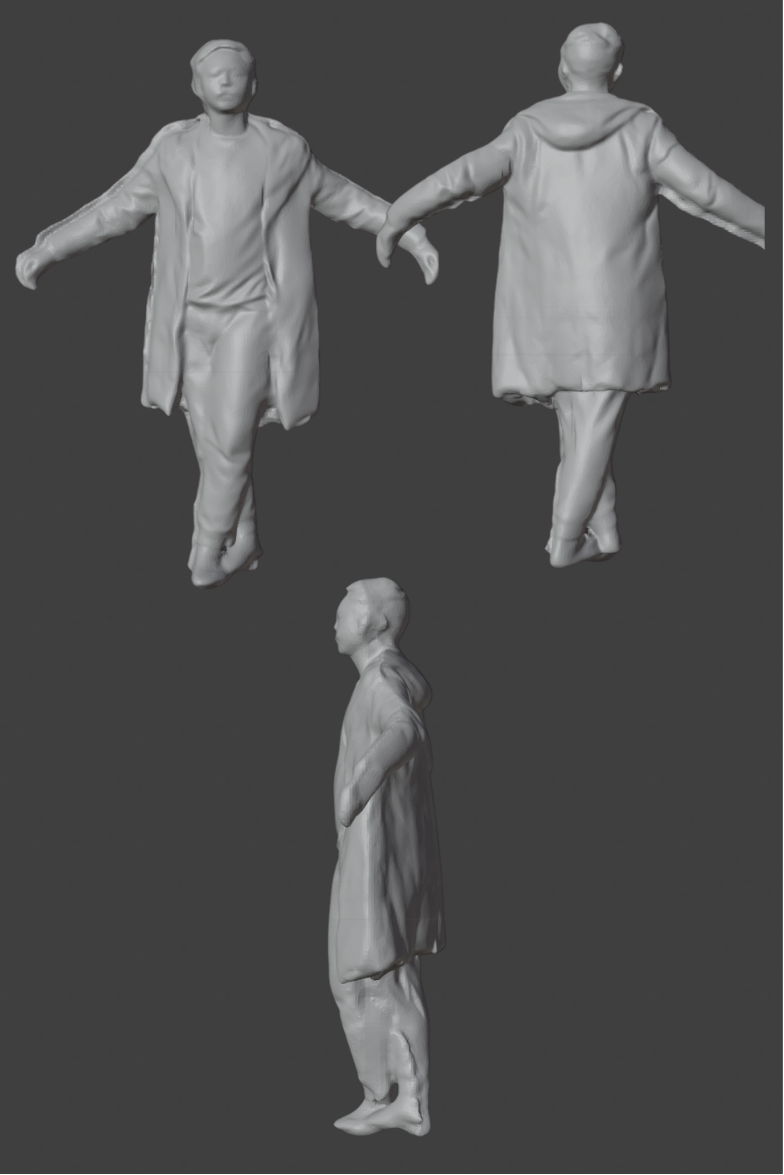} &
    \includegraphics[width=0.19\linewidth]{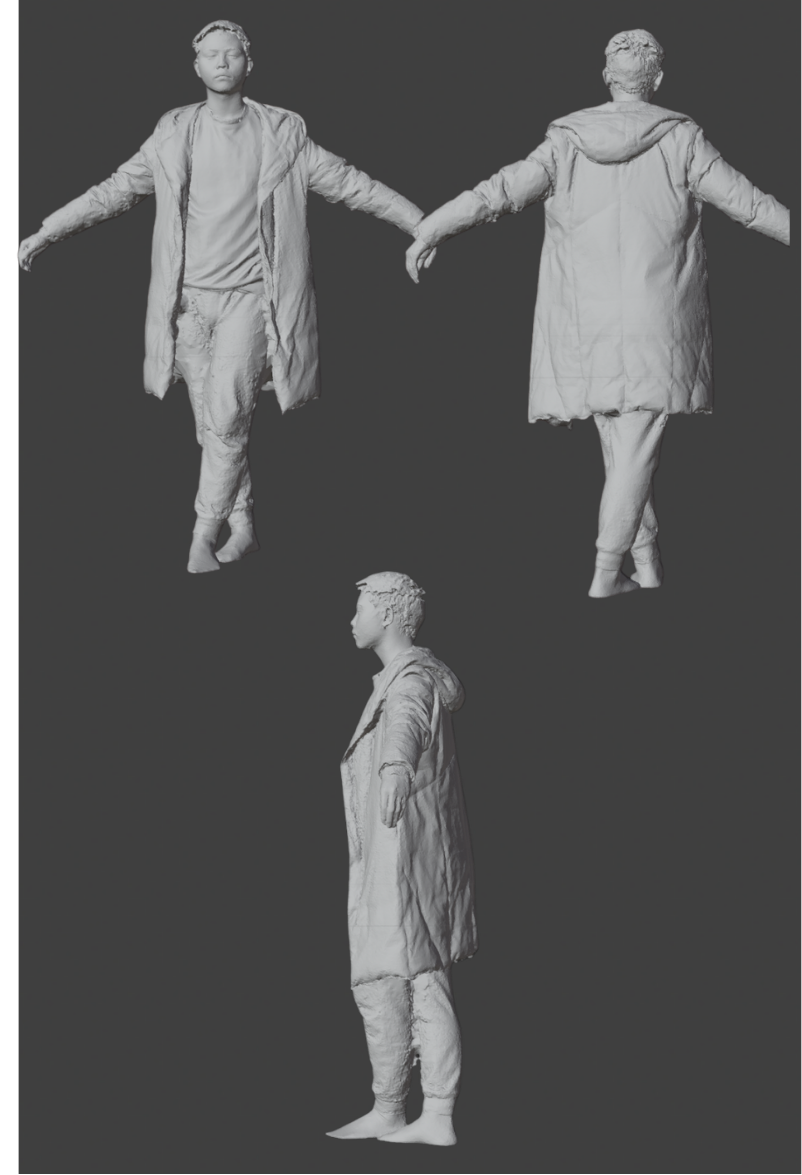} \\
    \rotatebox[origin=lb]{90}{\footnotesize \hspace{1.5cm} Avatar ID 0007} &
\includegraphics[width=0.19\linewidth]{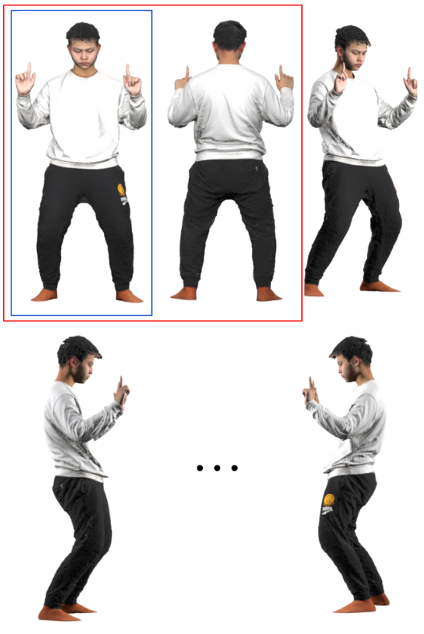} &
    \includegraphics[width=0.19\linewidth]{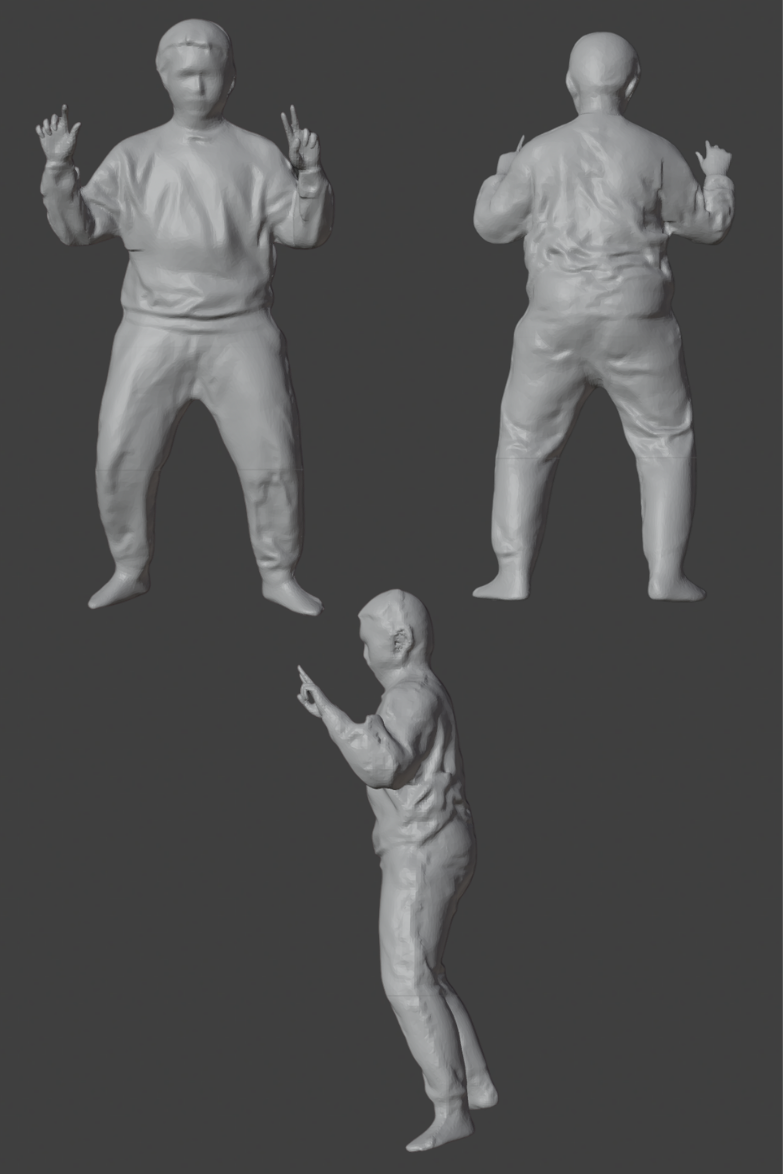} &
    \includegraphics[width=0.19\linewidth]{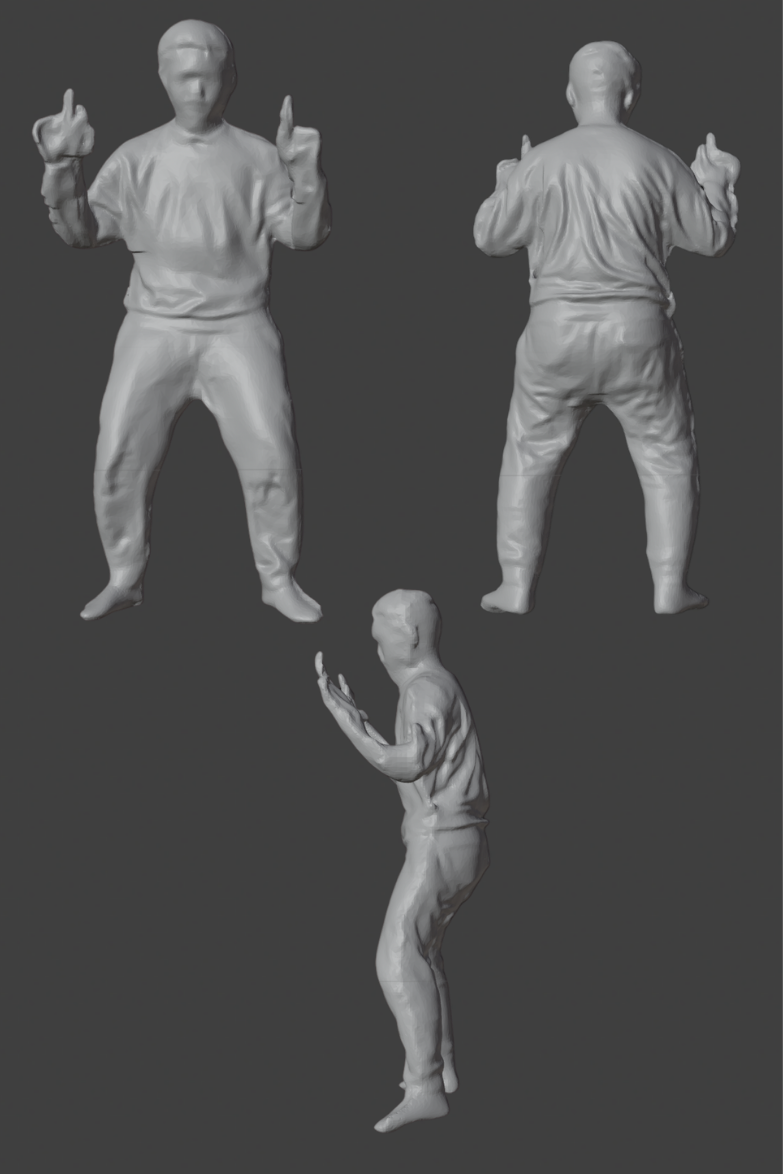} &
    \includegraphics[width=0.19\linewidth]{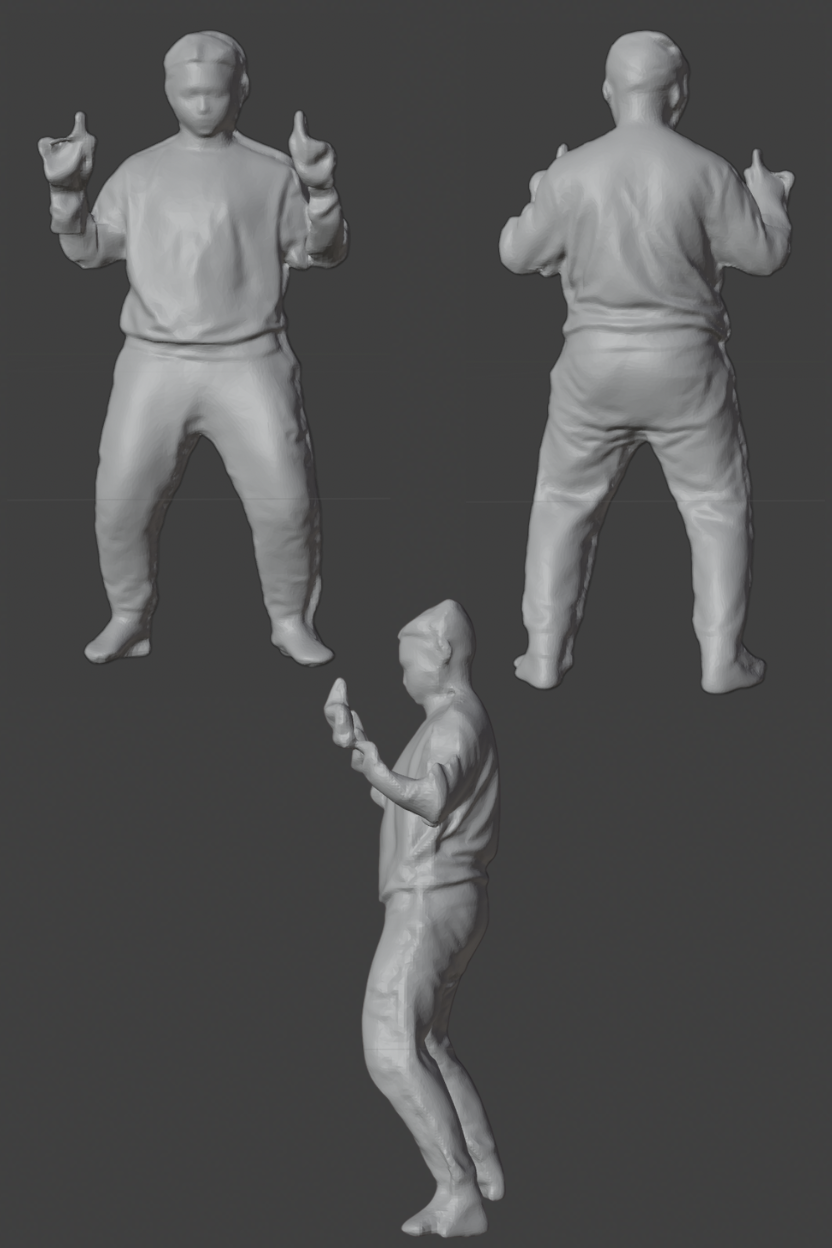} &
    \includegraphics[width=0.19\linewidth]{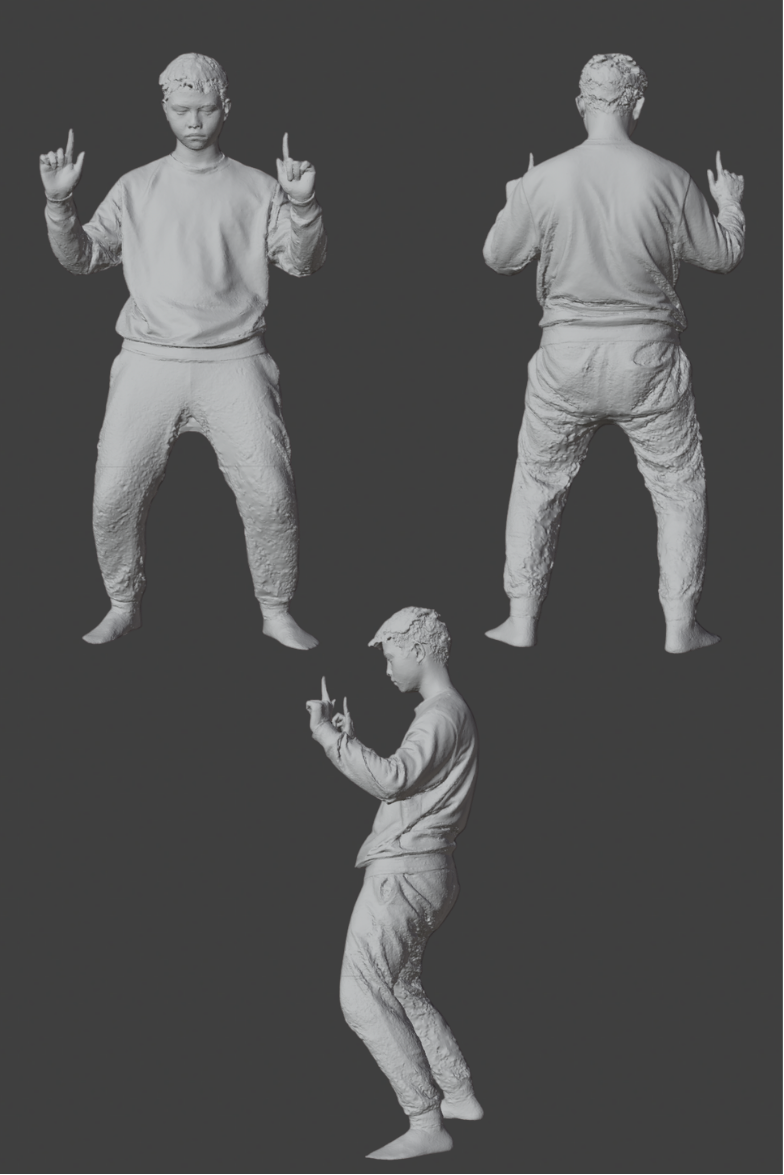} \\
\rotatebox[origin=lb]{90}{\footnotesize \hspace{1.5cm} Avatar ID 0057} &
\includegraphics[width=0.19\linewidth]{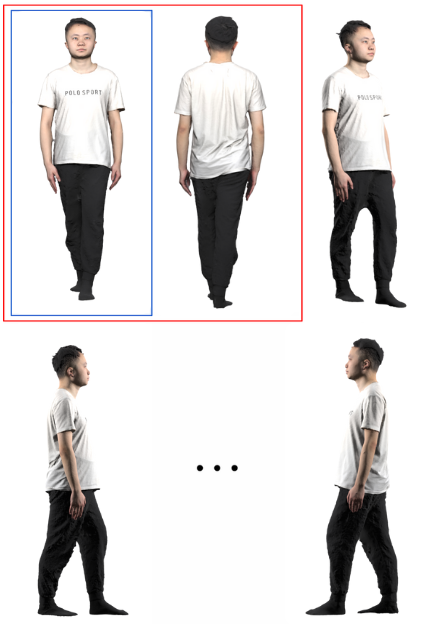} &
    \includegraphics[width=0.19\linewidth]{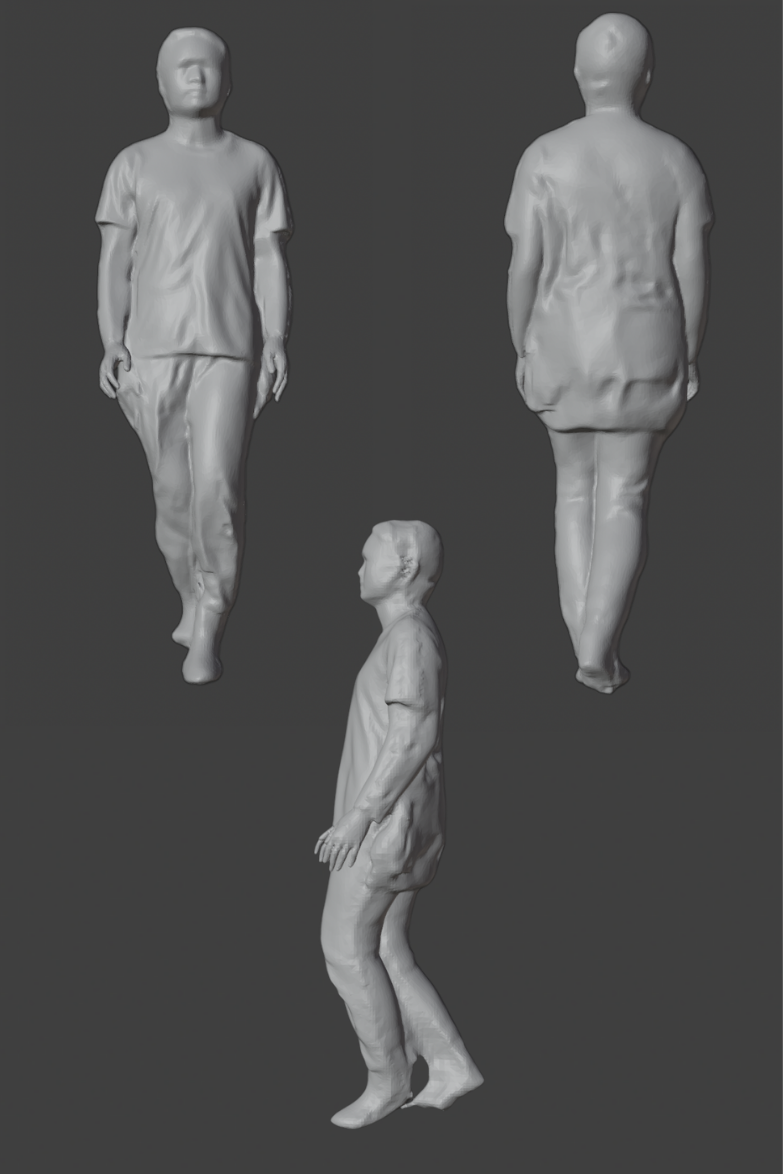} &
    \includegraphics[width=0.19\linewidth]{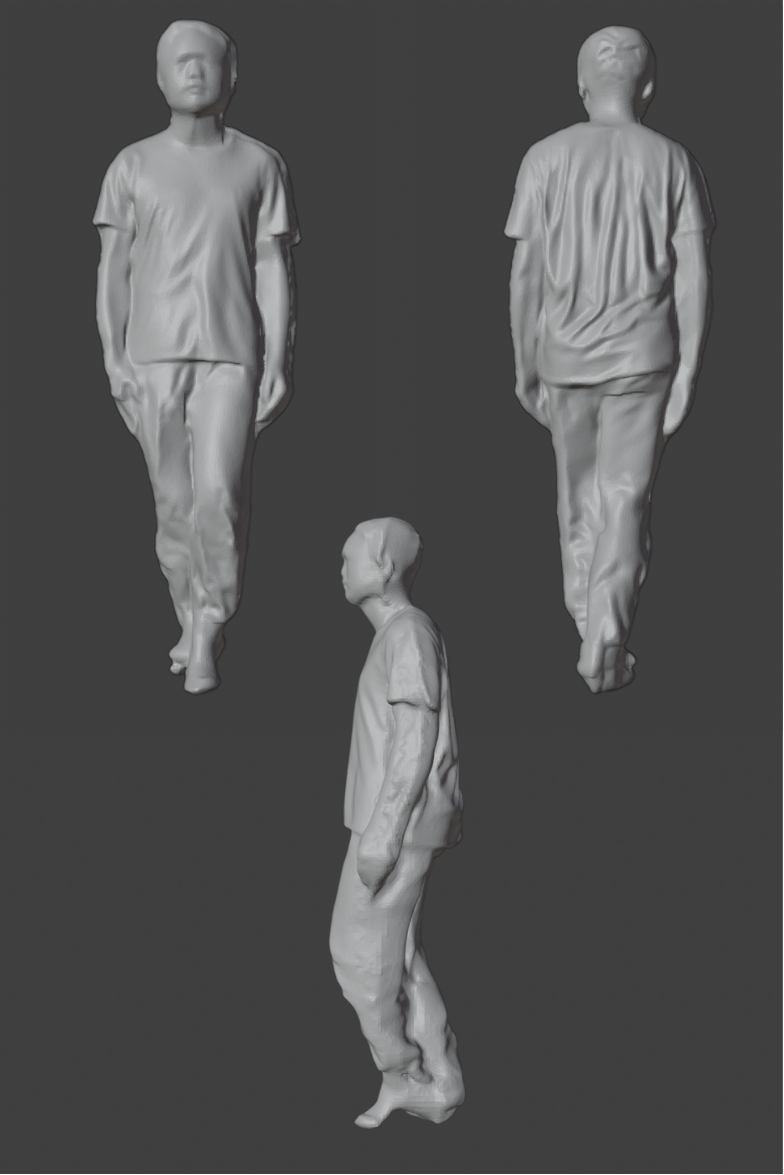} &
    \includegraphics[width=0.19\linewidth]{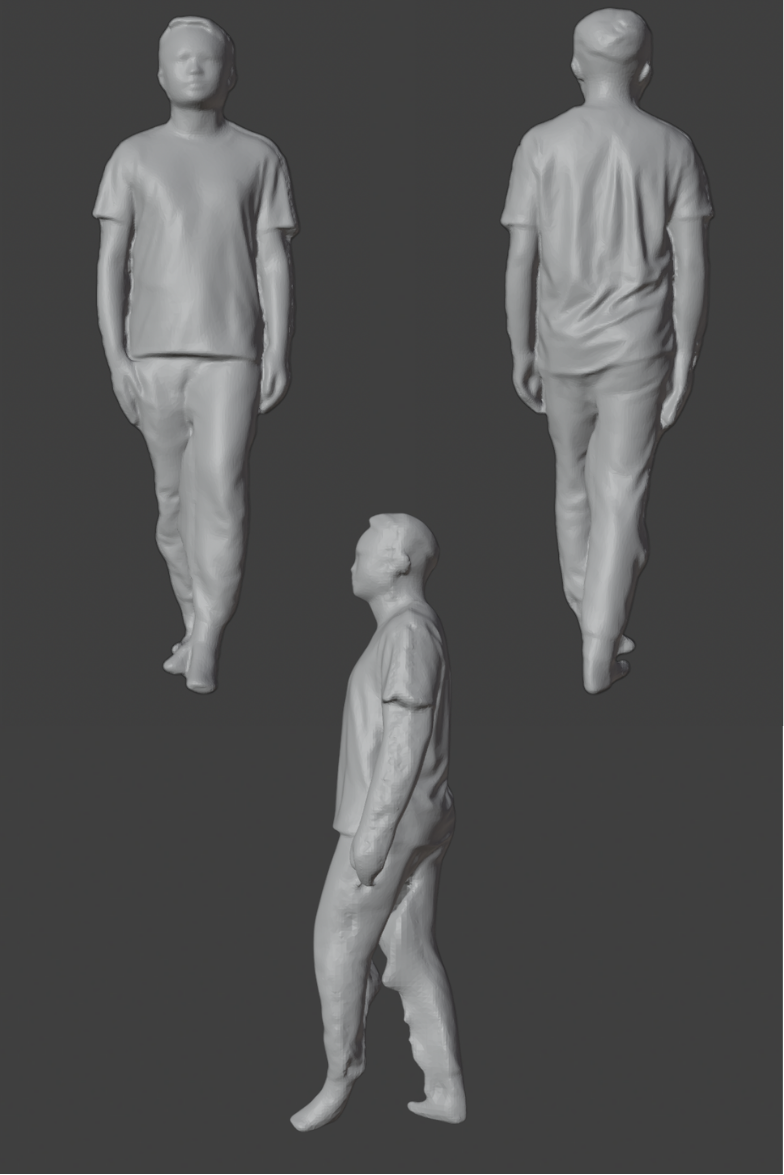} &
    \includegraphics[width=0.19\linewidth]{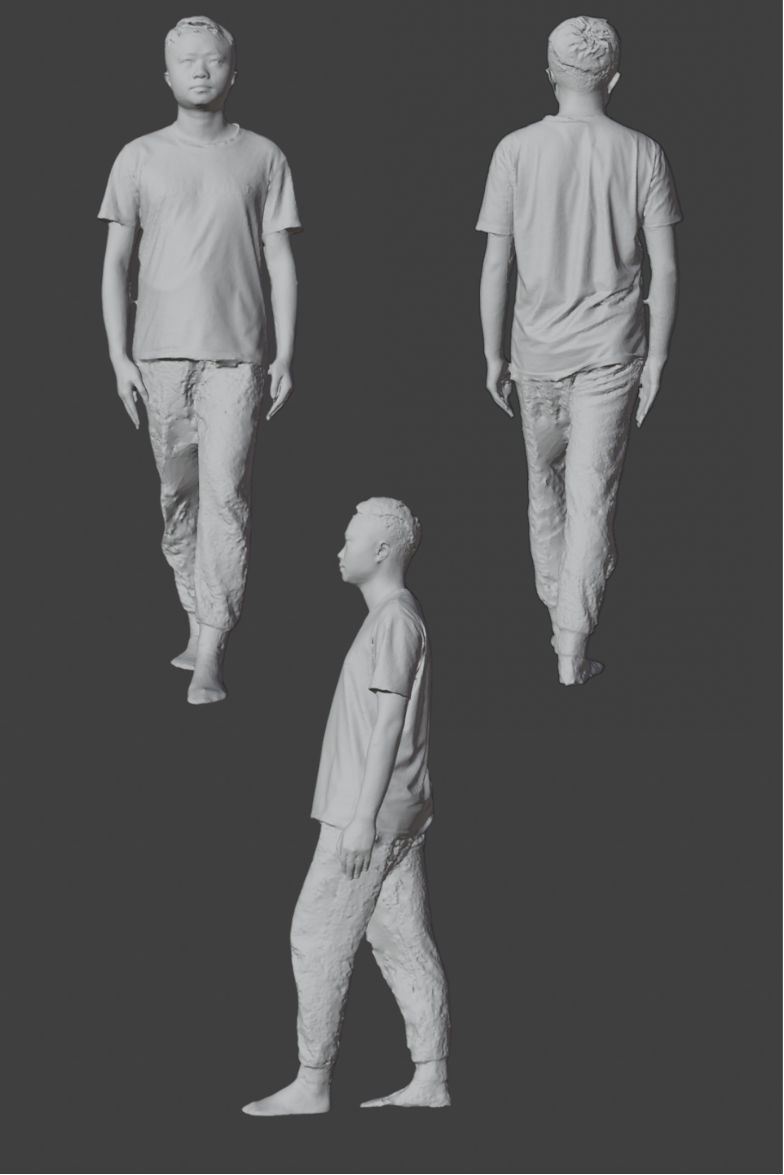} \\
    \rotatebox[origin=lb]{90}{\footnotesize \hspace{1.5cm} Avatar ID 0181} &
\includegraphics[width=0.19\linewidth]{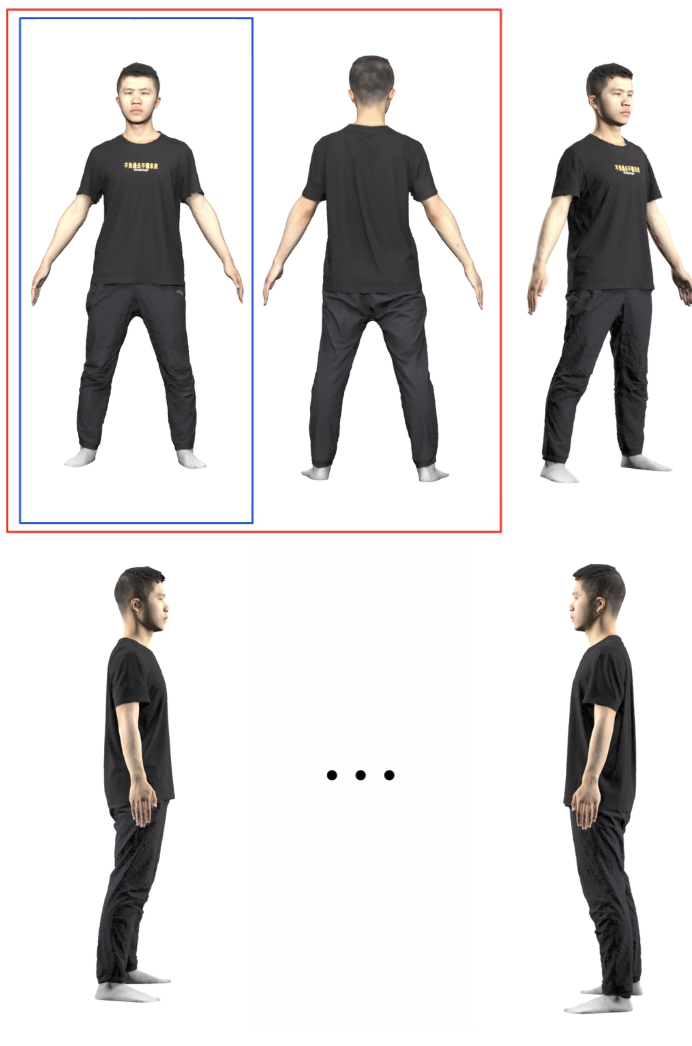} &
    \includegraphics[width=0.19\linewidth]{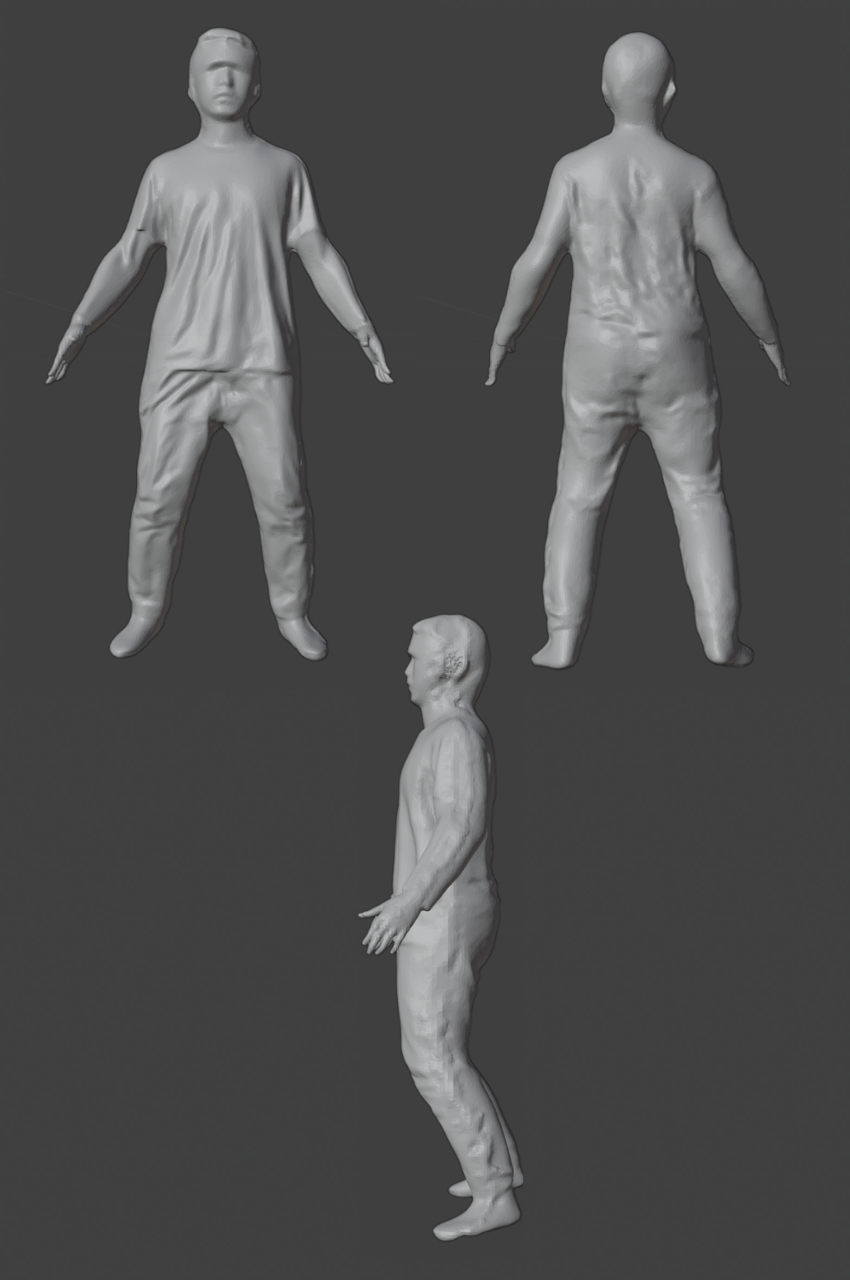} &
    \includegraphics[width=0.19\linewidth]{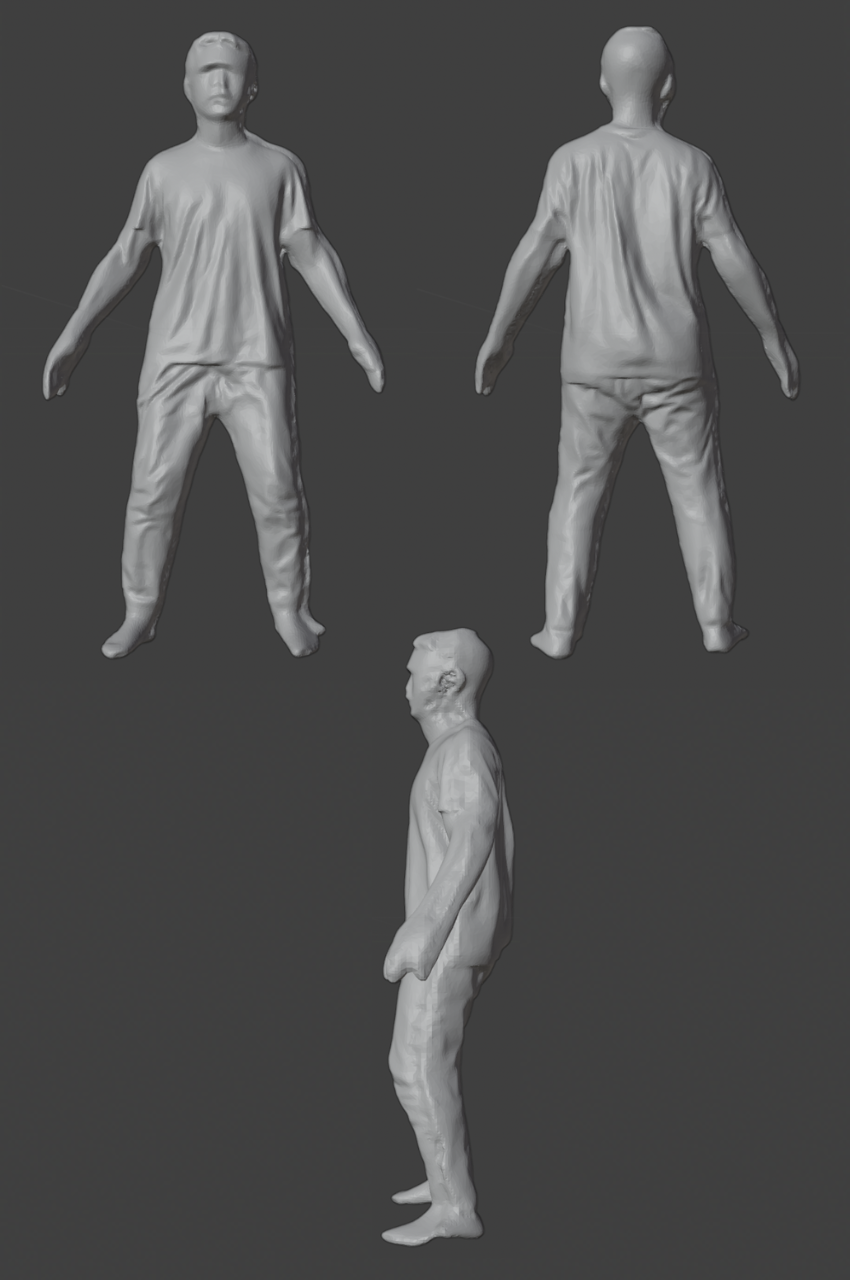} &
    \includegraphics[width=0.19\linewidth]{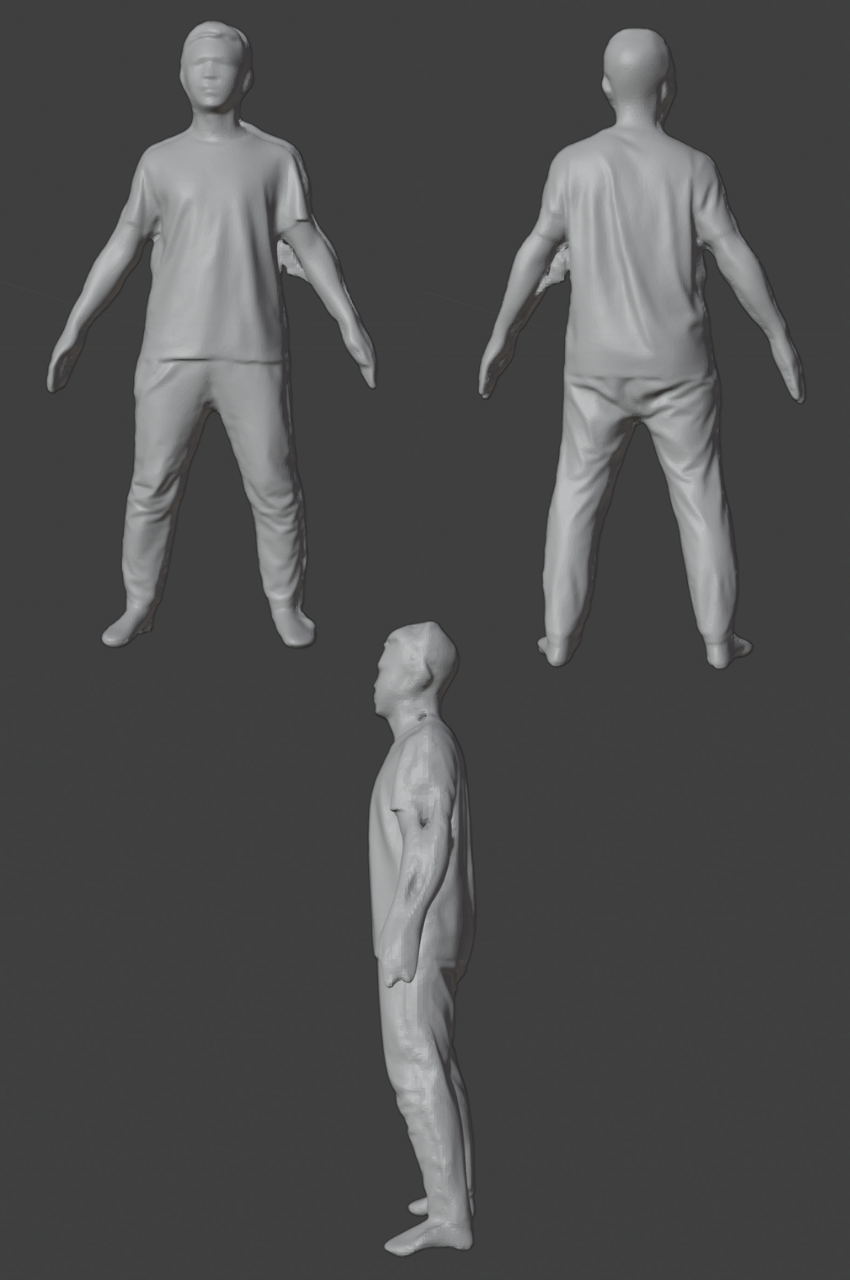} &
    \includegraphics[width=0.19\linewidth]{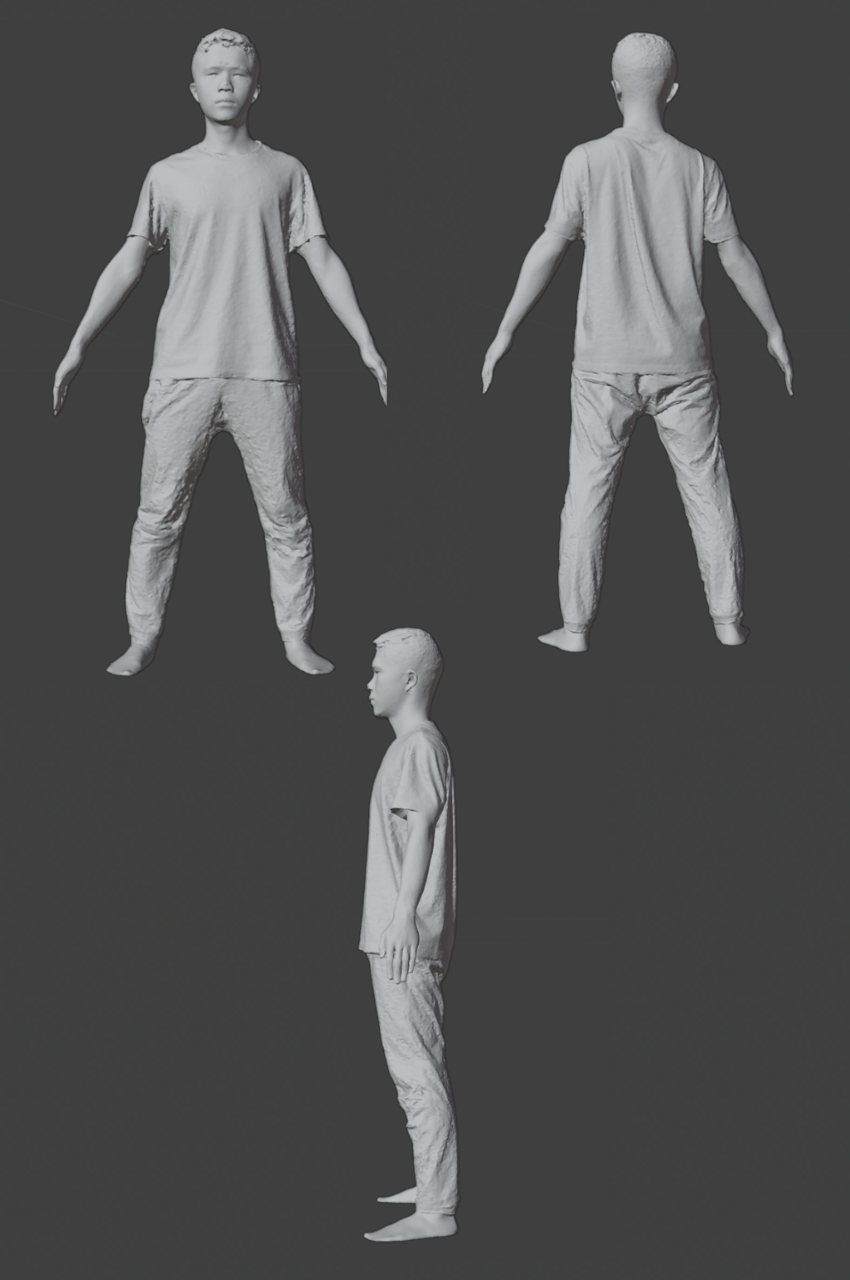} \\
    \end{tabular}
    \caption{Additional qualitative results. ECON takes a single front-view input (blue bounding boxes), 2-view MExECON uses front and back views (red bounding boxes), and 8-view MExECON uses all available views. MExECON yields more accurate body shape and pose estimation, along with enhanced geometric detail (particularly in the back). All meshes are shown from 3 viewpoints for clarity. 
    }
    \label{fig:qualitative_results_supp}
\end{figure*}
\setkeys{Gin}{draft}

\paragraph{Additional qualitative results.} Fig.~\ref{fig:qualitative_results_supp} shows qualitative results between ECON~\cite{xiu2023econ}, 2-view MExECON and 8-view MExECON for another group of selected test set avatars from THuman 2.1~\cite{tao2021function4d}. Visual comparison further confirms MExECON's superiority in recovering accurate shapes, poses, and fine details, especially for 8 views.

Note that Fig.~\ref{fig:qualitative_results_supp} shows more realistic hands in ECON avatars, achieved by replacing reconstructed hands with cropped SMPL-X hands. This step can also be incorporated into MExECON without introducing compatibility issues.

{
    \small
    \bibliographystyle{ieeenat_fullname}
    \bibliography{main}
}

\end{document}